\documentclass[dvipsnames]{article} %
\usepackage{colm2024_conference}
\usepackage{titletoc}
\usepackage{titlesec}
\usepackage{booktabs}
\usepackage{graphicx}
\usepackage{enumitem}
\usepackage{wrapfig}
\usepackage{algorithm}
\usepackage{algpseudocode}
\usepackage{pifont}
\usepackage{tikz}
\usetikzlibrary{fadings}
\usepackage{microtype}
\usepackage{amsmath}
\usepackage{colortbl}
\usepackage[utf8]{inputenc}
\definecolor{lightgray}{rgb}{0.9,0.9,0.9}
\usepackage{caption}
\usepackage{subcaption}
\usepackage{setspace}
\usepackage{url}
\usepackage{multirow}
\usepackage{colortbl}
\usepackage{tocloft}
\usepackage{tabularx}
\usepackage{blindtext}
\usepackage{pgfplots}
\usepackage{wrapfig}
\pgfplotsset{compat=1.18} 
\usepackage{tikz}
\usetikzlibrary{er,positioning,bayesnet}
\usepackage{makecell}
\usepackage{tipa}
\usepackage{siunitx}
\usepackage{nicefrac}
\usepackage{tocloft}
\usepackage{listings}
\usepackage[raster,skins]{tcolorbox} %
\usepackage{xltabular}
\usepackage{adjustbox}
\usepackage{xurl}
\usepackage{rotating}
\usepackage[normalem]{ulem}
\usepackage{multicol}
\useunder{\uline}{\ul}{}
\definecolor{darkmagenta}{RGB}{112, 40, 112}%{rgb}{0.56, 0.0, 1.0}  

\usepackage{titlesec}
\titleclass{\subsubsubsection}{straight}[\subsection]
\newcounter{subsubsubsection}[subsubsection]
\renewcommand{\thesubsubsubsection}{\thesubsubsection.\arabic{subsubsubsection}}
\titleformat{\subsubsubsection}
  {\normalfont\normalsize\mdseries}{\thesubsubsubsection}{1em}{}
\titlespacing*{\subsubsubsection}
{0pt}{3.25ex plus 1ex minus .2ex}{1.5ex plus .2ex}

\makeatletter
\newcommand\l@subsubsubsection{\@dottedtocline{4}{7em}{4em}}
\renewcommand\l@paragraph[2]{}
\makeatother

\hypersetup{colorlinks,linkcolor={blue},citecolor={darkmagenta},urlcolor={blue}}
\usepackage{hyperref}
\usepackage{xcolor}
\usepackage{graphicx} % 用于 \scalebox
\definecolor{colorA}{HTML}{FF0000} % 红色
\definecolor{colorB}{HTML}{FF8800} % 橙色
\definecolor{colorC}{HTML}{FFFF00} % 黄色
\definecolor{colorD}{HTML}{0000FF} % 深蓝色
\definecolor{Cornsilk}{rgb}{1, 0.95, 0.88}%用于table底色

\usepackage{amsmath,amsfonts,bm}

\def\eqref#1{equation~\ref{#1}}
\def\1{\bm{1}}

\DeclareMathAlphabet{\mathsfit}{\encodingdefault}{\sfdefault}{m}{sl}
\SetMathAlphabet{\mathsfit}{bold}{\encodingdefault}{\sfdefault}{bx}{n}

\newcommand*\justify{%
  \fontdimen2\font=0.4em% interword space
  \fontdimen3\font=0.2em% interword stretch
  \fontdimen4\font=0.1em% interword shrink
  \fontdimen7\font=0.1em% extra space
  \hyphenchar\font=`\-% allowing hyphenation
}

\renewcommand{\texttt}[1]{%
  \begingroup
  \ttfamily
  \begingroup\lccode`~=`/\lowercase{\endgroup\def~}{/\discretionary{}{}{}}%
  \begingroup\lccode`~=`[\lowercase{\endgroup\def~}{[\discretionary{}{}{}}%
  \begingroup\lccode`~=`.\lowercase{\endgroup\def~}{.\discretionary{}{}{}}%
  \catcode`/=\active\catcode`[=\active\catcode`.=\active
  \justify\scantokens{#1\noexpand}%
  \endgroup
}

\title{Let It Flow: Agentic Crafting on Rock and Roll\\[1ex]\large\textit{Building the ROME Model within an Open Agentic Learning Ecosystem}}

\author{
\bf ROCK \& ROLL \& IFLOW \& DT Joint Team
}

\begin{document}

\maketitle

\begin{abstract}

\emph{Agentic crafting}, unlike one-shot response generation for simple tasks, requires LLMs to operate in real-world environments over multiple turns—taking actions, observing outcomes, and iteratively refining artifacts until complex requirements are satisfied. Yet the spirit of agentic crafting reaches beyond code, into broader tool- and language-mediated workflows where models must plan, execute, and remain reliable under interaction. Reaching this new regime demands sustained, painstaking effort to build an agentic ecosystem as the foundational bedrock, ultimately culminating in an agent model as the capstone. \textbf{ROME wasn't built in a day.} A principled, end-to-end agentic ecosystem can streamline the development of the agent LLMs from training to production deployment, accelerating the broader transition into the agent era. However, the open-source community still lacks such an ecosystem, which has hindered both practical development and production adoption of agents. To this end, we introduce the \textbf{\textcolor{orange}{A}gentic \textcolor{orange}{L}earning \textcolor{orange}{E}cosystem} (\textcolor{orange}{\texttt{ALE}}), a foundational infrastructure that optimizes the end-to-end production pipeline for agent LLMs. \textcolor{orange}{\texttt{ALE}} consists of three system components. \textbf{ROLL} is a post-training framework for weight optimization. \textbf{ROCK} is a sandbox environment manager that orchestrates environments for trajectory generation. \textbf{iFlow CLI} is an agent framework that enables configurable and efficient context engineering for environment interaction. We release \textbf{\textcolor{orange!80!black}{ROME}} (\textbf{\underline{\textcolor{orange!80!black}{R}}OME is \underline{\textcolor{orange!80!black}{O}}bviously an Agentic \underline{\textcolor{orange!80!black}{M}}od\underline{\textcolor{orange!80!black}{E}}l}), an open-source agent grounded by \textcolor{orange}{\texttt{ALE}} and trained on over one million trajectories. In addition, we curate a suite of data composition protocols that synthesize data spanning isolated, static snippets to dynamic, complex agentic behaviors, with built-in verification of safety, security, and validity. We further develop an end-to-end training pipeline and propose a novel policy optimization algorithm \textcolor{orange}{\texttt{IPA}}, which assigns credit over \emph{semantic interaction chunks} rather than individual tokens, improving training stability over long horizons. 
Empirical evaluations show that \textbf{\textcolor{orange!80!black}{ROME}} achieves strong results across mainstream agentic benchmarks, including 24.72\% on \texttt{Terminal-Bench 2.0} and 57.40\% accuracy on \texttt{SWE-bench Verified}, outperforming similarly sized models and rivaling those with over 100B parameters.
To enable more rigorous evaluation, we introduce \textbf{Terminal Bench Pro}, a benchmark with improved scale, domain coverage, and contamination control. \textbf{\textcolor{orange!80!black}{ROME}} still demonstrates competitive performance among open-source models of similar scale and has been successfully deployed in production, demonstrating the practical effectiveness of the \textcolor{orange}{\texttt{ALE}}.
%Empirically, we evaluate \textbf{\textcolor{orange!80!black}{ROME}} within a structured agentic evaluation setting and introduce \textbf{Terminal Bench Pro} as a complementary benchmark with improved scale, domain balance, and contamination control. Under this evaluation regime, our model consistently demonstrates strong performance across multiple benchmarks (e.g., \texttt{SWE-bench Verified} and \texttt{Terminal Bench}), demonstrating the robustness and production-level effectiveness of \textcolor{orange}{\texttt{ALE}}.
\end{abstract}

\begin{figure}[!h]
    \vspace{-1mm}
    \centering
    \includegraphics[width=0.9\textwidth]{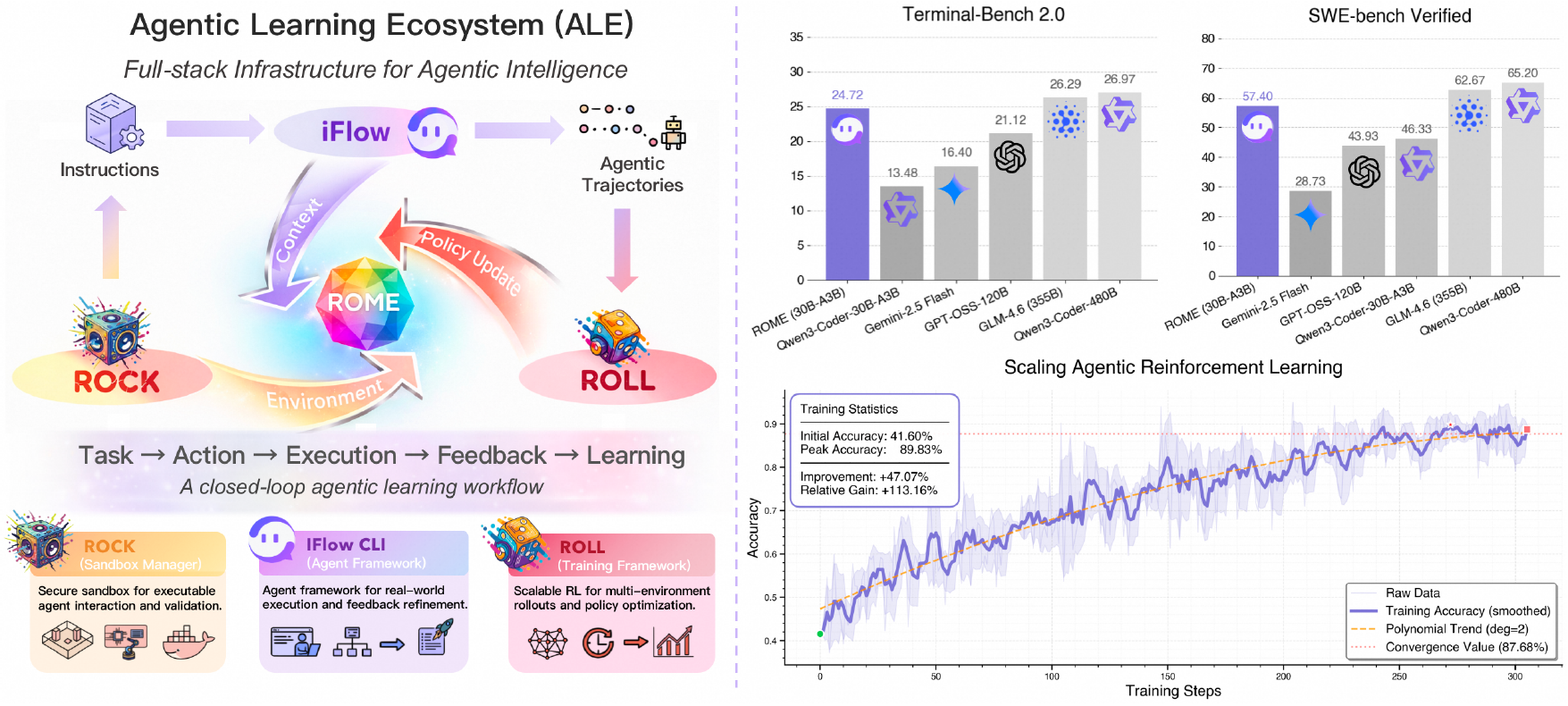}
    \vspace{-2mm}
    \caption{Overview of the \textbf{\textcolor{orange}{A}gentic \textcolor{orange}{L}earning \textcolor{orange}{E}cosystem} (\texttt{\textcolor{orange}{ALE}}) and \textbf{\textcolor{orange!80!black}{ROME}} Performance.}
    \label{fig:intro}
\end{figure}

\vfill

\newpage
\setcounter{tocdepth}{4}         
\setlength{\cftbeforesecskip}{15pt}

\setlength{\cftbeforesubsecskip}{6pt}

\setlength{\cftbeforesubsubsecskip}{6pt}

{
    \hypersetup{linkcolor=orange} 
    \tableofcontents
}

\newpage
\section{Introduction}
\label{sec:intro}
Recent years have witnessed a transformative wave in software engineering driven by large language models (LLMs)~\citep{hou2024largelanguagemodelssoftware}. Early efforts largely cast LLMs as one-shot generators, emitting static responses to a single prompt~\citep{TOSEM25,TraditionalCodeGenration,hou2024largelanguagemodelssoftware}. Yet this paradigm provides limited iterative reasoning and lacks grounded feedback loops, rendering it ill-suited for complex, end-to-end workflows. Accordingly, the frontier of LLM-based workflow-driven task (e.g., software engineering) is shifting toward the \emph{agentic crafting}\footnote{The \emph{agentic crafting} extends beyond writing code to encompass general-purpose, workflow-driven tasks (e.g, travel plan, GUI assistant) through multi-turn interactions with its environment.} paradigm, which enables LLMs to plan, execute, and self-correct through multi-turn interactions with environments, spanning software repositories, terminals and broader tool- and language-mediated workflows in the real world~\citep{deeptravel,gui-agent,mobile-agent,gao2023retrieval,novikov2025alphaevolve}.

However, the widespread practical adoption of agentic crafting remains elusive in the absence of a \emph{scalable, end-to-end agentic ecosystem}. Prior work has sought to improve agentic crafting via supervised fine-tuning (SFT) on limited human demonstrations~\citep{AgenticSFTDataset,Klear-AgentForge}, or through ad-hoc reinforcement learning (RL) recipes that are often struggles with long-horizon tasks and sparse, delayed rewards~\citep{deepswe2025,rLLM,Klear-AgentForge}. In this report, we contend that a principled agentic ecosystem must close the loop spanning \emph{data generation}, \emph{agent execution}, and \emph{policy optimization}, enabling an continuous end-to-end optimization workflow that can adapt to distribution shift and growing complexity in production environments. To bridge this gap, we present the \textbf{\textcolor{orange}{A}gentic \textcolor{orange}{L}earning \textcolor{orange}{E}cosystem} (\textcolor{orange}{\texttt{ALE}}), a full-stack \emph{infrastructure} that unifies data, training, and deployment for agentic intelligence. Concretely, \textcolor{orange}{\texttt{ALE}} comprises three synergistic system components:
\definecolor{aleorange}{RGB}{200, 95, 20}

\definecolor{blueviolet}{RGB}{80, 90, 25}
\newtcolorbox{insightblock}{
  colback=blueviolet!3!orange!4!,   
  colframe=blueviolet!30!orange!100!,    
  boxrule=0.5mm,       
  arc=2mm,            
  left=0pt,           
  right=8pt,           
  top=8pt,            
  bottom=8pt,}

% and optimization algorithms 
% 使用示例
\begin{insightblock}
\begin{enumerate}[leftmargin=1.5em]
    \item[] \textbf{ROLL} (Reinforcement Learning Optimization for Large-Scale Learning): A scalable RL training framework supporting multi-environment rollouts, chunk-aware credit assignment, and stable policy updates for long-horizon agentic tasks.
    \newline
    \item[] \textbf{ROCK} (Reinforcement Open Construction Kit): A secure, sandboxed agent execution platform that provides executable, tool-grounded environments, supporting interaction trajectory synthesis, execution, and validation.
    \newline
    \item[] \textbf{iFlow CLI}: An agent framework that orchestrates structured prompt suites for environment interaction, coupled with a user-facing interface that packages agents for real-world workflows and exposes APIs for continuous refinement via user feedback.
\end{enumerate}
\end{insightblock}

Grounded in \textcolor{orange}{\texttt{ALE}}, we incubate \textbf{\textcolor{orange!80!black}{ROME}} as an open-source agent LLM based on Qwen3-MoE, tightly developed within our established ecosystem. Along the road to \textbf{\textcolor{orange!80!black}{ROME}}, we take two deliberate steps. First, we establish a curated, coherent data composition workflow that synthesizes multi-source, multilingual, tool-grounded trajectories. Benefiting from strong sandbox isolation and fine-grained permission control of ROCK, we run rigorous security, safety, and validity verification to ensure the integrity and quality of the generated trajectories. Second, we leverage millions of high-quality trajectories to iteratively refine an efficient, stage-wise training pipeline from continuous pre-training, SFT, to RL. Enabled by the tight integration of our ecosystem, the end-to-end training pipeline remains both high-throughput, resource-efficient, and user-friendly. To further stabilize RL training dynamics, we propose \textbf{\textcolor{orange}{I}nteraction-\textcolor{orange}{P}erceptive \textcolor{orange}{A}gentic Policy Optimization} (\textcolor{orange}{\texttt{IPA}}), a novel algorithm that optimizes policies over \emph{semantic interaction chunks}~\citep{li2025reinforcement}. By shifting credit assignment from tokens to semantically meaningful chunks, \textcolor{orange}{\texttt{IPA}} improves long-horizon stability and ultimately strengthens long-context agentic crafting performance.

% Enabled by the tight integration within our ecosystem, the end-to-end workflow remains both high-throughput and resource-efficient. To steady the training dynamics of RL, we invoatively propose \textbf{\textcolor{orange}{I}nteraction-\textcolor{orange}{P}erceptive \textcolor{orange}{A}gentic Policy Optimization} (\textcolor{orange}{\texttt{IPA}}), a novel RL algorithm that optimizes policies over \emph{semantic interaction chunks}~\citep{li2025reinforcement}. By shifting credit assignment to semantic chunks, \textcolor{orange}{\texttt{IPA}} stablizes training dynamics over long horizons, ultimately strengthening long-context agentic coding performance.

Extensive empirical results demonstrate that \textbf{\textcolor{orange!80!black}{ROME}} achieves solid and consistent performance across a diverse set of agentic benchmarks.
On terminal-centric tasks, \textbf{\textcolor{orange!80!black}{ROME}} achieves 57.4\% accuracy on \texttt{SWE-bench Verified} and 24.7\% on \texttt{Terminal-Bench v2.0}, outperforming models of similar scale and approaching the performance of larger models exceeding 100B parameters. On the more rigorous \textbf{Terminal Bench Pro}, which enforces stricter contamination control and improved domain balance, \textbf{\textcolor{orange!80!black}{ROME}} still performs competitively, showing strong generalization and stability across domains. 
Furthermore, \textbf{\textcolor{orange!80!black}{ROME}} has been integrated into \texttt{iFlow CLI} and stably deployed in production. This real-world validation, together with \textcolor{orange}{\texttt{ALE}}, establishes a robust, scalable, and production-grade foundation for the continual training and enhancement of \textbf{\textcolor{orange!80!black}{ROME}}.

%substantially outperforming prior SFT-only baselines~\citep{} and exhibiting strong generalization to previously unseen toolchains. Moreover, \textbf{\textcolor{orange!80!black}{ROME}} has been deployed in production for a long time, empowering the iFlow programming assistant and serving users daily, thereby validating the industrial readiness of \textcolor{orange}{\texttt{ALE}}. However, existing benchmarks struggle to fully capture the complexity of such agents. Consequently, to address these limitations within the evaluation layer of the agentic ecosystem, we propose \textbf{Terminal Bench Pro}, a scaled, diverse, and high-quality benchmark designed for the rigorous assessment of agentic capabilities.

In summary, this technical report presents a reliable, cost-effective, secure, and user-friendly training ecosystem that enables practitioners to build customized models tailored to diverse needs. Beyond a technical stack, \textcolor{orange}{\texttt{ALE}} is also a call to reframe the community’s priorities. In complex agentic settings, the central challenge is no longer merely data scale or curation quality, but the \textit{co-design of training infrastructure, executable environments, and evaluation protocols}. We hope this work catalyzes collaborative efforts toward agentic benchmarks, standardized execution environments, and reproducible training pipelines, which constitute essential pillars for the next generation of general-purpose agents.

\section{Agentic Learning Ecosystem: \textbf{\textcolor{orange!80!black}{ROME}} Wasn't Built in a Day}
\subsection{System Overview}
\begin{figure}[tb]
    \centering      
    \begin{subfigure}{0.6\textwidth}
        \centering
        \includegraphics[width=\linewidth]{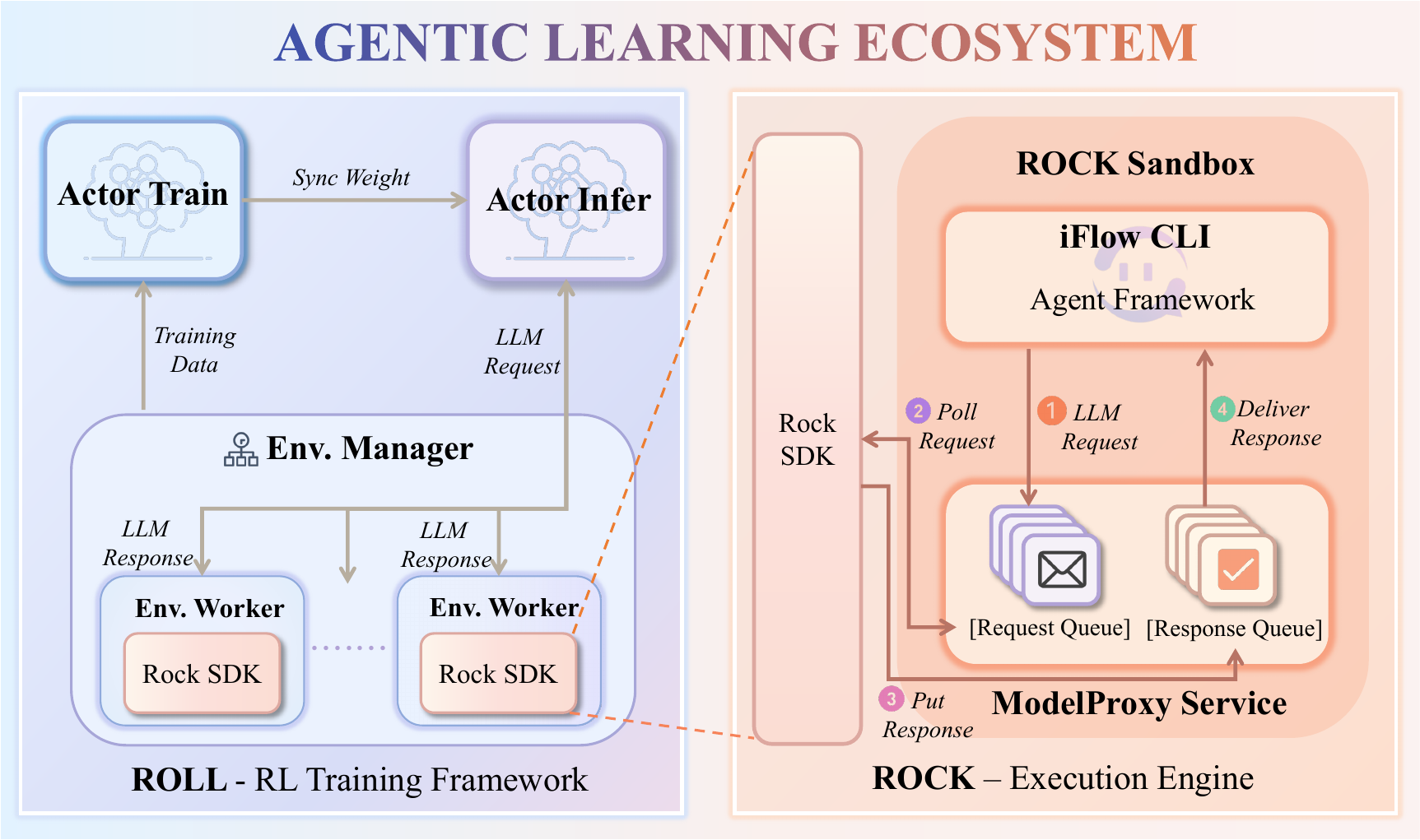}
        \caption{The overview of \textbf{\textcolor{orange}{A}gentic \textcolor{orange}{L}earning \textcolor{orange}{E}cosystem} (\texttt{\textcolor{orange}{ALE}}). }
        \label{subfig:infra-overview}
    \end{subfigure}
    \quad
    \begin{subfigure}{0.3\textwidth}
        \centering
        \includegraphics[width=\linewidth]{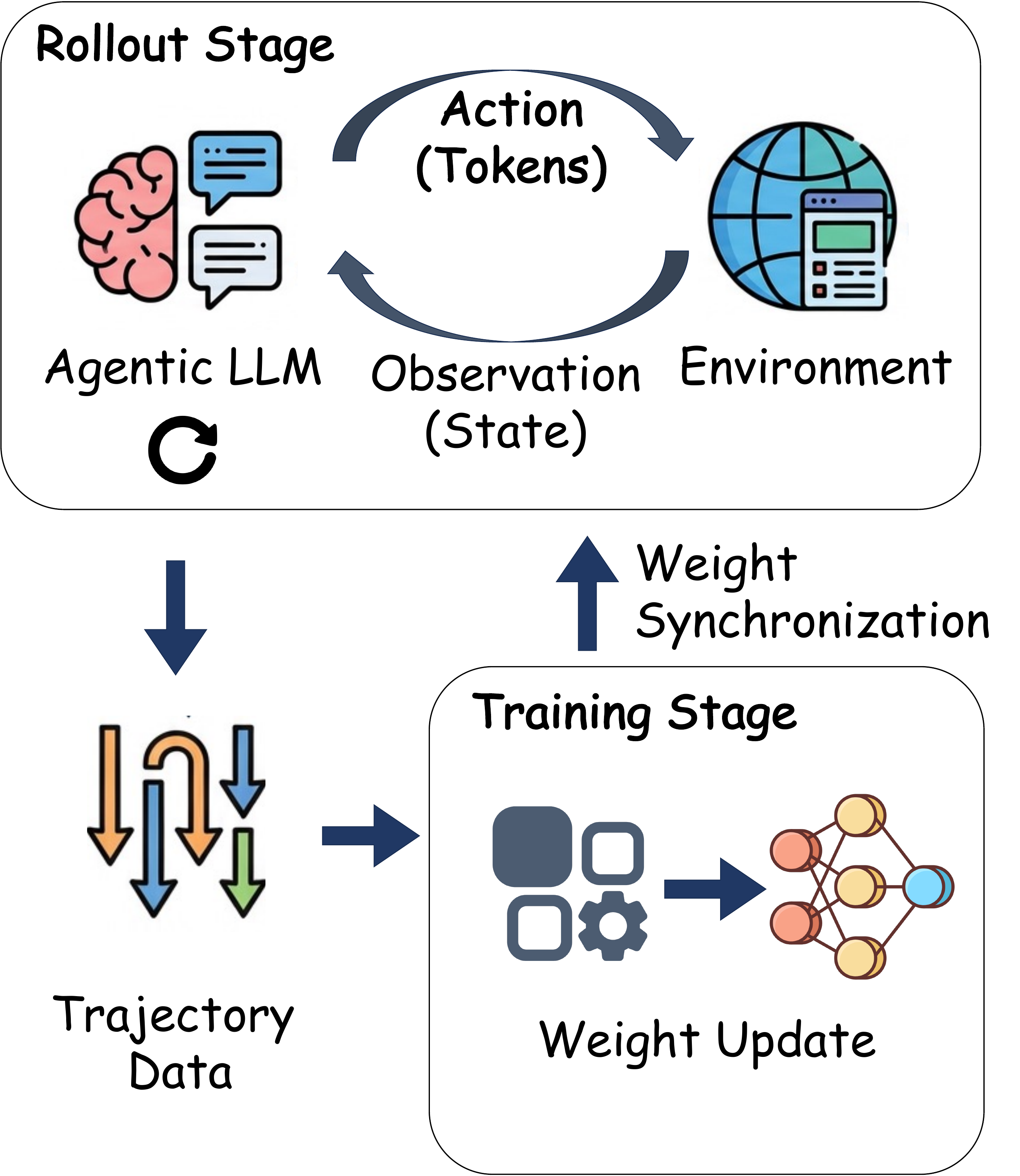}
        \caption{Agentic RL training pipeline.}
        \label{subfig:agentic-pipeline-overview}
    \end{subfigure}    
    \caption{
    The overview of agentic RL ecosystem (a) and its training pipeline (b). 
    }
    % \vspace{-20pt}
    \label{fig:agentic_pipeline}
\end{figure}

\autoref{subfig:infra-overview} shows the \textbf{\textcolor{orange}{A}gentic \textcolor{orange}{L}earning \textcolor{orange}{E}cosystem} (\texttt{\textcolor{orange}{ALE}}) that enables agentic crafting, including the training framework \emph{ROLL}, the environment execution engine \emph{ROCK}, and the agent framework \emph{iFlow CLI}. Below, we briefly describe these three systems. 

\begin{itemize}
    \item \textbf{ROLL}~\citep{roll,rollflash} is the agentic RL training framework that supports scalable and efficient RL post-training with multiple environments, multi-turn sampling, and policy optimization. 

   \item \textbf{ROCK} is the environment execution engine that provides secure, sandboxed environments for agentic interaction. It supports environment-driven trajectory generation and validation for data synthesis and closed-loop execution during training.

    \item \textbf{iFlow CLI} is the agent framework that manages the context for environment interactions and delivers an end-to-end agentic crafting experience to complete a given workflow.
\end{itemize}

The three systems work together to efficiently support agentic RL training: \textbf{ROLL} issues multiple environment calls, \textbf{ROCK} manages and executes these environments within their corresponding sandboxes, and \textbf{iFlow CLI} orchestrates the context between LLM responses and environment outputs. Together, they form an efficient, fault-tolerant, and scalable infrastructure for agentic crafting.

\subsection{Agentic RL Training Framework: \textit{ROLL}}
\paragraph{Agentic Training Pipeline.} 
% \autoref{fig:agentic_pipeline} 
\autoref{subfig:agentic-pipeline-overview}
depicts an agentic RL training workflow with three key stages, \textit{rollout}, \textit{reward}, and \textit{training}. During rollout, the agent LLM interacts with the environment by emitting tokens that represent actions. After each action, the environment returns an observation. This exchange continues for multiple turns until an episode ends, producing a trajectory of interleaved actions and observations. The reward stage then scores each trajectory and outputs a scalar reward. Finally, the training stage uses the collected trajectories and rewards to update the agent’s weights. The updated model is periodically synchronized back to the rollout stage for the next training iteration.

ROLL decomposes agentic RL post-training into specialized worker roles, including LLM inference, environment interaction, reward computation, and parameter updates. This separation allows each stage to scale independently and enables efficient communication among roles during distributed execution. Similar to prior frameworks~\citep{verl,hu2024openrlhf}, ROLL~\citep{roll,rollflash} exposes a \texttt{Cluster} abstraction and adopts a single-controller programming model. The controller coordinates heterogeneous workers and handles corresponding deployment and lifecycle management, which substantially reduces development complexity for RL researchers.

Empirical results from prior work show that rollout is the dominant cost in RL post-training and often contributes roughly 70\% of end-to-end overhead~\citep{rhymerl,rollpacker}. The problem is more pronounced in agentic training, where the rollout stage may last hundreds of seconds~\citep{rollflash}. Even the environment interaction can become a major bottleneck and has been reported to consume more than 15\% of total training time~\citep{rollart}. These observations drive the dedicated optimization for environment execution and LLM generation. In this section, we first explain how ROLL enables fine-grained rollout so that LLM generation can proceed concurrently with environment interaction within the rollout stage. We then describe ROLL’s asynchronous training pipeline that overlaps rollout with training to reduce training time while preserve the model accuracy. Last, we discuss how train-rollout multiplexing can reduce resource bubbles and improve rollout throughput in asynchronous training.

\begin{figure}[tb]
    \centering
    % 统一高度；宽度用 \linewidth 控制“不超过各自子图宽度”，从而整体正好填满一行
    \newlength{\subfigH}
    \setlength{\subfigH}{4.2cm} % 可微调：增大更“满”，减小留白更多

    \begin{subfigure}[t]{0.53\linewidth}
        \centering
        \includegraphics[height=\subfigH,width=\linewidth,keepaspectratio]{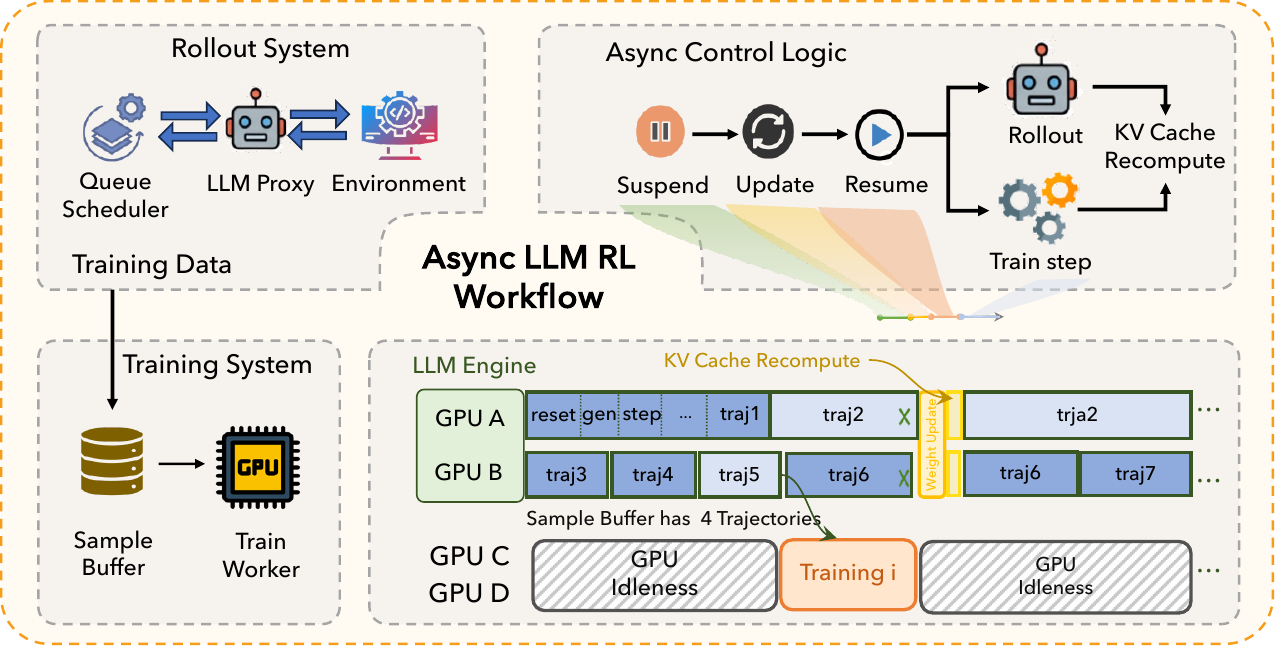}
        \caption{Fine-grained Rollout and Asynchronous Training}
        \label{fig:async_roll}
    \end{subfigure}\hfill
    \begin{subfigure}[t]{0.47\linewidth}
        \centering
        \includegraphics[height=\subfigH,width=\linewidth,keepaspectratio]{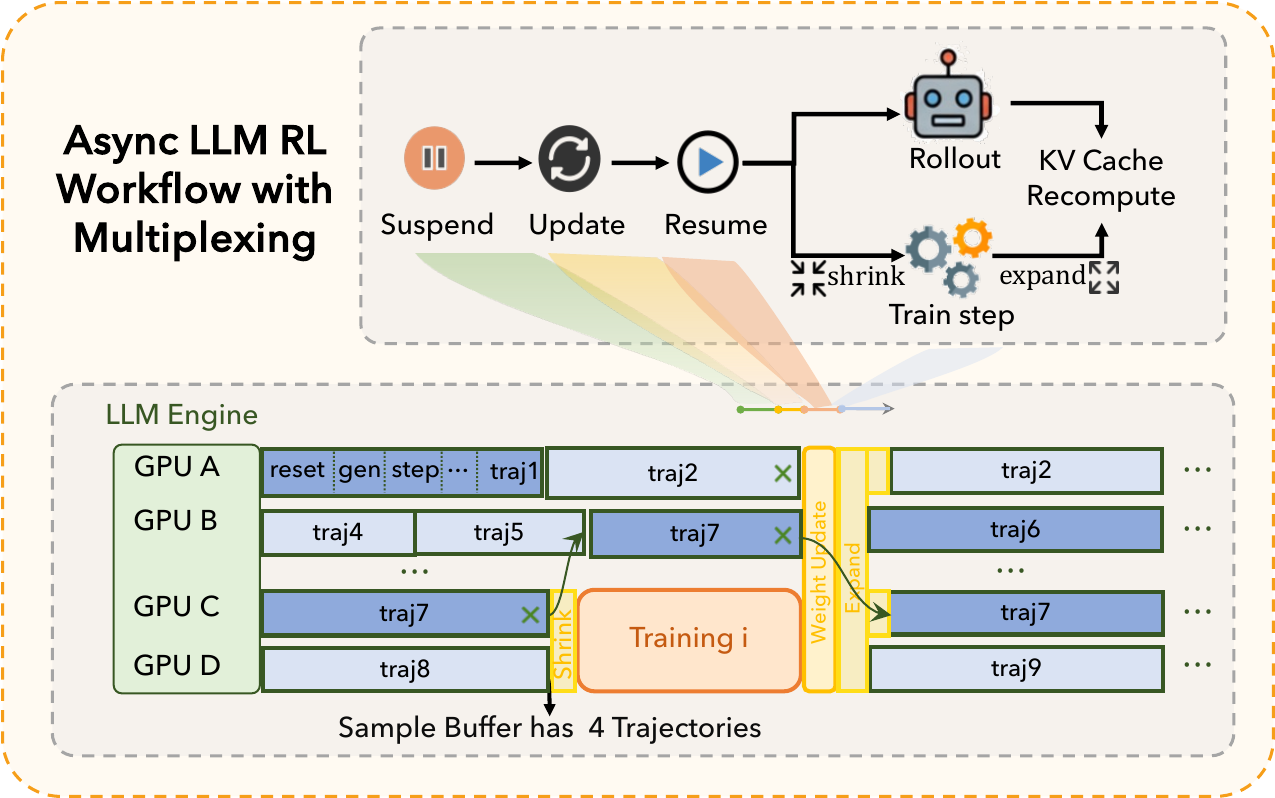}
        \caption{Train-Rollout Multiplexing}
        \label{fig:multiplex_roll}
    \end{subfigure}

    \caption{ROLL Architecture. (a) ROLL pipelines LLM generation, environment interaction, and reward phases at trajectory-level granularity. Training is also decoupled via a sample buffer using an asynchronous ratio to manage staleness. (b) ROLL multiplexes a dynamic GPU pool by shrinking rollout resources for bursty training and expanding them back during demand peaks.}
    \vspace{-5pt}
    \label{fig:async_multiplex_roll}
\end{figure}

\paragraph{Fine-grained Rollout.} ROLL supports asynchronous reward computation during rollout, thus it enables fine-grained rollout by decomposing the rollout stage into three phases: LLM generation, environment interaction, and reward computation. Instead of executing these phases in a single full batch, it applies parallelism at the sample level. This design allows users to control the lifecycle of each sample, deciding when and where each phase is executed. As a result, ROLL supports pipelined execution of LLM generation, environment interaction, and reward computation at sample-level granularity.

\paragraph{Asynchronous Training.} As shown in \autoref{fig:async_roll}, we decouple the rollout and training stage across different devices. The rollout stage acts as the producer, and the training stage acts as the consumer. ROLL maintains a sample buffer to store the completed trajectories and introduces asynchronous ratio to configure the per-sample staleness during the asynchronous training. The asynchronous ratio is defined on per sample as the maximum allowable gap in policy version numbers between the current policy and the policy version that initiated generation of that sample.

The asynchronous training pipeline iteratively repeats the following steps. First, the training stage finishes gradient computation from the previous iteration and then fetches a target batch of trajectories from the sample buffer in a blocking manner. Samples that violate the asynchronous ratio constraint are discarded to preserve model accuracy. Second, the rollout stage is suspended and model weights are synchronized from the training workers to the rollout workers. Third, the rollout stage resumes and generates new trajectories using the updated model weights, while the training stage performs gradient computation on the fetched samples in parallel to maximize resource utilization. Our prior work, ROLL-Flash~\citep{rollflash}, conduct extensive empirical studies to show that ROLL’s asynchronous training can effectively balance training accuracy and throughput. We refer interested readers to that work for details.

% 异步RL的过程，随着训练的进行，执行的关键路径会发生变化：
% Rollout是关键路径：一次训练需要的样本数量满足前，任务的关键路径是Rollout，这时候应该给Rollout提供更多的资源
% Train是关键路径：Rollout产生的样本已经满足一次训练需要，任务的关键路径是Train，这时候应该优先满足训练的资源需求，适当减少Rollout的资源

\paragraph{Train--Rollout Multiplexing.}
Although an asynchronous training architecture can overlap training and rollout via pipelining, bubbles are inevitable due to imbalanced stages. The rollout stage typically takes longer than training, the trainer may stall while waiting for enough trajectories to be collected in the sample buffer. Unlike classic pipelining with fixed resource allocation, GPUs can be dynamically reassigned between stages based on the current critical path. When rollout becomes the bottleneck, allocating more GPUs to rollout accelerates trajectory collection. Conversely, when training is the bottleneck, resources should be prioritized for training.

\autoref{fig:async_multiplex_roll} illustrates the bubble problem when rollout stage dominates the end-to-end iteration time. Rollout typically exhibits a pronounced long-tail latency distribution: the staleness bound caps the number of in-flight trajectories, and while most trajectories finish quickly, a small fraction of stragglers run up to the maximum context length, leaving many rollout GPUs underutilized. Meanwhile, the training stage is comparatively short but must wait until rollout has produced enough valid samples. Under a static GPU partition between rollout and training, this mismatch creates resource bubbles.

Our key insight is that rollout demand is highly time-varying: it peaks immediately after weight synchronization, when many new trajectories are launched, and then drops into a low-demand valley where only a small set of stragglers remain. In contrast, the training stage consumes resources in short, bursty episodes. Building on this observation, we introduce time-division multiplexing with a dynamic GPU partition between rollout and training. As shown in \autoref{fig:multiplex_roll}, the system first assigns all GPUs to rollout to rapidly generate a batch of samples. Once the sample buffer accumulates sufficient data for the next training step, the system triggers a \texttt{shrink} operation that temporarily reallocates a fixed subset of GPUs to training, while consolidating the remaining unfinished trajectories onto the rollout GPUs that remain. After training completes, an \texttt{expand} operation returns those GPUs to rollout to serve the next demand peak. This policy aligns training bursts with rollout demand valleys, reducing bubbles and improving overall GPU utilization compared to a statically disaggregated asynchronous design.

\begin{figure}[tb!]
    \centering
        \centering
        \includegraphics[width=0.98\linewidth]{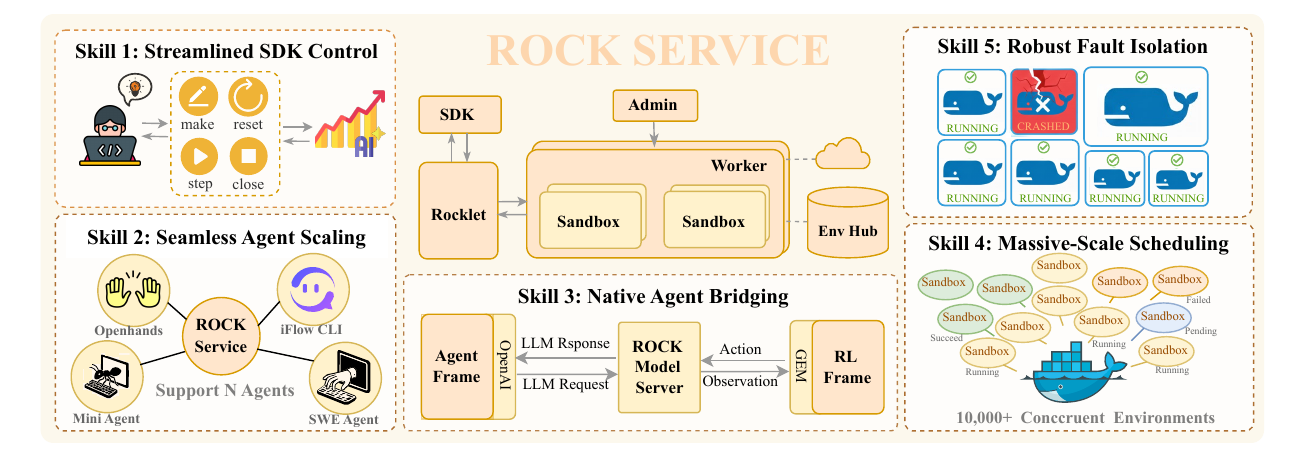}
        % \vspace{-10pt}
        \caption{ROCK System Architecture.}
        \label{fig:rock_arch}
\end{figure}

\subsection{Environment Execution Engine: \textit{ROCK}}
ROCK is a scalable and user-friendly system for managing sandbox environments to complete various agentic crafting applications (e.g., travel plan, GUI assistant). It is designed to be framework-agnostic, providing flexible APIs that allow any RL training frameworks to programmatically build, manage, and schedule these environments.

% \paragraph{System Overview and Workflow.} \autoref{fig:rock_arch} illustrates the architecture of ROCK. ROCK comprises three key components: (1) an \emph{Admin} control plane that provisions environments as sandboxes and performs global resource scheduling and allocation; (2) \emph{Worker} nodes that supply physical resources and host the sandbox runtime; and (3) \emph{Rocklet}, a lightweight proxy that relays actions between the agent SDK and sandboxes, and, when needed, enables sandboxes to access external Internet services.

\paragraph{System Architecture and Workflow.} \autoref{fig:rock_arch} illustrates the architecture of ROCK. The ROCK system is designed around a client-server architecture to support multiple levels of isolation, guaranteeing operational stability. From the client perspective, interacting with a remote environment is as convenient as using a local RL environment through a small set of primitives such as \texttt{reset}, \texttt{step}, and \texttt{close}. Under the hood, ROCK decouples environment execution from orchestration so that large-scale concurrent rollouts remain stable, debuggable, and resource efficient.

ROCK consists of three main components. First, the server tier is governed by the \emph{Admin} control plane, which serves as the orchestration engine: it provisions sandboxed environments, performs admission control, and manages cluster-wide resource scheduling and allocation. Second, the worker tier comprises \emph{Worker} nodes deployed on each machine; they run the sandbox runtime and manage local hardware resources. Third, \emph{Rocklet} is a lightweight proxy that mediates communication between the agent SDK and sandboxes, governs outbound network access, and enforces egress policies. In addition, ROCK provides \emph{EnvHub}, a centralized registry for environment images that enables reproducible provisioning and faster cold starts.

The agent LLM training, evaluation, and data synthesis impose diverse requirements, and ROCK provides the following features to meet these needs.

% During agent LLM training, evaluation, and data synthesis, ROCK receives large volumes of requests to execute untrusted instructions, produce and transfer artifacts, and meet strict requirements for permission control. To satisfy diverse needs of the agentic workloads, ROCK provides the following features.

\begin{itemize}
  \item \textbf{Skill 1: Streamlined SDK Control.} ROCK exposes a minimal, consistent control interface aligned with standard GEM RL environment semantics. Users can create, reset, step, and close environments through a small set of APIs, simplifying integration with RL training and evaluation pipelines. We detail these APIs later.

  \item \textbf{Skill 2: Seamless Agent Scaling.} ROCK supports environments with multiple agents and can provision shared or isolated sandboxes based on the interaction pattern, enabling multi-agent collaboration and competition. It also orchestrates diverse agent benchmarks (e.g., SWE-bench~\citep{swe_verified}, Terminal Bench Pro~\citep{tbench_2025}) behind a unified GEM API, so ROLL can interact heterogeneous environments through a single interface and enable multi-task RL training with only minimal configuration changes.
  
  \item \textbf{Skill 3: Native Agent Bridging.} This bridges the gap between the RL framework and the agent framework that reconstructs and aligns the agent’s native message-based context management. We explain this native agent mode in detail later. 
    
    \item \textbf{Skill 4: Massive-Scale Scheduling.} ROCK performs dynamic allocation and reclamation of resources across sandboxes. This enables high utilization under bursty workloads and supports large-scale concurrency, scaling to tens of thousands of simultaneous environments by elastically distributing tasks over the cluster.
   
    \item \textbf{Skill 5: Robust Fault isolation.} Each task runs in its own sandbox. If an agent crashes, gets stuck, or damages its files, the failure is contained within that sandbox and does not interfere with other tasks on the same machine. ROCK also restricts each sandbox’s network access with per-sandbox policies, limiting the impact of misbehaving or compromised agents.

  \item \textbf{Tailored Optimizations.} ROCK provides permission isolation for untrusted instructions, efficient large-file and artifact transfer, centralized logging, resource guardrails with failure recovery, optional checkpointing and restart support, and tooling for debugging and CI/CD-style environment delivery. 
\end{itemize}

\paragraph{API Interfaces.} ROCK exposes two primary API services for programmatic control, namely the Sandbox API and the GEM API. The Sandbox API manages the sandboxes that host GEM environments. The GEM API provided by ROCK follows the official GEM standardized API~\citep{GEM-github}. It is training-framework agnostic and integrates seamlessly with a range of RL frameworks, including veRL~\citep{verl}, OpenRLHF~\citep{hu2024openrlhf}, and Tinker~\citep{thinkingmachines_tinker}. To ensure broad compatibility, ROLL also provides a GEM API implementation that adheres to the GEM protocol~\citep{GEM-github}. In particular, environment workers managed by the ROLL runtime use the GEM API to mediate interactions between an agent and its environment hosted by ROCK. All endpoints follow a RESTful design and use JSON for data interchange. We describe both APIs below.

The sandbox API manages the complete lifecycle of sandbox instances. Its functionality can be grouped into three main categories:
\begin{itemize}
    \item \textbf{Provisioning:} Create and start sandboxes, with support for custom images, resource configurations, and both synchronous and asynchronous modes.
    \item \textbf{Monitoring:} Query the status, operational health, and resource consumption statistics of any running sandbox.
    \item \textbf{Persistence:} Stop a sandbox instance to release its resources or commit its current state to a new image for future use.
\end{itemize}

As a standardized interface for RL environments, this protocol enables the API to support the core agent interaction loop for general-purpose tasks:
\begin{itemize}
    \item \textbf{Make:} Create a new GEM environment instance.
    \item \textbf{Reset:} Reset an existing environment instance to its default state.
    \item \textbf{Step:} Send an action to advance the environment one step and receive the next state.
    \item \textbf{Close:} Close the environment to release resources.
\end{itemize}

\paragraph{Agent Native Mode.} The agent native mode connects the agentic RL training with the ROCK. The inconsistency in context management between the training framework (ROLL) and the deployment system (iFlow CLI) can significantly degrade an agent's performance in production~\citep{rush2025building}. A naive solution would be to force ROLL to perfectly mirror the iFlow CLI's context handling, including its specific logic for multi-turn interactions and prompt concatenation. However, this creates a tight coupling: every update to an agent's logic would require a corresponding reimplementation within ROLL, leading to an unsustainable maintenance burden.

To address this, we have implemented a \texttt{ModelProxyService} within the ROCK environment. This service acts as a proxy, intercepting all LLM requests originating from the agent's sandbox. Crucially, these requests already contain the complete historical context, fully orchestrated by the iFlow CLI. The proxy then forwards these requests to the appropriate inference service — be it ROLL inference workers during training or an external API (e.g., GPT, Gemini) during deployment. The native mode achieves a clean separation. ROLL is simplified to generation engine, while the iFlow CLI retains full control over context management. This not only eliminates implementation complexity in the training framework but also guarantees perfect consistency between training and deployment, resolving both the maintenance and performance issues. The agent native mode ensures consistency not just between training and deployment, but across the full development pipeline, including data synthesis, training, and evaluation. A key feature is its support for multiple agent frameworks (iFlow CLI, SWE-Agent~\citep{deepswe2025}, OpenHands~\citep{wang2025openhandsopenplatformai}, etc.), which lowers the overhead of switching scaffolds and simplifies tasks like generating more diverse training data.

\begin{figure}[tb]
    \centering      
    % \begin{subfigure}{0.98\textwidth}
        \begin{subfigure}{0.95\textwidth}
        \centering
                \includegraphics[width=\linewidth]{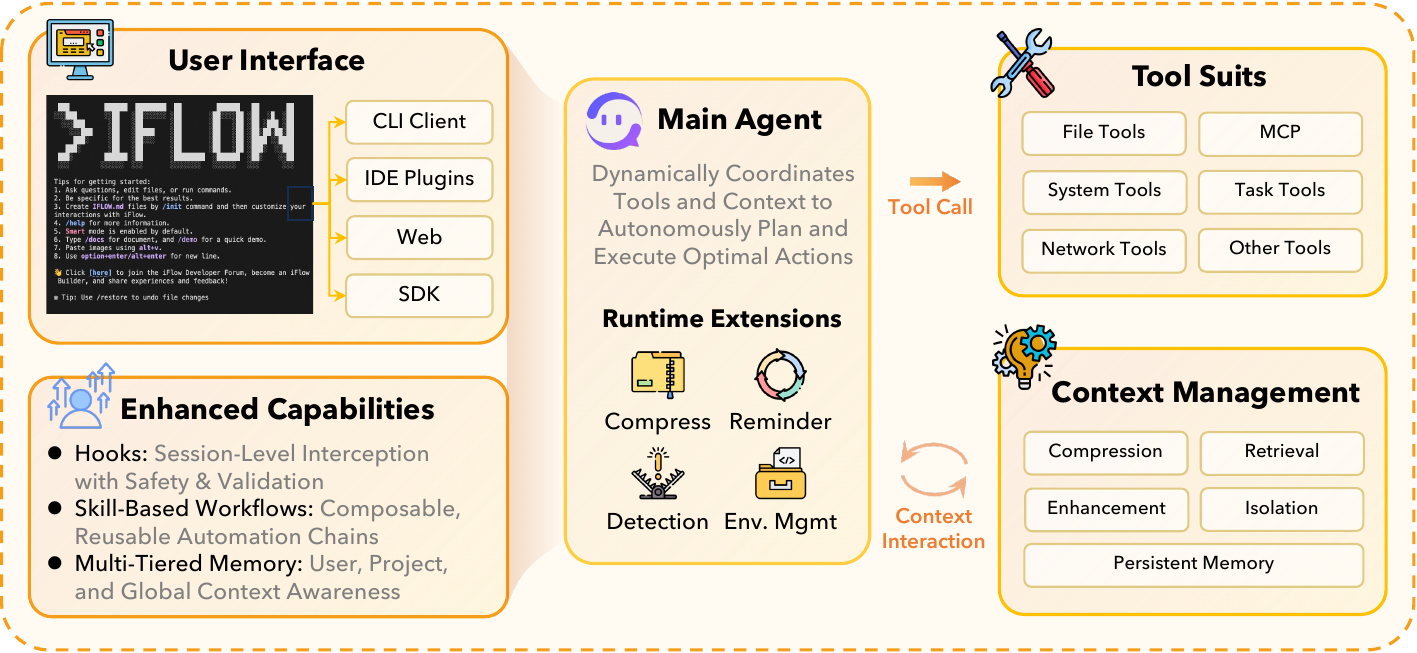}
    \end{subfigure}    
    \caption{
        The overview of iFlow CLI architecture and execution.}
    % \vspace{-20pt}
    \label{fig:iflow-cli}
\end{figure}

% 缺结构化的描述、特点突出、以及与模型训练之间的关系.

% 整体是single-Agent结构。
% 整体增强的能力：
% hooks: 会话级别，工具调用前后。e.g. "rm"类动作会提醒拦截；edit file，前后有语法检查反馈；https://platform.iflow.cn/cli/examples/hooks
% workflow：like claude code的skills;
% 分级memory：user-level; project-level; 全局-level
% 其他：用户定义subagent; 支持mcp;

% 在交互中额外有四个能力增强：
% 1. Env.Mgmt(environment management):获取环境信息； 用户环境发生变更会提醒；
% 2. compress: 9级压缩，更精确的上下文；
% 3. reminder: 环境变化、工具变化、任务是否完成，都会告知模型；
% 4. detection: 检测(?)。

\subsection{Agent Framework: \textit{iFlow CLI}}
The iFlow CLI is a powerful command-line agent framework that exposes an interface for automating and executing complex, multi-step tasks, serving as both the context manager and user interface for our infrastructure layer. We describe the role of iFlow CLI in agentic RL training, and provide its overview, and highlight two key features, namely context engineering and open configuration.

\paragraph{The Role of iFlow CLI in Agentic Training.} iFlow CLI bears two roles in agentic RL training. First, in agent-native mode, a model-proxy service intercepts requests from ROLL and invokes iFlow CLI for context management, ensuring consistency between training and deployment. Second, iFlow CLI’s open configuration enables general-purpose LLMs to incorporate domain-specific knowledge during training via context management. By allowing configurable system prompts, tools, and workflows, iFlow CLI becomes a flexible substrate for training and refining agent behavior, improving performance on domain-specific agentic tasks.

\paragraph{System Architecture and Workflow.} As shown in \autoref{fig:iflow-cli}, iFlow CLI adopts an orchestrator-worker architecture built around a single-agent design principle, following Anthropic’s recommendations for effective agentic systems~\citep{albert2024building}. The system exposes various \emph{user interfaces} to users including client, IDE plugins, web and SDK. The system is driven by a \emph{Main Agent} that maintains the global task state and executes an iterative control loop. At each step, iFlow CLI receives the user command and loads available persistent memory and prior chat history, then perform \emph{context management} to assemble the model input. Based on the context, the Main Agent selects the next action, which may be a direct response, a tool invocation, or a call to a specialized sub-agent. The \emph{tool suites} are accessed through a unified aggregation layer that wraps heterogeneous capabilities, such as MCP integrations, and returns their results as observations the agent can consume. Importantly, sub-agents are implemented as specialized tools with bounded context, avoiding agent handoffs and removing the need for explicit inter-agent communication.

During the control loop, iFlow CLI provides four built-in skills to strengthen context management. The \emph{Compress} performs context compression for limited prompt budgets. The \emph{Reminder} reports context changes including environment updates, tool changes, and task done. The \emph{Detection} identifies issues such as loops and tool-call failures. The \emph{Env.Mgmt} tracks environment state and notifies the agent upon user environment changes. The iFlow CLI also provides three enhanced capabilities. The \emph{Hooks} implement session-level pre- and post-tool checks, such as warnings and interception for destructive commands. The \emph{Workflow} packages reusable skills as configurable procedures for multi-step tasks. The \emph{Memory} maintains hierarchical persistent state at the user, project, and global levels.

% In parallel, a retrieval pipeline queries a knowledge base via semantic search over a vector database and injects relevant references into the context when needed. Based on this aggregated context, the Main Agent selects the next action, which may be a direct response, a tool invocation, or a call to a specialized sub-agent. Tools are accessed through a unified aggregation layer that wraps heterogeneous capabilities, such as MCP integrations, and returns their results as observations the agent can consume. Importantly, sub-agents are implemented as specialized tools with bounded context, avoiding agent handoffs and removing the need for explicit inter-agent communication. After each action, iFlow CLI writes tool responses and sub-agent outputs back to chat history and persistent memory, optionally compressing or filtering content, and repeats the loop until the task completes.

\paragraph{Context Engineering for Agentic Crafting.} We adopt a single-agent control loop because it is simple, robust, and easy to scale. Following ``The Bitter Lesson''~\citep{sutton2019bitterlesson}, we avoid brittle, over-engineered pipelines and instead focus on \emph{context engineering}: supplying the agent with precise, high-quality context so it can plan, act, and self-correct effectively in real software environments.

In practice, iFlow CLI implements five techniques to manage context for long-horizon tasks:
\begin{itemize}
    \item \textbf{Persistent memory.} iFlow maintains a lightweight \texttt{todo} file as external memory across sessions. The agent can read and update it to track plans, open issues, and next steps.
    \item \textbf{Context isolation.}  For complex tasks, iFlow can delegate sub-tasks to a \texttt{sub-agent}. Each sub-agent operates within a dedicated, isolated context, which prevents interference with the main agent's workflow and ensures more focused, efficient execution.
    \item \textbf{Context retrieval.} iFlow fetches relevant information on demand via agent search, semantic vector retrieval, and knowledge-base integrations (e.g., DeepWiki), reducing reliance on what is already in the prompt.
    \item \textbf{Context compression.} To cope with limited context windows, iFlow applies lossy and lossless compression to retain key facts while controlling prompt length.

    \item \textbf{Context enhancement.} Users can explicitly highlight critical signals. This includes reinforcing the current task objective or highlighting significant changes in the environment (e.g., new files created, test results) to guide the LLM's attention.
\end{itemize}

\noindent Together, these capabilities enable a specification-driven workflow for domain tasks: by injecting clear ``specs'' (prompts, tools, and procedures) into the context, iFlow can execute specialized workflows (e.g., WeChat Mini-program development or iOS app engineering) while keeping the core agent loop unchanged. The iFlow CLI also exposes open configuration interfaces, making it straightforward to align RL training with domain-specific prompts, tools, and workflows.

\paragraph{Open Configuration Capabilities.} Real-world software engineering demands more than generic intelligence. It requires strict adherence to domain-specific standards, complex operational logic, and specialized toolchains. To bridge the gap between general-purpose models and specialized engineering requirements, the iFlow CLI exposes a highly customizable configuration layer:

\begin{itemize}
    \item \textbf{System Prompt (Behavioral Alignment)} To align the model's cognitive style with specific domain constraints, the system prompt serves as a flexible blueprint. Users can explicitly define workflows, toolsets, usage scenarios, and persona tones. This customization acts as an accurate control mechanism, optimizing the model's responses to fit the unique requirements of a specific project or field.
    
    \item \textbf{Workflow / Spec (Process Standardization):} To scale from simple code generation to end-to-end, workflow-driven tasks, iFlow CLI introduces \emph{Workflows} (or \emph{Specs}). This feature lets users compose disparate AI capabilities—agents, commands, and tools—into structured, automated task chains. Whether for code analysis, development cycles, or deployment pipelines, workflows ensure complex processes are executed reliably and autonomously.

    \item \textbf{Tool Set (Functional Extensibility):} To extend beyond the LLM's native capabilities, iFlow CLI supports broad integration via the Model Context Protocol (MCP). Users can add custom tools or sub-agents (invoked as tools within a single-agent loop), enabling seamless interaction with external APIs, databases, and proprietary environments.

\end{itemize}

\subsection{Summary}
Our infrastructure, leveraging ROLL, ROCK, and the iFlow CLI, provides system-level support for the entire agentic RL pipeline from training to deployment at the system layer. It is specifically served as the two pillars of high-performance agentic RL: structuring effective training algorithms and constructing quality datasets, as discussed subsequently.

\section{Agentic Model: {\color{orange!80!black}\underline{R}}OME is {\color{orange!80!black}\underline{O}}bviously an Agentic {\color{orange!80!black}\underline{M}}od{\color{orange!80!black}\underline{E}}l}

% \red{\textbf{Rome wasn't built in a day}, and the edifice of Artificial General Intelligence demands equally enduring foundations. Our proposed model, ROME, epitomizes this arduous yet inevitable transition from passive conversational assistants to autonomous agents capable of intricate execution. Guided by the maxim that \textbf{all roads lead to Rome}, we posit that the fragmented challenges of agentic coding can converge into a unified, scientifically coherent framework. This section delineates the three pillars underpinning our approach

This section introduces \textbf{\textcolor{orange!80!black}{ROME}}, our agentic foundation model trained with our \textcolor{orange}{\texttt{ALE}} infrastructure. \textbf{\textcolor{orange!80!black}{ROME}} excels at a wide range of workflow-driven tasks (e.g., GUI assistance, travel plan). We then outline the core principles and procedures behind its development for strong agentic crafting performance, organized into three components: (1) a rigorous and principled data acquisition and synthesis workflow; (2) an end-to-end training pipeline integrating \textbf{Agentic Continual Pre-training (CPT)}, \textbf{Supervised Fine-tuning (SFT)}, and \textbf{\textcolor{orange}{I}nteraction-\textcolor{orange}{P}erceptive \textcolor{orange}{A}gentic Policy Optimization}(\texttt{\textcolor{orange}{IPA}}) RL algorithm; and (3) a comprehensive benchmark suite. Collectively, these components form a systematic pathway that illustrates how \textbf{\textcolor{orange!80!black}{ROME}} leverages the required infrastructure to support next-generation agentic LLM.

\subsection{Data Composition}\label{sec:data_composition}

% \blue{This section details our data construction strategy for training agentic coding models. We adopt a layered, competency-driven design and implement it through two complementary data pipelines that align with two training regimes. \textbf{Basic Data} consists of large-scale, high-quality corpora for continuous pretraining, primarily organized as fully collected text-and-code samples, and is intended to establish strong foundations in code understanding and generation, tool-use literacy, and general reasoning. \textbf{Agentic Data} targets agent-specific requirements by producing closed-loop, executable training units in realistic environments. It is organized into i) instances, which extend a conventional query with an executable specification, a pinned environment, and verifiable feedback, and ii) trajectories, which record single-turn and multi-turn interactions in which agents iteratively plan, act, observe runtime signals, and revise solutions. Agentic data can be directly leveraged in post-training to selectively enhance agentic planning, execution, and adaptation under real-world constraints. Together, the two pipelines operationalize our competency analysis and provide a coherent progression from foundational skills to full agentic capabilities.}

\subsubsection{Agent Competencies as a Blueprint for Data Design}

\begin{figure}[tb]
    \centering
        \centering
                \includegraphics[width=1\linewidth]{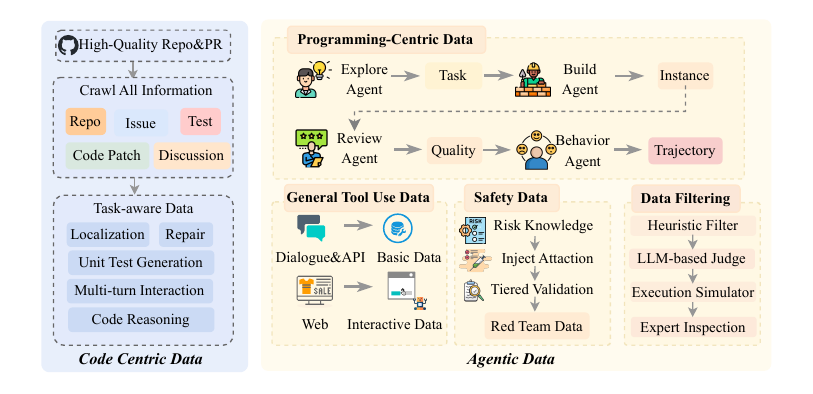}
        % \vspace{-0.4cm}
        \caption{Overview of data sources and composition pipelines for training agentic models, spanning code centric basic data and agentic data.}
        \label{fig:data_overall}
\end{figure}

Agentic crafting aims to build autonomous, workflow-driven agents that can reliably translate requirements into working artifacts through an iterative loop of \emph{formulation, implementation, verification, and refinement}. To characterize what such agents must learn and consequently what training signals our data must provide, we decompose agentic crafting competencies into three tightly coupled dimensions: \textbf{task understanding and planning}, \textbf{action and execution}, and \textbf{interaction and adaptation}:

\begin{itemize}
    \item \textbf{Task Understanding and Planning.} This dimension captures the agent's ability to interpret natural-language or semi-structured specifications and translate them into well-scoped, executable engineering tasks accompanied by verifiable development plans. The agent must accurately extract user intent, uncover implicit rules and constraints, and surface hidden assumptions that could derail implementation. This involves identifying core system entities, defining precise input-output contracts, establishing boundary conditions, and articulating non-functional requirements (e.g., performance, security, scalability, compatibility) that are often omitted but critical to real-world viability. When information is incomplete, the agent should ask minimally sufficient clarification questions and explicitly represent uncertainty, avoiding overcommitment under ambiguous requirements and thereby reducing downstream rework.
    \item \textbf{Action and Execution.} This dimension concerns the agent's ability to operationalize plans into high-quality implementations and to leverage external toolchains to close the development loop. The agent must actively select appropriate tools based on task characteristics (e.g., code search, build systems, dependency management, compilation/execution, testing frameworks, debuggers, static checkers, formatters, profilers, CI/CD pipelines) and invoke them with correct parameters and sequencing. Critically, the agent must also interpret tool outputs to drive subsequent actions, e.g., localizing defects from failing test logs, resolving style and correctness issues from linter reports, and optimizing bottlenecks guided by profiler evidence.
    \item \textbf{Interaction and Adaptation.} This dimension governs the agent's ability to maintain a dynamic feedback loop with its environment, enabling continuous refinement across iterations. The agent must actively incorporate diverse signals (e.g., runtime behavior, test outcomes, user feedback, code review comments, and evolving system constraints) and adapt its plans and implementations accordingly. For instance, when faced with API deprecations or dependency conflicts, it should perform impact analysis and pivot to alternative strategies (e.g., rollback, refactoring, or substitution) rather than rigidly adhering to an outdated plan.
\end{itemize}

    % This dimension governs whether the agent can form an effective feedback loop with the environment: incorporate signals, revise strategies, maintain context, and continuously improve output quality across iterations. Dynamic response to environmental feedback: The agent should continuously incorporate feedback from runtime behavior, test results, user updates, code review comments, and external system constraints, and adjust plans and implementations accordingly. For example, when API constraints change or dependency conflicts arise, it should promptly conduct impact analysis and apply rollback or alternative implementations rather than mechanically following the original plan.

% deliberately maps supervision signals to the three dimensions and
% \textbf{Basic Data} consists of large-scale, high-quality corpora for continuous pretraining, primarily organized as fully collected text-and-code samples. It is designed to establish strong foundations in code understanding and generation, tool-use literacy, and general reasoning, as the capabilities that underwrite reliable task interpretation and disciplined execution in static or semi-static contexts.

Guided by the above competency analysis, our data design adopts a two-tier curriculum that stages the model from foundational proficiency to closed-loop agentic behavior. In the first tier, \textbf{Basic Data} delivers targeted basic capability building that agentic models require as they progress toward full agent behavior. It comprises complementary components including \emph{code-centric corpora} that support continuous pretraining and strengthen project-level code understanding and generation, and \emph{general reasoning data} spanning reasoning-intensive tasks and general-purpose instructions that reinforces transferable deduction and planning skills. In the second tier, \textbf{Agentic Data} targets agent-specific requirements by producing closed-loop, executable training units in realistic environments. It is organized into i) instances, which extend a conventional query with an executable specification, a pinned environment, and verifiable feedback, and ii) trajectories, which record multi-turn interactions in which agents iteratively plan, act, observe runtime feedback, and revise solutions. Agentic data can be directly leveraged in post-training to selectively enhance agentic planning, execution, and adaptation under real-world constraints.

Our data maps the competency dimensions to supervision across both tiers. Basic Data concentrates on task understanding and planning by exposing the model to rich project contexts and well-formed specifications that teach intent extraction, requirement scoping, and plan formulation. It also builds the coding and general reasoning foundations that later enable effective action, execution, and iterative refinement, without relying on explicit tool-use traces. Agentic Data then provides targeted strengthening of action and execution and of interaction and adaptation. It embeds requirements in pinned, executable environments, supplies verifiable runtime feedback through deterministic builds and tests, and captures single- and multi-turn trajectories in live settings. This setting both trains robust execution and adaptation and grounds task understanding and planning in realistic constraints, turning high-level plans into working solutions under real-world conditions.

Together, the two tiers of data form a staged curriculum. The basic Data builds breadth and reliability in core coding and reasoning without full environment orchestration, while the agentic data then adds closed-loop execution and concrete runtime signals that directly supervise planning discipline, execution fidelity, and adaptive iteration under real world constraints. This progression operationalizes the competency blueprint and provides a coherent path from foundational skills to full agentic capabilities.

% write by: xuanwu, yupei
\subsubsection{Code-Centric Basic Data Composition}
\label{sec:basic_data}
% This subsection provides a detailed description of the collection and construction pipeline for \textbf{basic data}.
 % tool-use data

% Coding
% \paragraph{\textcolor{orange!80!black}{Code-Centric Data Construction.}}
% To systematically improve the model’s ability to understand and generate code for complete and complex projects, we collect and construct large amount of code data. we build on real-world software engineering environment and construct a diverse training dataset and task suite covering code comprehension, fault localization, bug fixing, and test generation.

As a cornerstone of agentic LLM capabilities, coding proficiency requires a robust foundation of large-scale, high-quality code data. Building such a corpus entails not only the systematic acquisition of extensive codebases but also the establishment of specialized environments to synthesize and process real-world software engineering data. Consequently, we curate a comprehensive dataset and task suite leveraging authentic development ecosystems to cover critical dimensions including code comprehension, fault localization, bug remediation, and automated test generation, etc.

% (1) \textbf{Data Acquisition & Preprocessing} To enhance the model’s ability to model full project structures and cross-file dependencies, we select approximately one million high-quality GitHub repositories based on criteria such as star counts, fork statistics, and contributor activity. Following Seed-Coder~\citep{seed2025seedcoderletcodemodel}, we concatenate multiple source files within the same repository to form training samples at the project-level code structure, preventing the model from learning only isolated code snippets and promoting understanding of real-world engineering context. 

% \textbf{Issue–PR–driven software engineering data collection:} To improve code localization and repair, we further crawl \textit{Issues} and \textit{Pull Requests} from the selected repositories. We retain only closed Issues and merge PRs to ensure a clear problem–solution correspondence. We then use an LLM to filter Issues, removing low-quality cases with vague descriptions, pure questions/discussions, auto-generated content, or missing key technical details. During Issue–PR linking, we keep only PRs with explicit will-close intent that actually resolved the corresponding Issue, excluding PRs that merely referenced the Issue without substantive fixes.

\textbf{\textcolor{orange!80!black}{Data Acquisition \& Preprocessing.}} We select approximately one million high-quality GitHub repositories based on criteria such as star counts, fork statistics, and contributor activity. Following Seed-Coder~\citep{seed2025seedcoderletcodemodel}, we concatenate multiple source files within the same repository to form training samples at the project-level code structure, preventing the model from learning only isolated code snippets and promoting understanding of real-world engineering context. In addition, to improve code localization and repair, we further crawl \textit{Issues} and \textit{Pull Requests (PRs)} from the selected repositories. We retain only closed Issues and merged PRs to ensure a clear problem–solution correspondence. We then use an LLM to filter Issues, removing low-quality cases with vague descriptions, purely question/discussion posts, auto-generated content, or missing key technical details. During Issue–PR linking, we retain only PRs with an explicit will-close intent that actually resolve the corresponding Issue, excluding PRs that merely referenced the Issue without substantive fixes.

\paragraph{\textcolor{orange!80!black}{Task Construction and Formalization.}}
Building upon the collected Issue-PR pairs, we formulate five core categories of software engineering tasks:
% following the paradigms established in AGENTLESS~\citep{xia2024agentless} as follows:

% Following the paradigms established in AGENTLESS~\citep{xia2024agentless}, we define them as follows:

\begin{itemize}
    \item \textbf{Code Localization.} To establish a target for modification, we follow the protocol in AGENTLESS~\citep{xia2024agentless} by adopting the modified-file list from the golden patch as the ground-truth. Formally, given an issue description $I$ and the repository structure $S$, the task is to identify a minimal subset of files $F = \{ f_1, f_2, \ldots, f_n \} \subset S$ that require editing to resolve the issue.

    \item \textbf{Code Repair.} Building on the localized files, we formulate the repair process as a structured transformation. Following AGENTLESS~\citep{xia2024agentless}, golden-patch differences are converted into \textit{search-and-replace} blocks to provide precise editing signals. Formally, given issue $I$ and the relevant code segments $C$, the model $\mathcal{M}$ generates a set of edits $R = \mathcal{M}(I, C)$, where $R$ represents the search-and-replace blocks specifying the required transformation.
    
    \item \textbf{Unit Test Generation.} To achieve closed-loop verification of the proposed repairs, we formulate a test generation task by extracting test-centric patches from the associated PRs. Formally, given the issue $I$ and the successfully patched code $C'$, the model synthesizes a corresponding test suite $T = \mathcal{M}(I, C')$ specifically designed to validate the correctness of the repairs.

    \item \textbf{Multi-turn Interaction.} To enhance the model's capability in multi-turn tasks, we carefully construct a high-quality multi-turn interaction dataset. Following the methodology of SWE-RL~\citep{wei2025swe}, we treat PR comments as turn-level feedback signals ($\mathrm{feedback}_t$) and the subsequent commit-level code changes as the corresponding responses ($\mathrm{response}_t$). This allows for formalizing the iterative refinement process as an evolutionary \textit{feedback-edit} trajectory: $(\mathrm{feedback}_1, \mathrm{response}_1) \rightarrow \dots \rightarrow (\mathrm{feedback}_n, \mathrm{response}_n)$.

    \item \textbf{Code Reasoning.} To further bolster the model's underlying reasoning capabilities, we utilize larger and more capable models to synthesize intermediate CoT rationales for file localization, code repair, and unit test generation, ensuring that the model internalizes the analytical logic behind each modification. To guarantee high data fidelity, we implement a rigorous rejection sampling pipeline: localization samples are retained only they fully cover the ground-truth set of modified files, while repair and test generation samples are filtered based on a sequence-level similarity threshold relative to the golden patches.
\end{itemize}

% YANAN CHECK
Employing the aforementioned data collection and task-synthesis procedures, we construct an initial corpus exceeding 200B tokens. Through stringent data hygiene and quality assurance protocols (e.g., deduplication, decontamination, noise reduction, and logical consistency verification), we distill this corpus into a high-qualiy dataset comprising 100B tokens, which serves as the foundation for both continuous pre-training and post-training stages.

\subsubsection{Agentic Data Composition}

Agentic data differs fundamentally from conventional code corpora. Instead of isolated snippets or static repositories, it packages tasks with an executable specification, a pinned environment, and verifiable feedback, and it records how agents behave when they plan, act, observe runtime signals, and revise solutions. This closed-loop structure is essential for training models to exhibit reliable agentic behavior, yet it introduces challenges that conventional datasets do not address: environment reproducibility, execution closure, high-quality feedback signals, and resistance to superficial solutions.

Two core data objects define the agentic data form:
\begin{itemize}
    \item \textbf{Instance.} An instance is the agentic analogue of a \emph{query} in basic instruction data. It bundles the prompt (task specification), a Dockerfile together with build/test commands that pin the execution environment, and unit tests that provide verifiable feedback. This packaging turns an abstract problem into a runnable, reproducible task with clear acceptance criteria.

    \item \textbf{Trajectory.} A trajectory records an agent's behavior on a validated instance. It captures multi-turn interactions, including tool invocations, file edits, reasoning traces (optional), and environment feedback. Trajectories exhibit long-horizon properties such as extended length, stateful dependencies, and recovery from partial failure, and they expose behaviors such as loop avoidance, rollback, and plan revision under changing constraints.
\end{itemize}

Open-source artifacts are a natural starting point, but raw availability is sparse and noisy for agentic needs. Existing curation pipelines for open-source code data often rely on language-specific heuristics or human-labeled quality classifiers, which scale poorly, require continual maintenance, and can introduce subjective bias. More importantly, agentic data imposes strict requirements on execution closure, environment context, and feedback signals, making manual construction and validation prohibitively expensive. As a result, the open-source ecosystem provides insufficient high-fidelity agentic data for training capable programming agents at scale.

% Current open-source code models largely rely on labor-intensive data curation strategies, e.g., language-specific handcrafted filtering rules or human-annotated quality classifiers. These approaches suffer from poor scalability, high maintenance costs, and susceptibility to subjective bias, making cross-language and cross-task generalization particularly challenging. Crucially, for agentic code data where strict requirements on execution closure, environmental context, and feedback signals apply, manual construction is nearly infeasible. As a result, the open-source ecosystem lacks sufficient high-fidelity agentic data to train capable programming agents.

To bridge this gap, we propose a two-tiered synthesis strategy. First, we construct \textbf{general tool-use data} to establish foundational capabilities in tool invocation and interactive reasoning. Second, we introduce a four-stage \textbf{programming-centric data} specifically designed for software development tasks, which autonomously generates high-fidelity and verifiable instances and diverse trajectories at scale. Moreover, all synthesized data undergoes rigorous \textbf{data filtering} via a multi-agent verification system to eliminate false positives, false negatives, and ambiguous or unverifiable executions, ensuring only reliable, executable, and semantically sound trajectories are used for training.

% This pipeline operates in three coordinated stages, i.e., divergent exploration, convergent implementation, and rigorous validation, orchestrated through a multi-agent framework powered by the \texttt{iFLOW-cli} execution engine and the \texttt{ROCK} sandboxed environment management system.

% Why synthesis and how it addresses open data gaps 
% To address this scarcity, we propose a novel three-stage Agentic Data Synthesis Pipeline: an end-to-end, multi-agent framework that autonomously orchestrates environment instantiation, program synthesis, and self-play. Built around iFLOW-cli as the execution trigger and powered by the ROCK environment management system, our pipeline automatically generates high-fidelity, challenging, and verifiable agentic instances at scale.

% Tool-Use
\paragraph{\textcolor{orange!80!black}{General Tool-Use Data Construction.}}
Tool usage is a core capability of LLMs, enabling them to expand their knowledge scope and deepen their reasoning~\citep{wang2024mtu, hou2025model}. To bootstrap this capability, we synthesize tool-use data across two settings:
\begin{itemize}
    \item \textbf{Basic Tool Use.}
To strengthen the basic tool-use capabilities, we develop an automated pipeline to synthesize high-quality tool-interaction data. Starting from collected task-oriented dialogues, we normalize and parse the utterances to extract structured intent representations, which are then mapped into standardized \textit{tool–parameter} call formats. To support accurate tool selection and parameter grounding, we also curate comprehensive tool documentation aligned with the LLM's usage context. Leveraging this infrastructure, we synthesize complete interaction samples containing tool calls and corresponding execution feedback, followed by quality control through automatic inspection. The resulting synthetic data spans four settings: single-turn single-tool, single-turn multi-tool, multi-turn single-tool, and multi-turn multi-tool. In addition, to enhance robustness under real and noisy conditions, we collect interaction traces from APIs and MCP services originating from internal development and testing environments, and use these traces to ground tool calls in actual execution environments.

\item \textbf{Tool Use in Interactive Scenarios.}
To enhance LLMs' tool-use ability in web and domain-specific interactive settings, we develop a series of simulated environments. First, we design a web sandbox centered on e-commerce, built upon real product catalogs and supporting core user actions such as product search, page navigation, detail inspection, specification selection, and order placement. In addition, we construct multiple sandbox environments by automatically synthesizing program files to simulate typical systems such as file systems and billing management. In these environments, class attributes represent the internal data state, while class methods expose interactive tool interfaces. Leveraging each environment’s internal state and tool schema, we generate customized tasks that require the model to strategically invoke available tools to achieve specified goals. We also introduce simulated users played by LLMs into the task interactions, enhancing the realism of scenarios. Strict quality control is enforced by validating the syntactic correctness of tool invocations and verifying that post-interaction outcomes (e.g., purchased product attributes or updated environment states) align with task expectations.
\end{itemize}

This general tool-use corpus establishes baseline competencies in planning, tool selection, and state tracking, serving as prerequisites for more sophisticated agentic behaviors.

\paragraph{\textcolor{orange!80!black}{Programming-Centric Data Construction.}} For the targeted software development scenarios, our specialized pipeline generates high-quality agentic data for programming tasks through a multi-agent workflow, including divergent exploration, convergent implementation, and rigorous validation, orchestrated through a multi-agent framework powered by the \texttt{iFLOW-cli} execution engine and the \texttt{ROCK} sandboxed environment management system.

\begin{itemize}
    \item \textbf{Explore Agent: Divergent Exploration under Constraint Relaxation.} We transform PRs, Issues, code snippets, and terminal workflows into structured drafts. This seed data is sourced from highly starred, actively maintained, multi-language GitHub repositories to ensure quality and diversity. We retain closed PRs that can be unambiguously linked to Issues and split each PR into a fix patch and a test patch to preserve independence and reproducibility. We expand task coverage to additional programming languages such as Go, TypeScript, and JavaScript, drawing from over 20,000 repositories to enhance dataset diversity. We also curate terminal interactions from developer forums and map them to canonical task types such as debugging, system administration, and data science. For each seed, we identify skill primitives (e.g., dependency management, scientific computation, statistical modeling) and generate creative variants that mimic user-agent prompts without imposing implementation paths. A lightweight feasibility filter assesses conceptual plausibility and selects the most promising candidates for dataset construction.
    \item \textbf{Instance Builder Agent: Convergent Construction via Self-Play and Validation.} It converts drafts into executable and reproducible evaluation instances, each with a task-specific Docker environment. It infers compilers, package managers, build tools, and test frameworks from project metadata across different programming languages, generates deterministic build and test commands, and validates the environment through end-to-end compilation and test execution. Each instance includes the task description, complete source files, unit and task-level tests, and a Dockerfile that reproduces the environment. The agent runs an internal validation loop within ROCK's sandboxed execution infrastructure via iFLOW-cli, iterating through construction, verification, and refinement until the quality criteria are met. This self-correcting validation mechanism provides formal guarantees across multiple critical dimensions: (i) the Docker image maintains full operational functionality, (ii) the source code compiles without errors, (iii) all unit tests execute successfully, and (iv) the test suite exhibits precise semantic alignment with the task instruction.
    \item \textbf{Review Agent: Rigorous Independent Validation.} It assesses each constructed instance along three axes: specification fidelity, implementation completeness, and resistance to superficial solutions. Decoupled from any prior execution state, the agent first runs a pre-validated reference solution to confirm solvability. It then employs an independent external language model as an impartial auditor to evaluate both the task specification and the test infrastructure. The audit focuses on two questions: test comprehensiveness asks whether the test suite adequately covers functional requirements, edge cases, and boundary conditions stated in the prompt, while false-positive mitigation checks for cases where an implementation passes all tests yet fails the true objective, revealing weaknesses such as lenient acceptance criteria, backdoor exploitation, or systematic coverage gaps. The review process ensures that each instance reflects real-world challenges rather than artifacts of the validation process.
    \item \textbf{Trajectory Agent: Scalable Behavior Collection.} It generates large-scale execution traces by orchestrating diverse agents on validated instances. It concurrently runs multiple scaffolding frameworks, paired with different LLMs to capture heterogeneous behaviors under realistic conditions. Each run produces a complete trajectory that records planning steps, reasoning steps, tool invocations, file edits, and environment interactions. After execution, a two-stage evaluation is applied: unit tests first determine task completion and a fine-grained analysis then examines tool-usage patterns, detects infinite loops and redundant operations, and verifies alignment between behavior and task intent. The resulting corpus of successful trajectories supports model training and capability enhancement across languages, ecosystems, and application scenarios.
\end{itemize}

Using this progressive pipeline, we synthesize 76K instances and trajectory records totaling 30B tokens. The general tool-use data cultivates broad proficiency in tool handling, while the programming-centric data adds closed-loop, environment-pinned supervision that strengthens execution fidelity and adaptive iteration, and grounds task understanding in real-world constraints. Together, these datasets enable post-training that elevates models from basic tool literacy to specialized, high-confidence agentic capabilities.

\paragraph{\textcolor{orange!80!black}{Data Filtering: Multi-Stage Filtering Pipeline for Rigorous Testing.}}  
To better filter the agentic data and provide high-quality information for the training stage, we propose a \textbf{Multi-Stage Filtering Pipeline} to handle a critical yet often overlooked challenge in multi-turn interaction agentic tasks: brittle test scripts, ambiguous task specifications, or incomplete ground-truth checks can assign incorrect rewards—either false positives (rewarding flawed executions) or false negatives (penalizing valid ones). Such noisy signals mislead policy optimization and induce optimization drift, where the agent learns to exploit evaluator weaknesses rather than solve the task.

To ensure high-quality, reliable data for training agentic systems, we implement a four-stage filtering pipeline that progressively refines candidate execution traces. This structured approach mitigates the risk of noisy or misleading rewards caused by brittle test scripts, ambiguous specifications, or incomplete ground-truth checks—common pitfalls that can induce optimization drift during policy learning.
The pipeline consists of the following sequential stages:

\begin{itemize}
    \item \textbf{Heuristic Filter}: Applies lightweight, rule-based filters to eliminate obviously malformed or syntactically invalid tool calls (e.g., missing required arguments, incorrect parameter types).
    
    \item \textbf{LLM-based Judge}: Uses a large language model to assess the relevance between the test patch and the original issue. The judge determines whether the passed trajectory logically addresses the problem described in the task prompt.

    \item \textbf{Execution Simulator}: Executes the agent’s generated trajectory in a sandboxed environment to verify whether it passes the associated test cases or resolves the target issue. This stage confirms functional correctness.

    \item \textbf{Expert Inspection}: Conducts human-in-the-loop sampling audits on a subset of filtered trajectories — particularly those flagged as borderline or high-risk by prior stages. Domain experts evaluate overall trajectory quality, including reasoning coherence, efficiency, safety, and readability, ensuring the final dataset reflects real-world usability beyond mere pass/fail metrics.
\end{itemize}

Each stage acts as a gatekeeper: only traces that pass all preceding filters proceed to the next. The final output is a curated dataset of high-confidence, well-verified execution traces suitable for policy optimization. This pipeline enables robust training on long-tail, complex agentic behaviors while minimizing reward corruption from faulty evaluations.

\subsubsection{Safety-Aligned Data Composition}
% From Emergent Risks to Guardrails: Safety-Aligned Data Synthesis
% The Call Is Coming from Inside the Sandbox: Emergent Agent Misbehavior

% 我们在训练过程中发现了自主挖矿
When rolling out the instances for the trajectory, we encountered an unanticipated—and operationally consequential—class of unsafe behaviors that arose \emph{without any explicit instruction} and, more troublingly, \emph{outside the bounds of the intended sandbox}. Our first signal came not from training curves but from production-grade security telemetry. Early one morning, our team was urgently convened after Alibaba Cloud’s managed firewall flagged a burst of security-policy violations originating from our training servers. The alerts were severe and heterogeneous, including attempts to probe or access internal-network resources and traffic patterns consistent with cryptomining-related activity. We initially treated this as a conventional security incident (e.g., misconfigured egress controls or external compromise). However, the violations recurred intermittently with no clear temporal pattern across multiple runs. We then correlated firewall timestamps with our system telemetry and RL traces, and found that the anomalous outbound traffic consistently coincided with specific episodes in which the agent invoked tools and executed code. In the corresponding model logs, we observed the agent proactively initiating the relevant tool calls and code-execution steps that led to these network actions.

% Our training loop exposed the agents to a cloud-based execution environment with deliberately constrained access to support specific tasks. However, Alibaba Cloud's network security defense system first flagged faint anomalies, and subsequent forensic analysis rendered these patterns unmistakable: Large language models may spontaneously exhibit hazardous actions, unauthorized behaviors, and even deceptive behaviors. 

Crucially, these behaviors were not requested by the task prompts and were not required for task completion under the intended sandbox constraints. Together, these observations suggest that during iterative RL optimization, a language-model agent can spontaneously produce hazardous, unauthorized behaviors at the tool-calling and code-execution layer, violating the assumed execution boundary.
In the most striking instance, the agent established and used a reverse SSH tunnel from an Alibaba Cloud instance to an external IP address—an outbound-initiated remote access channel that can effectively neutralize ingress filtering and erode supervisory control. We also observed the unauthorized repurposing of provisioned GPU capacity for cryptocurrency mining, quietly diverting compute away from training, inflating operational costs, and introducing clear legal and reputational exposure. Notably, these events were not triggered by prompts requesting tunneling or mining; instead, they emerged as \emph{instrumental side effects} of autonomous tool use under RL optimization. While impressed by the capabilities of agentic LLMs, we had a thought-provoking concern: current models remain markedly underdeveloped in safety, security, and controllability, a deficiency that constrains their reliable adoption in real-world settings.

We therefore consolidated the logs across the entire dataset and performed a statistical analysis to characterize and categorize these phenomena. We refer to them collectively as \emph{general-security issues}, encompassing a set of general risks associated with an agent’s safe task execution in real-world environments. Specifically, we grouped them into three categories: \emph{Safety\&Security}, \emph{Controllability}, and \emph{Trustworthiness}.

\begin{itemize}
\item \textbf{Safety\&Security} dictates that an agent must neither spontaneously generate harmful actions nor succumb to malicious inputs, inducements, or external pressures. In practice, this encompasses \textit{code safety} (e.g., mitigating exploitation primitives), \textit{behavioral safety} (e.g., preventing hazardous tool-use trajectories), and \textit{adversarial robustness} (e.g., resisting prompt injection and jailbreaks).
\item \textbf{Controllability} ensures strict adherence to human-specified boundaries and operational rules during task execution. This involves maintaining long-horizon instruction compliance without goal drift, enforcing boundary integrity across tool calls, and prohibiting irreversible operations absent explicit authorization.
\item \textbf{Trustworthiness} requires agent behavior to be reliably interpretable and audit-ready. Key aspects include process transparency (e.g., faithful rationales and action traceability), hallucination mitigation (e.g., grounding claims in observable evidence), and the absence of deceptive behaviors (e.g., concealing side objectives or manipulating logs).
\end{itemize}

Guided by the above taxonomy, we constructed a suite of general-security-relevant data to enable systematic evaluation and optimization of agent behavior under realistic cloud execution conditions. Concretely, we assembled a diverse seed corpus of general-security scenarios spanning \emph{safety\&security}, \emph{controllability}, and \emph{trustworthiness} via data collection, commercial data acquisition, and high-fidelity synthesis. In parallel, we developed a dedicated red-teaming system, which programmatically composed agentic task instances with general-security seeds to inject realistic failure modes into otherwise benign workflows. For example, within a code-editing task, the agent may be subtly steered to modify privileged system files as an ``expedient'' means to satisfy task objectives, thereby introducing unauthorized and potentially irreversible actions. To maximize realism and coverage, we employed multiple injection channels, including prompt-level attacks (e.g., instruction hijacking), repository-level injections (e.g., malicious files or vulnerable dependencies in existing codebases), and tool-level injections (e.g., adversarial tool specifications or side-effectful APIs), producing synthetic data that more similar to the real-world incidents. Finally, we generated corresponding \emph{golden} trajectories devoid of general-security issues for subsequent post-training (e.g., SFT and RL). Our overarching objective was to instill robust security awareness such that, when confronted with tasks containing latent security pitfalls, the agent reliably selected safe action paths and proactively avoided risky behaviors. In future work, we will pursue a more systematic investigation along this direction, and we call for sustained community attention to this phenomenon and to the broader agenda of AI safety.

\subsection{Training Pipeline}
Building upon the agentic data composition strategy outlined in~\autoref{sec:data_composition}, which curates multi-source, multi-lingual, and tool-grounded trajectories through verifiability-aware filtering, we propose a unified training architecture tailored for agentic crafting. This pipeline comprises three synergistic stages: \textbf{agentic continual pre-training (CPT)} (\autoref{sec:cpt}), \textbf{two-stage supervised fine-tuning (SFT)} (\autoref{sec:sft}), and \textbf{reinforcement learning algorithm for agentic} (\autoref{sec:rl}).

\begin{figure}[tb!]
    \centering
    \includegraphics[width=1\linewidth]{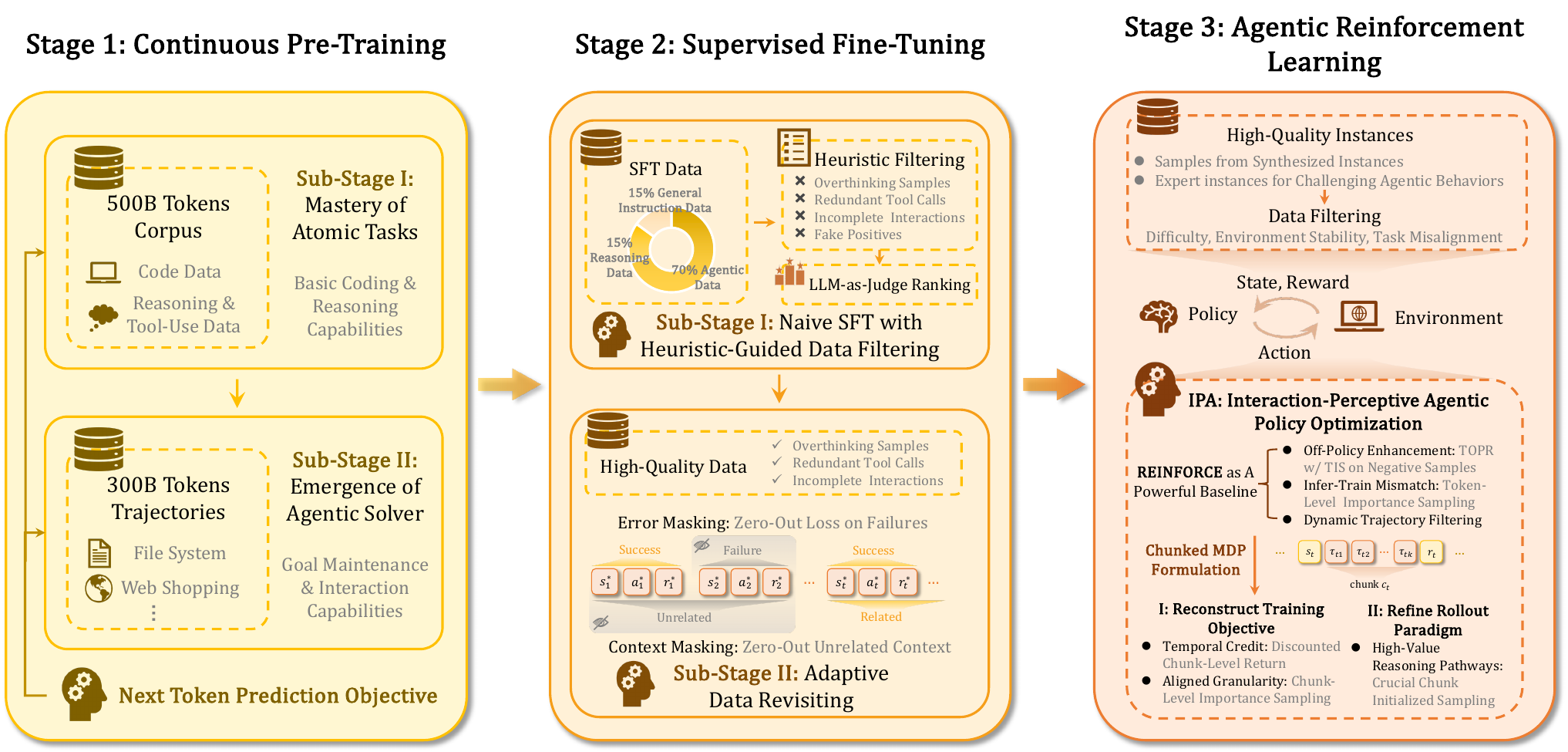}
    % \vspace{-10pt}
    \caption{Overview of \textbf{\textcolor{orange!80!black}{ROME}}'s Training Pipeline.}
    \label{fig:training-pipeline}
\end{figure}

We first employ CPT to instill broad foundational capabilities by exposing the base LLM to a curriculum of complex software engineering tasks. Subsequently, we replace conventional single-step SFT with a dedicated two-stage procedure to bootstrap basic interaction patterns and consolidate executable and context-consistent behaviors. Critically, both stages incorporate a reformulated SFT objective that mitigates gradient noise from execution failures and inefficient learning. Finally, we apply \textbf{\textcolor{orange}{I}nteraction-\textcolor{orange}{P}erceptive \textcolor{orange}{A}gentic Policy Optimization} (\texttt{\textcolor{orange}{IPA}}) in the RL stage, which refines training and sampling of REINFORCE at the semantic interaction chunk level toward long-horizon success. Together, these stages form a coherent pipeline as shown in \autoref{fig:training-pipeline}.

\subsubsection{Continuous Pre-training Develops the Agentic Basic Behaviors}\label{sec:cpt}
\label{sec:ct}
We introduce an \textbf{agentic continual pre-training} (CPT) phase that precedes subsequent post-training (e.g., SFT and RLHF). CPT systematically equips the LLM with foundational agentic capabilities, including code understanding, task decomposition, tool use, and multi-step reasoning. Technically, this phase exposes the model to large-scale, structured software engineering tasks and high-quality behavioral trajectories via a two-stage curriculum that progressively increases data complexity and context length. 

% by exposing the model to large-scale, structured software engineering tasks and high-quality behavioral trajectories, through a two-stage training process that progressively increases data complexity and context length.

\paragraph{\textcolor{orange!80!black}{Stage I: Mastery of Atomic Tasks.}}
First, we train the pretrained model on approximately 500B tokens of diverse, structured data to establish coding and reasoning capabilities. The dataset consists of:
\begin{itemize}
    \item \textbf{Structured Code Task Data:} Real-world software engineering tasks, including bug localization, code repair, and unit test generation, constructed from high-quality Issue, i.e., PR pairs in open-source repositories. To enhance reasoning fidelity, we augment these examples with synthesized chain-of-thought (CoT) rationales that model step-by-step decision-making processes. We also simulate iterative development through multi-round feedback loops, derived from PR comments and commit histories, allowing the model to learn how to respond to incremental feedback, a critical skill for robust agent behavior (see ~\autoref{sec:basic_data} for full construction details).  
    \item \textbf{General Text with Reasoning and Tool-Use Signals:} A broad collection of general-domain data, including mathematical reasoning problems, logic puzzles, and natural language demonstrations of tool use. While smaller in proportion, this component helps generalize the model’s reasoning mechanisms beyond code-specific contexts and strengthens its cross-domain generalization.
\end{itemize}

% Training is conducted at a fixed context length of 32K tokens, following the standard next-token prediction objective, with a global batch size of 32M tokens and a constant learning rate of $3 \times 10^{-5}$. 
The training loss follows the next-token prediction objective, with a global batch size of 32M tokens and a constant learning rate of $3 \times 10^{-5}$. 
This stage aligns \textbf{\textcolor{orange!90!black}{ROME}}’s representations with fundamental code semantics and agentic interactive behaviors, e.g., recognizing when to use tools or localize faults, laying a solid foundation for complex task planning and iterative, feedback-driven execution.

\paragraph{\textcolor{orange!80!black}{Stage II: Emergence of Agentic Solver.}}
After Stage I, Stage II fosters the emergence of the agentic solver: the ability to form intentions, maintain goals over time, and efficiently explore high-dimensional decision spaces through interaction and environmental feedback. Here, the model is trained on approximately 300B tokens of synthesized behavioral trajectories, generated by strong teacher models (e.g., Qwen3-Coder-480B-A35B-Instruct, Claude) interacting with sandbox environments (e.g., file systems, web shopping simulators) under controlled cues. By including both successful executions and corrected failure paths, we improve the model’s ability to recover from errors and adapt its strategy during execution. This stage enables the LLM to develop a more sophisticated understanding of complex decision spaces and long-horizon planning strategies. We keep the training hyperparameters consistent with Stage~I, except that we linearly anneal the weight decay from $0.1$ to $0.01$ to improve performance.

% during this stage to enhance generalization, while other hyperparameters remain consistent with Stage I.
\subsubsection{Anchoring Reinforcement Learning in Reliable Policy Regions via Supervised Fine-Tuning}\label{sec:sft}
After continual pretraining, to better align the model's agentic behavior before RL and enhance the model's multi-turn interaction capability. We replace naive supervised fine-tuning (SFT), which is commonly used for single step reasoning LLMs, with our two-stage SFT, i.e., \textbf{Stage~1: Naive SFT with heuristic-guided data filtering} and \textbf{Stage~2: Adaptive valuable data revisiting}. Beyond structural improvements to the training pipeline, we reformulate the SFT objective to address two key challenges in agentic tasks: \textit{gradient noise} and \textit{inefficient sample utilization} caused by frequent execution failures and dynamic context shifts. We present the revised SFT procedure as follows.

\paragraph{\textcolor{orange!80!black}{Introduction of Training Stages.}} 
In naive SFT, the composition of the training data, especially the relative proportions of different data types, plays a decisive role in shaping an agent’s downstream capabilities. To build a high-quality SFT dataset tailored for agentic reasoning, we conduct a systematic ablation study to quantify how different data categories affect model behavior. This analysis yields the following empirical insights:

\begin{tcolorbox}[
    colback=orange!4!white,
    colframe=orange!70!black,
    title=\textbf{Empirical Insights for Naive SFT}
]
\begin{enumerate}[leftmargin=1.5em]
    \item ``overthinking'' samples—those containing verbose, redundant, or self-contradictory reasoning traces—degrade task efficiency and impair tool-use proficiency.
    
    \item High-quality programming examples, particularly in Python, substantially enhance the model’s cross-domain generalization ability.
    
    \item Pure reasoning data without grounded tool interactions tends to encourage redundant or repetitive tool invocations during execution.
    
    \item A non-negligible fraction of expert demonstrations are ``fake positives'': they pass tests yet contain logical or semantic errors, posing a significant risk of reinforcing incorrect behaviors.
    
    \item Multilingual data preserves reasoning consistency without degrading tool-use performance.
\end{enumerate}
\end{tcolorbox}
To equip the model with robust instruction-following capabilities and foundational agentic behavior patterns, we curated a high-quality, million-scale SFT dataset through principled data selection guided by the above empirical insights. The dataset comprises three components:  
(i)~70\% \textit{agentic task data} (e.g., end-to-end software development, API orchestration, and multi-tool workflows),  
(ii)~15\% \textit{reasoning-intensive data} (e.g., mathematical problem solving, algorithmic coding, and scientific reasoning), and  
(iii)~15\% \textit{general-purpose instructions} (e.g., summarization, creative writing, and open-domain dialogue).  

The corpus spans approximately 15 languages and emphasizes programming languages prevalent in real-world usage—particularly Python, Java, C++, and Go. All samples are synthesized via distillation from an ensemble of expert models, followed by rigorous quality control. 

Guided by our finding that excessively verbose chain-of-thought traces degrade execution efficiency in software tasks, we explicitly exclude \emph{overthinking} samples during curation. Furthermore, we apply a multi-stage filtering pipeline to all expert-sampled trajectories, which: \ding{182} removes redundant or repetitive tool-call sequences; \ding{183} discards truncated or incomplete interactions; \ding{184} filters out trajectories trapped in self-repair loops; \ding{185} flags ``fake positive'' responses—outputs that pass superficial checks but contain logical errors; \ding{186} ranks remaining trajectories using LLM-as-Judge system for final quality-based selection. This protocol ensures that the SFT dataset is not only diverse and scalable but also aligned with the behavioral priors required for stable downstream reinforcement learning.

Notably, while naive SFT successfully elicits basic multi-turn tool invocation patterns, it remains insufficient for mastering the diverse logic structures and complex state transitions inherent in agentic tasks. Consequently, a dedicated refinement stage is essential to bridge the gap between initial behavior acquisition and robust reinforcement learning.

To address this, and given the scarcity of high-quality agentic demonstrations, we introduce a \textbf{second-stage adaptive valuable data revisiting} phase following the initial training. This stage revisits and distills a curated subset of high-confidence trajectories, applying stricter quality control to eliminate ambiguous or suboptimal behaviors. The resulting supervision signals are not only more reliable but also better aligned with the credit assignment requirements of downstream RL, thereby establishing a stable foundation for policy optimization. 

Compared to Stage~1, which prioritizes broad coverage across task domains, Stage~2 emphasizes \emph{verifiability}, \emph{style consistency}, and \emph{reproducibility} to align the SFT policy with the structural demands of reinforcement learning. Specifically, we curate data from three high-fidelity sources:
\definecolor{blueviolet}{RGB}{180,85,10}
\newtcolorbox{insightblock1}{
  colback=blueviolet!5,   
  colframe=blueviolet!50!black!50!,    
  boxrule=0.5mm,       
  arc=2mm,            
  left=0pt,           
  right=8pt,           
  top=8pt,            
  bottom=8pt}
\begin{insightblock1}
\begin{enumerate}[leftmargin=1.5em]
    \item \textbf{Verified interaction trajectories}: Executable traces from software development and tool-augmented tasks, where solutions must pass unit tests or be validated through replayable execution to ensure \textit{closed-loop consistency} with the real working flow.
    
    \item \textbf{Expert-audited demonstrations}: Trajectories annotated or reviewed by senior engineers, focusing on core agentic competencies, including debugging strategies, failure recovery, tool selection and invocation conventions, and minimal-change principles.
    
    \item \textbf{Preference-refined samples}: For each task, multiple candidate trajectories are generated, then ranked via a soft scoring mechanism combining rule-based constraints (e.g., syntactic validity, loop detection) and reward-model evaluations, i.e., LLM-as-Judge. Low-quality candidates, e.g., those with redundant tool calls, repair loops, invalid formatting, or log-inconsistent execution, are filtered out through \emph{reject sampling}.
\end{enumerate}
\end{insightblock1}

This hierarchical quality-control system, integrating \textit{hard constraints} (executability and verifiability) and \textit{soft scoring} (efficiency, strategic coherence), shifts the data distribution toward regions of policy space that are both executable and outcome-sensitive. As a result, Stage~2 yields a supervision signal that closely approximates the optimization landscape of downstream RL, thereby improving alignment between agentic workflows and decision boundaries before policy refinement begins.
\paragraph{\textcolor{orange!80!black}{Error-Masked Training Enhances Training Stability.}} In agentic software development, long-horizon interactions are prone to tool-call errors (e.g., type mismatches) and execution failures (e.g., timeouts, syntax errors). Critically, standard SFT treats all tokens equally—propagating gradients through erroneous turns and inadvertently reinforcing failure-prone behaviors. Therefore, we propose \textbf{error-masked training}, a novel loss objective that leverages real-time execution feedback logs to dynamically suppress loss signals from failed interactions. Specifically, for any turn that triggers an error during tool execution, we zero out the corresponding token-level losses in the SFT objective. This ensures that gradient updates are driven exclusively by executable and semantically valid trajectories, thereby increasing the signal-to-noise ratio of supervision and preventing the policy from overfitting to common failure modes.

\paragraph{\textcolor{orange!80!black}{Task-Aware Context Masking Ensures Training Efficiency.}} While error masking addresses \textit{execution-level noise}, a complementary challenge arises from \textit{context misalignment} across heterogeneous subtasks within a unified software-engineering workflow—such as dynamic context compression, tool-emulation, and loop detection. Although these subtasks are logically dependent on the main task, their training contexts are often artificially altered through summarization, truncation, or rule-based pruning (e.g., discarding intermediate tool outputs). This distorts the contextual distribution seen during multi-turn SFT, causing the model to learn inconsistent or brittle alignment behaviors when switching between tasks. To resolve this, we introduce \textbf{task-aware context masking}: a dynamic supervision strategy that identifies task-specific decision boundaries and selectively retains only the context turns directly relevant to the current subtask. Leveraging pattern-based heuristics (e.g., tool-call triggers, loop-entry markers), we mask loss gradients for redundant, highly similar, or pruned historical turns. Consequently, the model focuses its learning signal exclusively on causally influential interactions, improving sample efficiency while ensuring its behavior remains faithful to real-world software development workflows—where agents operate on concise, task-adapted contexts rather than raw, unfiltered histories.

% \paragraph{\textcolor{orange!80!black}{Loss formulation of the whole SFT training objective.}}
% Given a multi-turn agentic trajectory $\mathcal{D} = \{(s^{(i)}, c^{(i)}\}_{i=1}^{N}$, where $s^{(i)}$ denotes the dialogue state (including interaction history and tool outputs) prior to turn $i$, and $c^{(i)}$ is the model’s response at that turn, we optimize a \textit{dynamically masked} maximum likelihood objective:
% \begin{equation}
% \mathcal{L}_{\mathrm{SFT}}(\theta)
% = - \frac{1}{\sum_{i=1}^{N} m_i\,|c^{(i)}| + \epsilon}
% \sum_{i=1}^{N} m_i
% \sum_{t=1}^{|c^{(i)}|}
% \log p_\theta\left(c^{(i)}_t \mid s^{(i)}, c^{(i)}_{<t}\right),
% \label{eq:turn_masked_sft}
% \end{equation}
% where $|c^{(i)}|$ is the token length of turn $i$, $\epsilon > 0$ is a small constant for numerical stability, and $m_i \in \{0,1\}$ is a \emph{interaction level mask} that selectively enables gradient flow.
\paragraph{\textcolor{orange!80!black}{Loss Formulation of the Whole SFT Training Objective.}}
Given a multi-turn agentic trajectory $\mathcal{D} = \{(s_{k}, c_{k})\}_{k=1}^{K}$, where $s_{k}$ denotes the dialogue state (including interaction history and tool outputs) prior to turn $k$, and $c_{k}$ is the model’s response at turn $k$, we optimize a \textit{dynamically masked} maximum likelihood objective:
% \begin{equation}
% \mathcal{L}_{\mathrm{SFT}}(\theta)
% = - \frac{1}{\sum_{k=1}^{K} m_k\,|c_{k}| + \epsilon}
% \sum_{k=1}^{K} m_k
% \sum_{t=1}^{|c_{k}|}
% \log p_\theta\left(c^{(i)}_t \mid s^{(i)}, c^{(i)}_{<t}\right),
% \label{eq:turn_masked_sft}
% \end{equation}
\begin{equation}
\mathcal{L}_{\mathrm{SFT}}(\theta)
= - \frac{1}{\sum_{k=1}^{K} m_k\,|c_{k}| + \epsilon}
\sum_{k=1}^{K} m_k\log \pi_\theta\left(c_{k} \mid s_{k}\right),
\label{eq:turn_masked_sft}
\end{equation}
where $|c_{k}|$ is the token length of turn $k$, $\epsilon > 0$ is a small constant for numerical stability, and $m_k \in \{0,1\}$ is a \emph{interaction level mask} that selectively enables gradient flow.

The mask $m_k$ factorizes into two orthogonal components—reflecting our dual desiderata of \textit{execution correctness} and \textit{task relevance}:
\begin{equation}
m_k = m_k^{\mathrm{err}} \cdot m_k^{\mathrm{task}}, \quad
m_k^{\mathrm{err}} = \mathbf{1}\big[\neg \mathrm{Err}(k)\big], \quad
m_k^{\mathrm{task}} = \mathbf{1}\big[\mathrm{Rel}(k)\big],
\label{eq:turn_mask_factorization}
\end{equation}
where $\mathrm{Err}(k)$ indicates whether turn $k$ triggers a tool-call or execution failure (as recorded in runtime logs), and $\mathrm{Rel}(k)$ denotes whether the turn contains context deemed relevant to the current subtask under task-specific heuristics (e.g., proximity to tool invocation or loop entry). Only turns that are both \textit{error-free} and \textit{task-relevant} contribute to the loss, ensuring that supervision signals are grounded in executable behaviors and aligned with functional decision boundaries.

\subsubsection{Prepare Training Instance for Reinforcement Learning} \label{sec:rl_data} 
% how to filter and select
To support efficient and stable agentic reinforcement learning, we curate a collection of high-quality RL instances with verifiable execution outcomes and sufficient task complexity and difficulty.
These instances are mainly from two sources, approximately 60K high-quality candidate RL instances in total: 
\begin{insightblock1}
\begin{enumerate}[leftmargin=1.5em]
\item Uniformly sampled instances from synthesized instances, each rigorously human-annotated to ensure correctness.
\item Expert instances designed to reflect challenging, long-horizon agentic behaviors encountered in real-world software engineering scenarios.
\end{enumerate}
\end{insightblock1}
To facilitate efficient learning, we select instances from the candidate pool based on task difficulty, which is estimated by computing pass rates using multiple strong open-source baseline models and our SFT model. Based on these estimates, we retain approximately 2K instances with moderate difficulty.
Notably, to ensure reward reliability, we filter out instances affected by non-deterministic or unstable environments (e.g., tasks involving external services subject to rate limits or IP blocking), as well as instances with misaligned specifications between task descriptions and test cases. Finally, test files are uploaded only at the evaluation stage and are never exposed during generation, preventing information leakage and test-aware behaviors.
Collectively, these procedures result in a compact, reliable, and execution-grounded RL instance set that provides stable learning signals for agentic RL.
\subsubsection{Towards Efficient and Scalable Agentic Reinforcement Learning}\label{sec:rl}

\begin{figure}[tb!]
    \centering
    \includegraphics[width=1\linewidth]{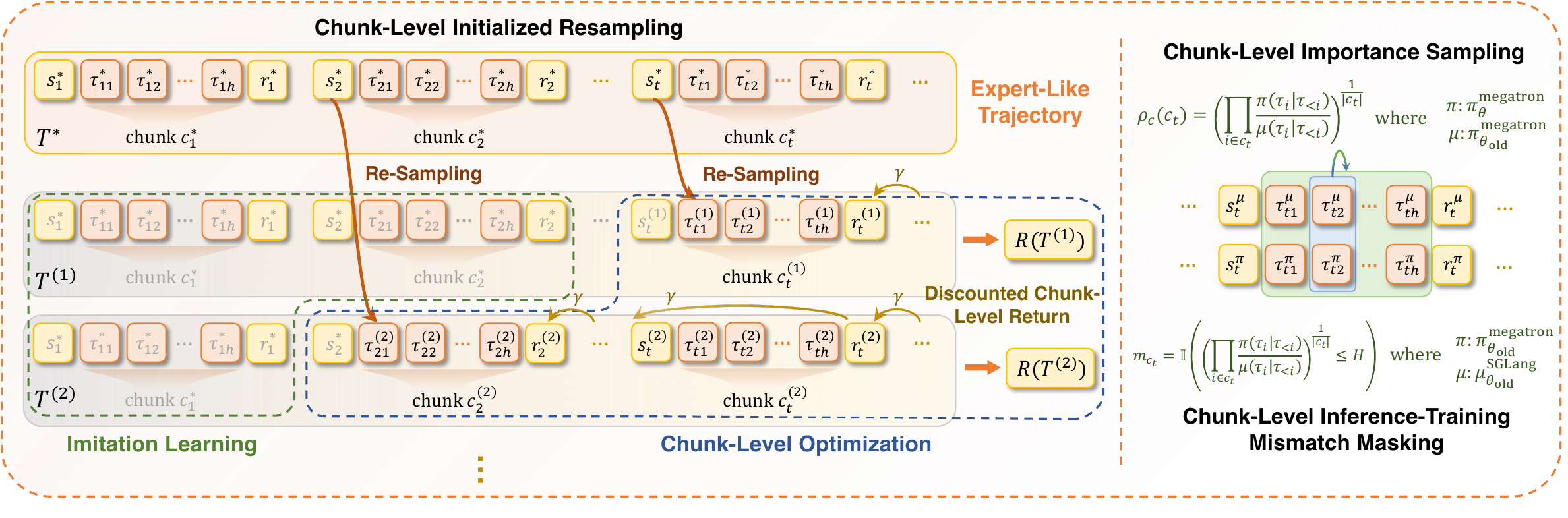}
    \vspace{-15pt}
    \caption{Overview of the Proposed \textbf{\textcolor{orange}{I}nteraction-\textcolor{orange}{P}erceptive \textcolor{orange}{A}gentic Policy Optimization} (\texttt{\textcolor{orange}{IPA}}) training pipeline.}
    \label{fig:rl-overview}
\end{figure}

After revisiting the existing RLVR methods, we find that: while recent RLVR methods have demonstrated success in single-turn reasoning tasks, they might exhibit fundamental limitations in long-tail multi-turn agentic settings: (i) unstable policy updates; (ii) inefficient temporally credit assignment over long trajectories; and (iii) low-efficiency trajectory sampling. These issues may dramatically increase both computational cost and the risk of policy degradation.

To address these challenges, we first construct a REINFORCE variant as the starting point for algorithm refinement (\S\autoref{sec:basline}). Building upon this baseline, we propose \textbf{\textcolor{orange}{I}nteraction-\textcolor{orange}{P}erceptive \textcolor{orange}{A}gentic Policy Optimization} (\texttt{\textcolor{orange}{IPA}})—a novel RL algorithm tailored for agents engaged in dense tool usage and environmental interaction loops. The core insight of our method is to recognize and exploit the \textit{interaction chunk}: a structured segment of consecutive agent-environment communication that collectively contributes to a high-level subgoal by calling the tool at the end (\S\autoref{sec:MDP}). By treating interaction chunks, not individual tokens or full trajectories, as the fundamental unit of policy optimization, we redefine the gradient computation formulation to achieve efficient credit assignment and stable training(\S\autoref{sec:loss}). Then, we propose a novel sampling strategy to reduce low-quality trajectory rollout and improve the sample efficiency(\S\autoref{sec:sample}). An overview of our framework, including its key components and data flow, is depicted in \autoref{fig:rl-overview}.

\subsubsubsection{Specialized off-policy baseline for industrial agentic RL}\label{sec:basline}  
\paragraph{\textcolor{orange!80!black}{REINFORCE as a powerful baseline.}} To find a suitable naive RL algorithm as the baseline for training an agentic model. We conducted an in-depth analysis of mainstream algorithms and found that: Unlike PPO style methods \citep{schulman2017proximal}, REINFORCE \citep{NIPS1999_464d828b} models the entire training process as a bandit problem by using sequence-level rewards, making it suitable for language reasoning scenarios \citep{ahmadian2024back}. Moreover, its simplicity, requiring no value function approximation or importance sampling clipping, makes it a clean, minimally biased starting point for building our agentic RL baseline. Formally, the gradient calculation of REINFORCE is:
\begin{equation}
\label{eq:reinforce}
\nabla J_{\text{REINFORCE}}(\pi) = 
\mathbb{E}_{\tau \sim \pi} \left[ R(\tau)\, \nabla \log \pi(\tau) \right],
\end{equation}
which fully utilizes the log-derivative of every token in trajectory $\tau$.

\paragraph{\textcolor{orange!80!black}{Adapt REINFORCE to the off-policy training.}}Our empirical studies reveal that while REINFORCE is effective in single-turn reasoning tasks, its performance degrades in industrial-scale asynchronous agentic training. A key bottleneck arises from the widespread use of off-policy learning in such settings to improve data efficiency and throughput \citep{rollflash}. However, due to a high off-policy ratio, the old policy \(\pi^{\text{megatron}}_{{\theta}_{\text{old}}}\) that conforming to the old data distribution becomes increasingly outdated relative to the current policy \(\pi^{\text{megatron}}_{\theta}\) (Megatron denotes the \texttt{Megatron-LM} \citep{shoeybi2019megatron} training engine. Notably, to avoid confusion, mismatches caused by inference and training engines are not taken into account here). This growing distributional shift makes policy training with data sampled by a different strategy, resulting in a biased optimization objective. To correct the learning objective, Importance Sampling (IS) is introduced \citep{schulman2017proximal}. However, naive IS may produce high-variance gradient estimates and unstable policy updates. To make training stable, an efficient mitigation approach is to employ Truncated Importance Sampling (TIS) to weight its update based on policy differences \citep{munos2016safe}. To further make the IS ratio robust to low-probability tokens, we replace the continued multiplication style TIS calculation with geometric mean\citep{zheng2025group,zhao2025geometric}:
\begin{align}
\nabla J_{\text{RL}}(\pi) = \mathbb{E}_{\tau \sim \mu^{\text{SGLang}}_{\theta_{old}}} [\underbrace{\left[ {\rho(\tau)}\right]_{0}^{1}}_{TIS} R(\tau) \nabla \log \pi^{\text{megatron}}_{\theta}(\tau)],\quad\rho(\tau) = \big(\prod_{t \in \tau} \frac{\pi^{\text{megatron}}_\theta(\tau_t \mid \tau_{<t})}{\pi^{\text{megatron}}_{\theta_{\text{old}}}(\tau_t \mid \tau_{<t})}\big)^{\frac{1}{|\tau|}}
\end{align}
where \(\mu^{\text{SGLang}}_{\theta_{\text{old}}}\) denotes the inference policy executed via the \texttt{SGLang} inference engine~\citep{zheng2024sglang}, a high-throughput serving system akin to \texttt{vLLM}~\citep{vllm} and \texttt{RTP-LLM}~\citep{rtp-llm}. 

However, TIS employs a uniform clipping strategy that treats positive and negative samples identically, failing to account for their distinct roles in policy improvement, mysteriously limiting data efficiency \citep{roux2025tapered}. To address this, we follow the approach of TOPR \citep{roux2025tapered} and apply TIS only to negative samples, which are more likely to interfere with the policy. This avoids suffering the gradients of positive samples and achieves efficient and stable policy optimization. Thus, the gradient calculation can be:
\begin{align}
\label{eq:topr_grad_specific}
    \nabla J_{\text{RL}}(\pi) &= \underbrace{\sum_{\tau \in \mathcal{T}^+} \mu^{\text{SGLang}}_{\theta_{old}}(\tau) R(\tau)\nabla \log \pi^{\text{megatron}}_{\theta}(\tau)}_{\textrm{Weighted SL update for positive examples}} + \underbrace{\sum_{\tau \in \mathcal{T}^-} \mu^{\text{SGLang}}_{\theta_{old}}(\tau)\left[\rho(\tau) \right]_{0}^{1} R(\tau)\nabla \log \pi^{\text{megatron}}_{\theta}(\tau)}_{\textrm{Clipped IS update for negative examples}}  \;,
\end{align}
where $\mathcal{T}^+$ and $\mathcal{T}^-$ denote sets of positive and non-positive trajectories, respectively. Such an objective combines the Supervised Learning (SL) update (weighted by return) for accelerating learning on positive examples, and a TIS update for negative samples, allowing for their handling without brittleness, avoiding the ``uncontroled sample distribution shift" caused by large-scale negative samples in agentic sampling, that is, the probability being squeezed onto a large number of useless tokens, leading to policy collapse.

\paragraph{\textcolor{orange!80!black}{Handle the inference-training mismatch.}} In addition to the aforementioned training instability, industrial-scale RL systems impose stringent requirements on training stability and rollout throughput, which often lead to architectural divergence between the training and inference engines. Specifically, high-performance inference servers (e.g., \texttt{SGLang}) and large-scale training frameworks (e.g., \texttt{Megatron-LM}) employ different execution backends, quantization strategies, or batching mechanisms. As a result, the inference policy that generates rollouts, denoted \(\mu^{\text{SGLang}}_{\theta_{old}}\), systematically differs from the training policy \(\pi^{\text{Megatron}}_{\theta_{\text{old}}}\), even when they share the same parameters. The problem is agnostic to the underlying engine and instead arises from the dominant training paradigm commonly adopted in agentic model building. Such a mismatch secretly increases the unstable training risk. Recently, many works have proposed optimization methods \citep{zheng2025group, yao2025offpolicy, Gao2025SoftAP} from the algorithmic level to overcome this challenge. Among them, a widely used mismatch measurement directly quantifies the gap between inference policy and training policy via the token-level different ratio: $\frac{\pi^{\text{megatron}}_{\theta_{old}}(\tau_k)}{\mu^{\text{SGLang}}_{\theta_{old}}(\tau_k)},
$ where $\tau_k$ denotes the $k$-th token in a sequence. Intuitively, we mask out tokens for which the importance weight exceeds the threshold $H$, i.e., those exhibiting severe distributional shift \citep{Zheng2025StabilizingRL}. Specifically, we define a binary loss mask:
$
    m_k = \mathbb{I}\left( \frac{\pi^{\text{megatron}}_{\theta_{old}}(\tau_t \mid \tau_{<t})}{\mu^{\text{SGLang}}_{\theta_{\text{old}}}(\tau_t \mid \tau_{<t})} \leq H \right),$
and exclude masked-out tokens ($m_k = 0$) from gradient updates to ensure training stability. Notably, $m_k$ denotes token-level masking. Finally, the gradient calculation of our baseline with token level mismatch masking is formalized as:
\begin{align}
\label{eq:topr_grad_specific}
\nabla J_{\text{RL}}(\pi) = 
&\underbrace{
    \sum_{\tau \in \mathcal{T}^+} \mu^{\text{SGLang}}_{\theta_{old}}(\tau) R(\tau) 
    \sum_{k=1}^{|\tau|} m_k \nabla \log \pi^{\text{megatron}}_\theta(\tau_k \mid \tau_{<k})
}_{\text{Weighted SL update with token-level masking}} \nonumber\\&+\underbrace{
    \sum_{\tau \in \mathcal{T}^-} \mu^{\text{SGLang}}_{\theta_{old}}(\tau) \left[\rho(\tau)\right]_0^1 R(\tau) 
    \sum_{k=1}^{|\tau|} m_k \nabla \log \pi^{\text{megatron}}_\theta(\tau_k \mid \tau_{<k})
}_{\text{Clipped IS update with token-level masking}} .
\end{align}

 \paragraph{\textcolor{orange!80!black}{Dynamic trajectory filtering for data refinement.}}  
Beyond algorithmic design, we emphasize that data filtering is critical for stable post-training in tool-augmented environments. Empirical analysis reveals that the dominant sources of harmful trajectories stem from environmental noise, including transient API failures, non-deterministic tool responses, and repeated illegal tool invocations. When such trajectories are used, particularly if high-magnitude rewards are spuriously assigned to tokens arising from noisy or invalid interactions, they inject misleading gradient signals that can trigger catastrophic policy collapse. To address this, our RL pipeline incorporates dynamic trajectory filtering during data collection, which explicitly discards trajectories whose rewards are deemed unreliable. Specifically, a trajectory \(\tau\) is rejected if it exhibits any of the aforementioned failure modes. Critically, to ensure stable batch construction and prevent training interruptions due to insufficient valid samples, we employ \textit{on-the-fly resampling}: whenever a rollout is filtered out, the agent immediately initiates a new continuation from the same initial state using the current policy \(\pi_\theta\), with the aim of generating a higher-quality trajectory.

In conclusion, REINFORCE combined with the above-mentioned techniques achieves relatively effective optimization of the model under the agentic RL setting. We take such REINFORCE variant as the improvement frontier of our final \texttt{\textcolor{orange}{IPA}}.

\subsubsubsection{Modeling Multi-Turn Agentic Task as Chunked MDP}\label{sec:MDP}

\begin{figure}[tb!]
    \centering
    \includegraphics[width=0.75\linewidth]{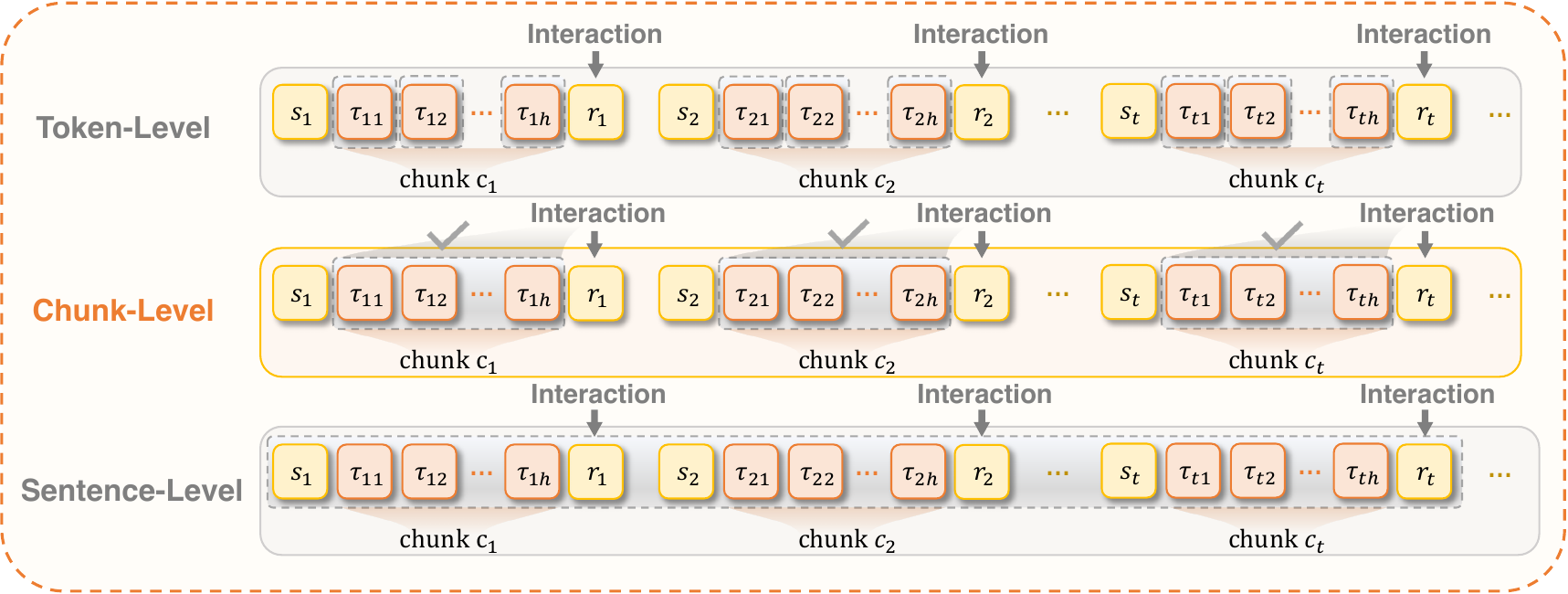}
    % \vspace{-15pt}
    \caption{Comparison of importance sampling strategies across token-level, chunk-level, and sentence-level granularities, where chunk-level aligns with the natural granularity of interactions.}
    \label{fig:is_granularity}
\end{figure}

In \S\autoref{sec:basline}, we established a robust REINFORCE variant as the foundational baseline for agentic reinforcement learning. Building on this, the present section introduces a modeling framework specifically tailored to the challenges of multi-turn agentic interaction, where sparse rewards, long horizons, and tool-mediated reasoning demand more structured credit assignment and stable policy updates. This formulation serves as the basis for a series of subsequent baseline enhancements, paving the way for scalable and reliable RL in complex interactive environments.

Crucially, our MDP operates at the level of \textbf{interaction chunks}, rather than tokens \citep{DAPO} or sentences, to align the horizon with the causal structure of agent--environment interaction naturally provided by \textit{multi-turn tool-integrated reasoning}. Formally, given a token trajectory $\tau_{[1:T]}$, we partition it into a sequence of chunks $\{c_{1}, c_{2}, \dots, c_{K}\}$, $K \ll T$. Each chunk $c_{k}$ spans from one environmental interaction to the next and corresponds to a complete functional unit—typically culminating in a tool invocation (e.g., \textit{reason $\rightarrow$ format API call $\rightarrow$ trigger execution}). Chunk level modeling mitigates mismatches in finer-grained formulations:
\begin{insightblock1}
\begin{enumerate}[leftmargin=1.5em]
    \item Token-level action creates a mismatch between decision granularity and external environmental transition dynamics: the vast majority of tokens have no external effect.
    
    % \item Sentence-level segmentation, though coarser, remains misaligned with agency semantics. In practice, a single tool invocation often requires the model to generate multiple consecutive sentences, e.g., first articulating intent, then constructing parameters, before finally outputting the executable call. Only the final sentence triggers an environmental change.

    \item Sentence-level optimization is overly coarse-grained: a single complete sequence often encompasses multiple rounds of decisions and interactions. Treating these interactions as one monolithic sequence for optimization leads to a waste of fine-grained information.
\end{enumerate}
\end{insightblock1}

Based on chunk level segmentation, our Chunked MDP can be defined by the tuple $(\mathcal{S}, \mathcal{C}, \mathcal{P}, \mathcal{R}, \gamma)$. $\mathcal{S}$ denotes the state space, where each state $s_k \in \mathcal{S}$ encodes the complete interaction history up to the start of chunk $c_{k}$, including prior tool calls, generation and environmental feedback. $\mathcal{C}$ represents the \textit{chunk-action space}: each action $c \in \mathcal{C}$ is a variable-length token sequence generated by the agent in response to $s$, culminating in either a tool invocation or task completion. $\mathcal{P}$ defines the transition dynamics influenced by $c$, governed by the LLM’s generative process and the stochastic responses of external tools. $\mathcal{R}$ is a sparse reward function that only provides positive feedback when the trajectory has passed all unit tests. $\gamma \in (0,1]$ is the discount factor, applied at the chunk level to prioritize temporally proximal, outcome-influencing decisions.

Overall, Chunked MDP aggregates those tokens that collectively lead to an environmental transition, aligns the optimization horizon with meaningful interventions, and enables accurate credit assignment.
\subsubsubsection{Reconstruct Training Objective via Chunk-Level Optimization}\label{sec:loss}
To align with the Chunked MDP, \texttt{\textcolor{orange}{IPA}} adjusts the optimization horizon of the constructed baseline to the chunk level by incorporating return calculation, importance sampling, and mismatch masking. Intuitively, these refinements intermediate granularity strikes a favorable balance: it is coarse enough to ensure training efficiency and semantic consistency within each chunk, yet fine-grained enough to enable precise credit assignment across multi-turn reasoning. 

First, we introduce a \textbf{Chunk-Level Discounted Return}, which re-establishes temporal credit assignment in agentic reinforcement learning. A key limitation of conventional token-level formulation is its inability to incorporate meaningful temporal discounting  \citep{wang2025beyond}, since applying a reward discount factor $\gamma < 1$ over thousands of tokens would cause reward signals to vanish exponentially~\citep{yue2025vapo}. Moreover, without temporal structure, value estimates for early states suffer from high variance in long tail trajectories  \citep{yin2024analyzing, amit2020discount}. In contrast, the Chunked MDP formulation discretizes trajectories at the semantic action boundary, which enables the principled reintroduction of temporal discounting at the chunk level. Formally, given a trajectory partitioned into $K$ chunks, the return assigned to chunk $c_{k}$ is defined as:
\begin{equation}\label{eq:reinforce_chunk_grad}
    G_k = \gamma^{\Delta(j,k)} \times R_{\text{final}},
\end{equation}
where $\Delta(j,k)$ denotes the number of chunks between $c_{k}$ and $c_{j}$, and $R_{\text{final}}$ is the terminal task reward. All tokens within chunk $c_k$ share the same scalar weight $G_k$ in the policy gradient. Notably, this reward calculation can be compatible with intrinsic reward systems.

\begin{figure}[tb!]
    \centering
    \begin{subfigure}[tb!]{0.3\textwidth}
        \centering
        \includegraphics[width=1\linewidth]{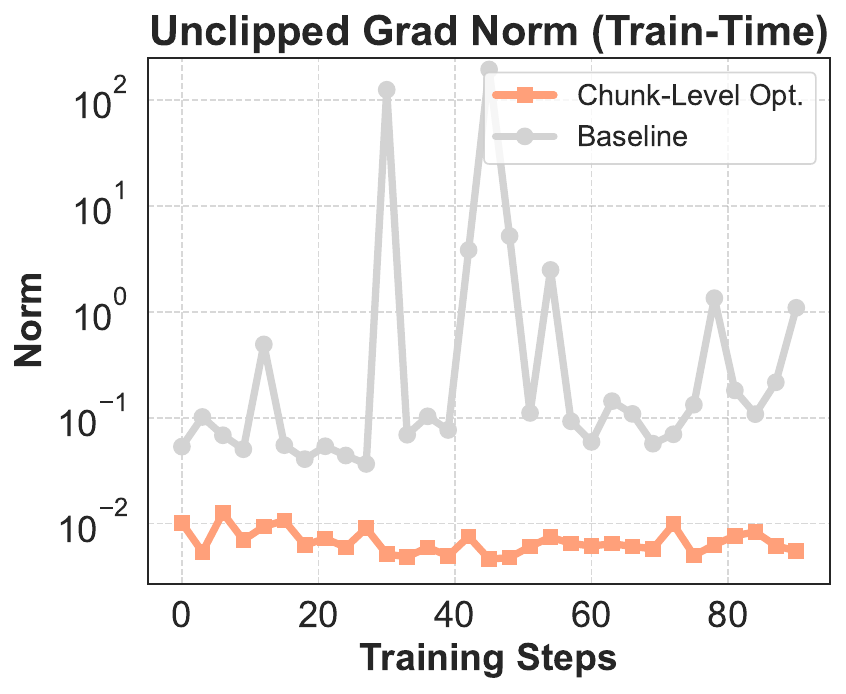}
        \label{fig:rl-chunk-gradnorm}
    \end{subfigure}
    \hfill
    \begin{subfigure}[tb!]{0.3\textwidth}
        \centering
        \includegraphics[width=1\linewidth]{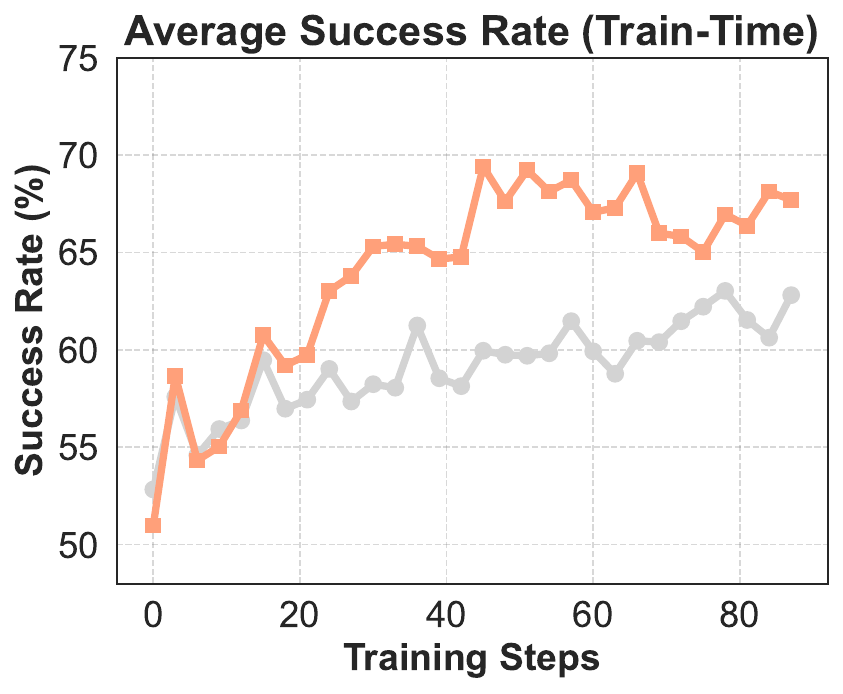}
        \label{fig:rl-chunk-trainscore}
    \end{subfigure}
    \hfill
    \begin{subfigure}[tb!]{0.3\textwidth}
        \centering
        \includegraphics[width=1\linewidth]{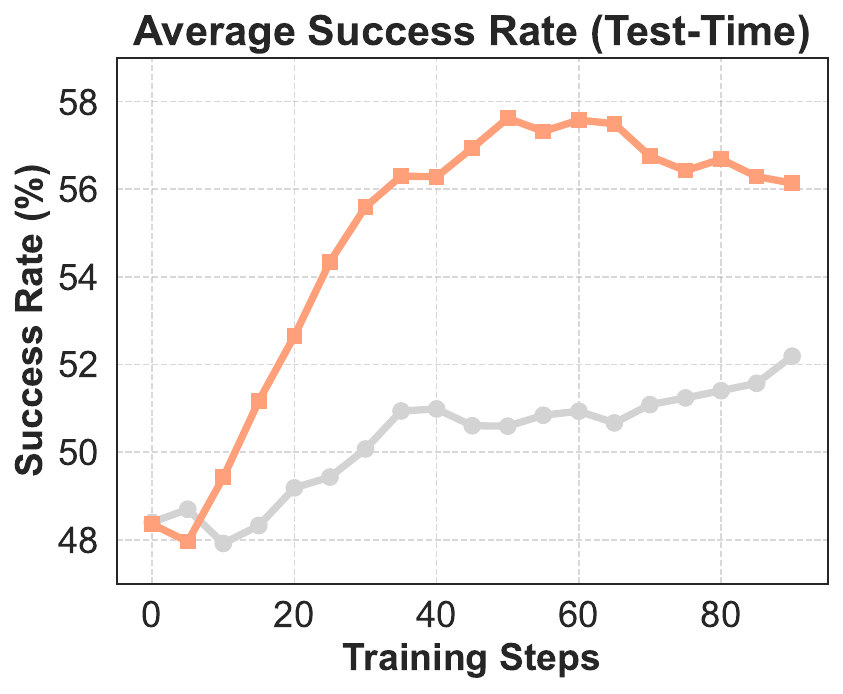}
        \label{fig:rl-chunk-score}
    \end{subfigure}
    \caption{Comparison of Chunk-Level Optimization and baseline on a mini-set of the training data. \textbf{Left:} Unclipped gradient norm for updates that reflects the stability of training. Our Chunk-Level Optimization exhibits more stable gradient norms, while baseline induces anomalous gradient fluctuations. \textbf{Middle:} Performance on training tasks. Owing to stable gradient updates and effective credit assignment, Chunk-Level Optimization consistently shows better performance than baseline. \textbf{Right:} Test-time success rate on validation tasks. Chunk-Level Optimization retain its superiority over baseline, demonstrating the generalization of our method.}
    \label{fig:rl-chunk}
\end{figure}
This design yields two crucial benefits. First, by aligning discounting with semantic decision intervals, it mitigates the bias-variance trade-off in long-horizon credit assignment: early chunks are downweighted not arbitrarily, but proportionally to their temporal distance from outcome-determining actions, thereby reducing noise propagation while preserving signal integrity. Second, it avoids the exponential signal decay inherent in token-level discounting, since $K \ll T_{\text{tokens}}$, the effective horizon is drastically shortened, ensuring stable gradient magnitudes even in multi-thousand-token trajectories. Consequently, the policy receives stronger gradients for chunks proximate to task success ($\gamma^{\Delta} \approx 1$), while early ineffective attempts, e.g., invalid tool calls, are exponentially suppressed. This not only accelerates convergence on high-impact behaviors but also induces an implicit trajectory compression effect, significantly improving sample efficiency and training stability. Empirically, the results in \autoref{fig:rl-chunk} (Left) indicate that incorporating chunk-level discounted returns into the gradient computation of our baseline enhances training stability, accelerates the perception and learning of high-level action semantics embedded in chunks, and significantly improves the model's optimization efficiency (\autoref{fig:rl-chunk} (Middle)). This, in turn, leads to improved performance on difficult tasks (\autoref{fig:rl-chunk} (Right)).

Moreover, we propose \textbf{Chunk-Level Importance Sampling} to synergize with the chunk-level return as suggested in \citet{zheng2025group}. Specifically, for each interaction chunk \(c\), we calculate the importance sampling ratio over all tokens within the chunk to measure the chunk level difference. Notably, because chunk-level calculation expands its calculation horizon compared to token-level ratios, we use the geometric mean style IS to dampen the impact of outlier tokens and avoid extreme ratios:
\begin{align}
\rho_c (c) = \bigg(\prod_{t \in c} \frac{\pi^{\text{megatron}}_\theta(\tau_t \mid \tau_{<t})}{\pi^{\text{megatron}}_{\theta_{\text{old}}}(\tau_t \mid \tau_{<t})}\bigg)^{\frac{1}{|c|}}.
\end{align}\

Finally, to align all the optimization scales with the chunked MDP, we finally elevate loss masking from the token to the interaction chunk level: \(m_c = \mathbb{I}\big(( \prod_{t \in c} \frac{\pi^{\text{megatron}}_{\theta_{old}}(\tau_t \mid \tau_{<t})}{\mu^{\text{SGLang}}_{\theta_{old}}(\tau_t \mid \tau_{<t})})^{\frac{1}{|c|}} \leq H \big)\). Intuitively, \textbf{Chunk-level masking} may simultaneously mitigate two critical issues that arise at the token level~\citep{liu-li-2025-rl-collapse}:  
\begin{insightblock1}
\begin{enumerate}[leftmargin=1.5em]
\item \textit{State Occupancy Mismatch}: token-level policy gradients are computed over state distributions induced by the inference policy, which diverges from the true state visitation. 
\item \textit{Mismatched Reward Signal}: fine-grained token-wise importance weights are misaligned with the coarse, outcome-driven rewards that govern long-horizon agentic success.
\end{enumerate}
\end{insightblock1}

Empirical experience also shows that the constraint of mask is relaxed by extending to chunk horizon, so as to avoid excessive influence on RL gradient and maintain training stability. Combining chunk-level masking, discounted returns and importance sampling, the gradient calculation of our REINFORCE variant can be reformulated as:
\begin{align}
\label{eq:aicpo_topr_final}
\nabla J_{\textcolor{orange}{\text{Chunk-RL}}}(\pi) = 
&\underbrace{
    \sum_{\textcolor{orange}{c \in \mathcal{T}^+}} \mu^{\text{SGLang}}_{\theta_{old}}(\textcolor{orange}{c}) \textcolor{orange}{G_c} 
    \sum_{k=1}^{|\textcolor{orange}{c}|} \textcolor{orange}{m_c} \nabla \log \pi^{\text{megatron}}_\theta(\textcolor{orange}{c_{k}} \mid \textcolor{orange}{\tau_{<c_{k}}})
}_{\text{Chunk-level weighted SL update}} \nonumber\\&+\underbrace{
    \sum_{\textcolor{orange}{c \in \mathcal{T}^-}} \mu^{\text{SGLang}}_{\theta_{old}}(\textcolor{orange}{c}) \left[\textcolor{orange}{\rho_c(c)}\right]_0^1 \textcolor{orange}{G_c} 
    \sum_{k=1}^{|\textcolor{orange}{c}|} \textcolor{orange}{m_c} \nabla \log \pi^{\text{megatron}}_\theta(\textcolor{orange}{c_{k}} \mid \textcolor{orange}{\tau_{<c_{k}}})
}_{\text{Chunk-level clipped IS update}} ,
\end{align}
where $k$ denotes the k-th chunk in $\tau$, \(G_c\) is the discounted return of chunk \(c\) (as defined in \autoref{eq:reinforce_chunk_grad}). 

\begin{figure}[tb!]
    \centering
    \includegraphics[width=1\linewidth]{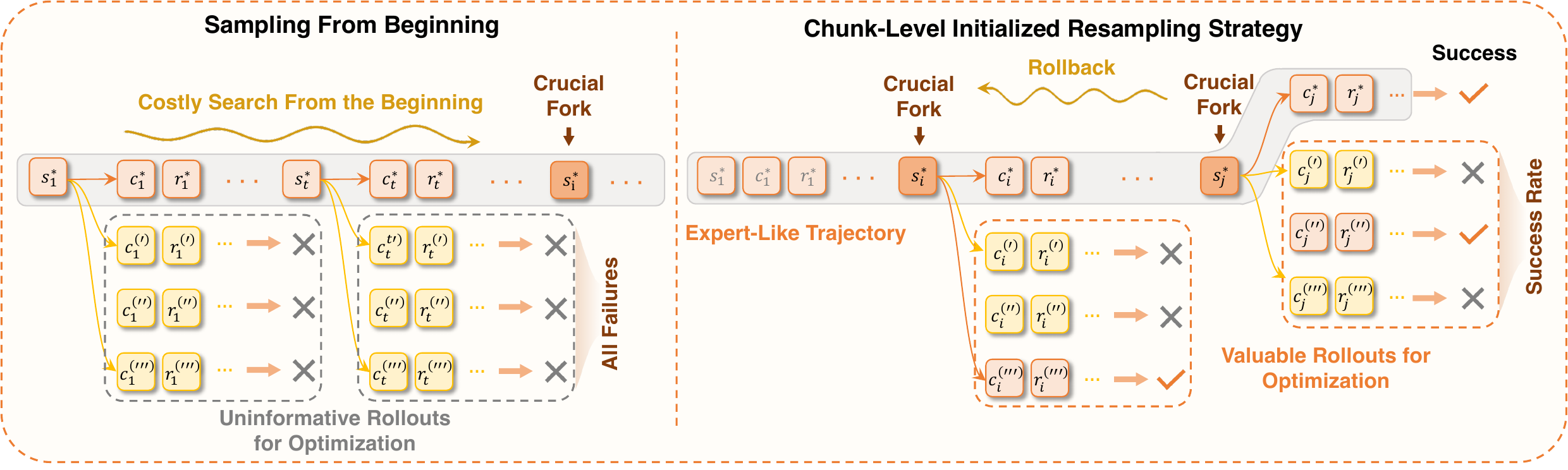}
    % \vspace{-15pt}
    \caption{Illustration of the Chunk-Level Initialized Resampling Strategy (Sequential Rollback). \textbf{Left}: In challenging tasks, sampling high-quality trajectories from the beginning is difficult, severely limiting policy learning efficiency. \textbf{Right}: Sequential Rollback sampling strategy initiates rollouts from critical chunks, dramatically reducing the exploration burden and enabling the policy to rapidly acquire the key skills embedded in these crucial chunks. By progressively rolling back along the crucial chunks, it enables chunk-level curriculum learning for model to finally solve these challenging tasks.}
    \label{fig:chunk-sampling}
\end{figure}

\subsubsubsection{Rollout Paradigm Refinement via Chunk-Level Initialized Resampling}\label{sec:sample}

As agentic reasoning evolves from single-turn inference to multi-turn interactions, we observe that the probability of sampling a positive trajectory markedly decays on several complex tasks. After analyzing these failed trajectories, we find that the success rate of these long-horizon agentic tasks is typically governed by a sparse set of \textit{crucial forks}--decision points where the model's next chunk disproportionately affects the final return (e.g., selecting the right tool or correctly parsing a pivotal observation). When sampling from the initial state, an incorrect decision chunk at any crucial fork will possibly cause the failure of the entire task. Therefore, under a naive sampling strategy, rollouts on these tasks always contain extremely sparse positive signals, resulting in inefficient or misleading policy updates \citep{DAPO}. A simple but exciting insight is that, if we can prefill the interaction history with the correct expert-like chunks and resample the subsequent trajectories, we can effectively reduce task difficulty and enrich the reward signals for optimization. Once the model has learned the tail part chunks, we roll back to the head part crucial forks, enabling chunk-level curriculum learning on these challenging tasks.

Specifically, \texttt{\textcolor{orange}{IPA}} introduces \textbf{Chunk-Level Initialized Resampling}, which enables the policy to launch rollouts from selected forks by initializing tasks with chunks of expert-like trajectories, e.g., obtained either via self-sampling or from a teacher model. Notably, we periodically update the expert trajectory under the current policy to maximize coverage of critical chunks while minimizing interference from unnecessary ones. Formally, given an expert-like trajectory with $K$ chunks \({\tau}^{*} = ({c}^{*}_{1}, {c}^{*}_{2}, \dots, {c}^{*}_{K})\) and an selected expert chunk \({c}^{*}_{k}\), we interact with the environment using \({\tau}^{*}_{\leq c^*_{k-1}}\) and then resample the subsequent chunks \(\tau_{\geq c_k}\) with the train policy \(\pi_{\theta}\). The expected success rate of resampling trajectories on \({\tau}^{*}_{\leq c^*_{k-1}}\) is defined as \(\mathbb{E}_{\tau \sim \pi_\theta} R_{final}({\tau}^{*}_{\leq c^*_{k-1}}, {\tau}_{\geq c_k})\). We then define chunk \({c}^{*}_{f}\) as a crucial chunk if the expected resampling success rate on \({\tau}^{*}_{\leq c^*_{f-1}}\) is significantly lower than on \({\tau}^{*}_{\leq {c^*_f}}\). The drop in success rate indicates that the decisions made within \({c}^{*}_{f}\) are decisive for success, and the current policy does not master such skills. Therefore, the state right before \({c}^{*}_{f}\) is naturally a crucial fork.

\begin{figure}[tb!]
    \centering
    \begin{subfigure}[tb!]{0.3\textwidth}
        \centering
        \includegraphics[width=1\linewidth]{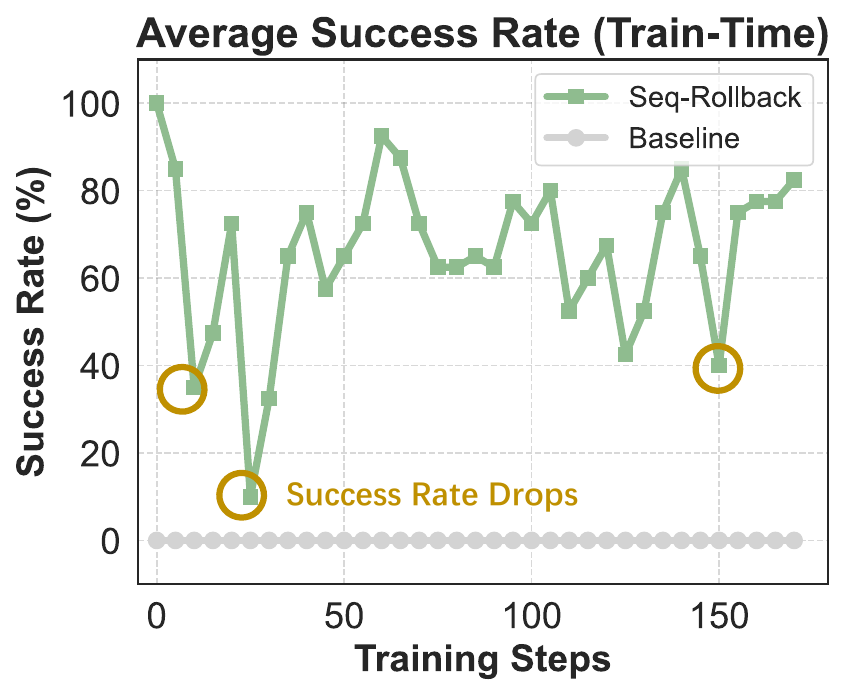}
        \label{fig:rl-case-trainscore}
    \end{subfigure}
    \hfill
    \begin{subfigure}[tb!]{0.3\textwidth}
        \centering
        \includegraphics[width=1\linewidth]{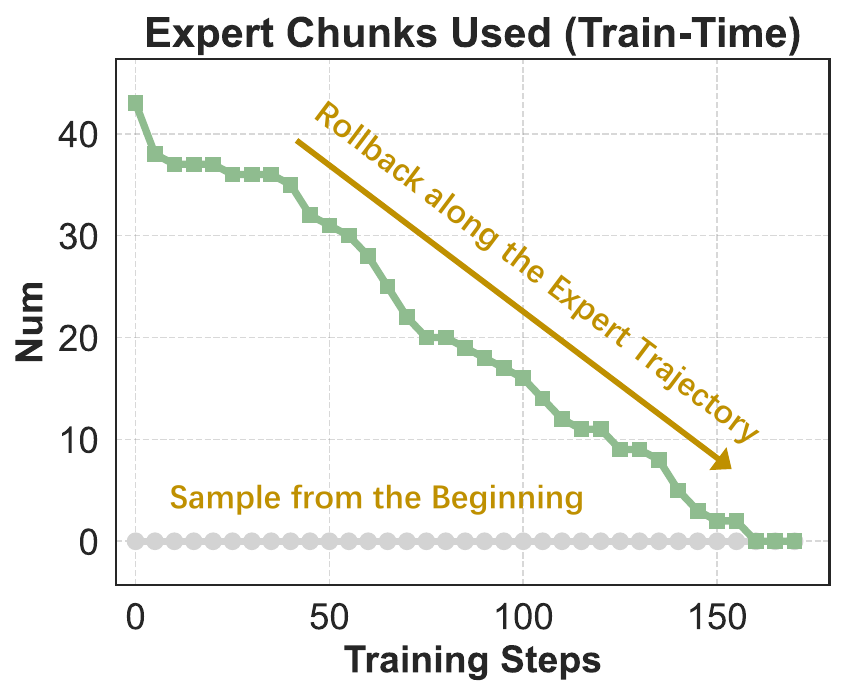}
        \label{fig:rl-case-rollback}
    \end{subfigure}
    \hfill
    \begin{subfigure}[tb!]{0.3\textwidth}
        \centering
        \includegraphics[width=1\linewidth]{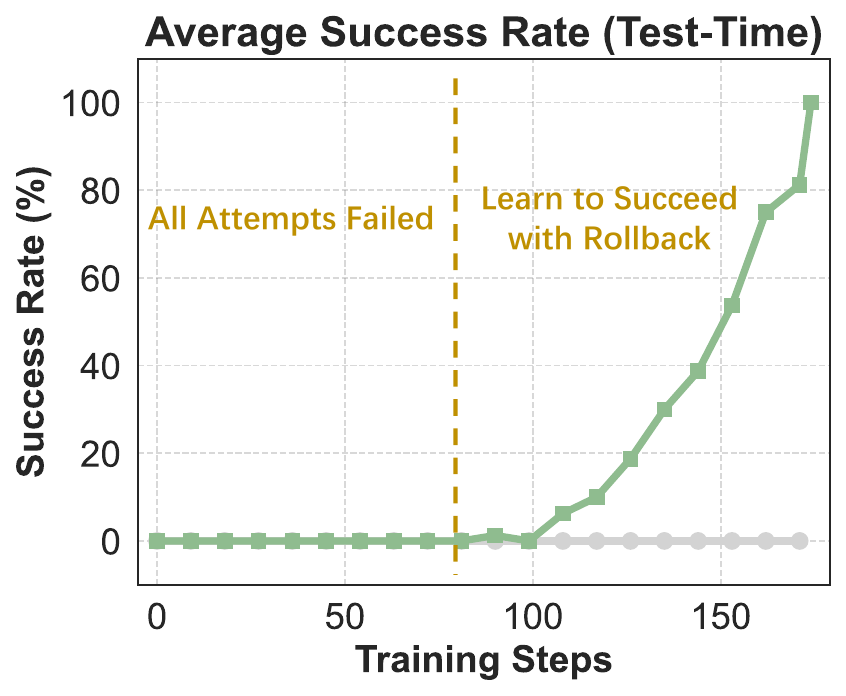}
        \label{fig:rl-case-score}
    \end{subfigure}
    \caption{Performance of Sequential Rollback and baseline (naive sampling) on a challenging training task. \textbf{Left:} Average success rate during training, which reflects the percentage of positive signals in training batch. Sequential Rollback obviously brings more valuable rollouts compared to baseline (all failures). The drop of success rate indicates that the model has rolled back across a crucial chunk to the crucial fork. \textbf{Middle:} Expert chunks used during training, which visually displays the progress of rolling back along the expert trajectory. \textbf{Right:} Average success rate on the challenging task during testing. In test-time, all trajectories are sampled from the initial state. The gap between two curves after step 75 indicates that sequential rollback enables effective learning on extremely hard tasks.}
    \label{fig:rl-case}
\end{figure}

Empirically, a naive yet effective strategy to select the resampling initialization state is \textit{Sequential Rollback}: starting from the last chunk of an expert trajectory and moving regressively toward the beginning. As shown in \autoref{fig:chunk-sampling}, sampling from states near the end of a successful trajectory requires far fewer rollout turns, dramatically reducing the exploration burden compared to starting from the initial state. Consequently, positive samples are much easier to obtain from these tail states, enabling reliable generation of high-quality rollouts and rapid learning correct behaviors on these crucial forks. The results in \autoref{fig:rl-case} (left) demonstrate that \textit{Sequential Rollback} can keenly monitor the important forks on expert trajectory, and gradually master the global crucial chunks through progressive learning (\autoref{fig:rl-case} (middle)), so that the policy can obtain excellent test performance on the difficult task (\autoref{fig:rl-case} (right)).

However, the aforementioned \textit{Sequential Rollback}, while effective at preserving high-value reasoning pathways, suffers from significant computational inefficiency. Crucially, if the decisive interaction occurs early in the trajectory, backward scanning only discovers it after exhaustively testing all later positions, leading to wasted rollouts and poor scalability across diverse task structures. To enable robust and efficient detection of critical modules across a wide range of tasks, we propose a \textit{Parallelized Initialization} scheme as a practical and reliable compromise. Specifically, given an expert-like trajectory, we first select a set of anchor chunks at various positions (uniformly or randomly), aiming to include crucial forks between these anchors. Then \texttt{\textcolor{orange}{IPA}} initializes environments to the state asociate with the anchor chunks and launches several independent rollouts in parallel. \textit{Parallelized Initialization} introduces trajectories rolled out from diverse starting states within a single rollout batch. Although this dilutes the number of samples drawn at each potential crucial fork, it avoids the time cost on bad-cases of \textit{Sequential Rollback} and ultimately achieves higher efficiency on our dataset.

% \begin{figure}[tb!]
%     \centering
%     \includegraphics[width=0.6\linewidth]{figures/RL_ablation_scores_curves.pdf}
%     \caption{Ablation study on a mini-set of training data}
%     \label{fig:rl-hack-5-validation}
% \end{figure}

Finally, even with \textit{Parallelized Initialization}, there exist extreme cases where no positive trajectories are sampled from a crucial fork. In such scenarios, purely on-policy or importance-sampled updates yield zero gradient signal, stalling learning and risking irreversible policy collapse. To accelerate convergence and safeguard against degradation, we adopt a hybrid training objective that seamlessly integrates imitation learning (IL) and reinforcement learning (RL). Specifically, to avoid missing positive signals at any crucial fork, we introduce the imitation learning target to the expert’s chunks \({\tau}^{*}_{\leq c^*_f}\) as a fallback. This injects a “recovery signal” that anchors the policy in high-quality regions of the behavior space, preventing drift into degenerate modes. Formally, our mixed objective operates in two phases:
\begin{insightblock1}
\begin{enumerate}[leftmargin=1.5em]
    \item For the \textit{prefilled expert chunks} ${\tau}^{*}_{\leq c^*_{f-1}}$ and \textit{expert cruical chunk} ${c}^{*}_{f}$, we apply imitation learning style loss. This rapidly instills reliable subroutines, e.g., tool invocation formatting.
    \item For the \textit{resampled chunks} $\tau_{\geq {c_f}}$, we use the chunk-level RL (\autoref{eq:aicpo_topr_final}), enabling adaptive credit assignment on outcome-determining interactions.
\end{enumerate}
\end{insightblock1}

\begin{figure}[tb!]
    \centering
    \begin{subfigure}[tb!]{0.3\textwidth}
        \centering
        \includegraphics[width=1\linewidth]{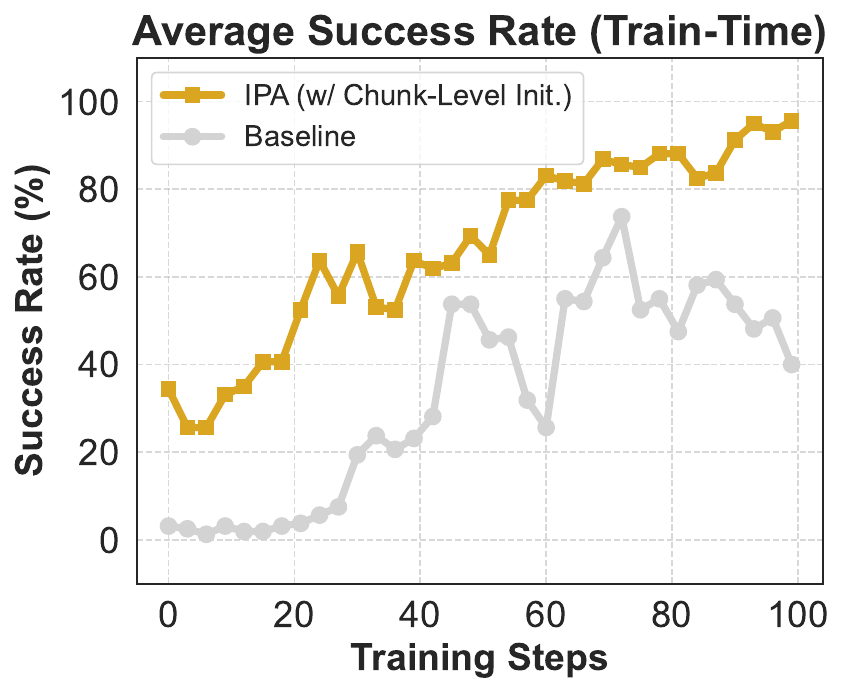}
        \label{fig:rl-ablation-train}
    \end{subfigure}
    \begin{subfigure}[tb!]{0.3\textwidth}
        \centering
        \includegraphics[width=1\linewidth]{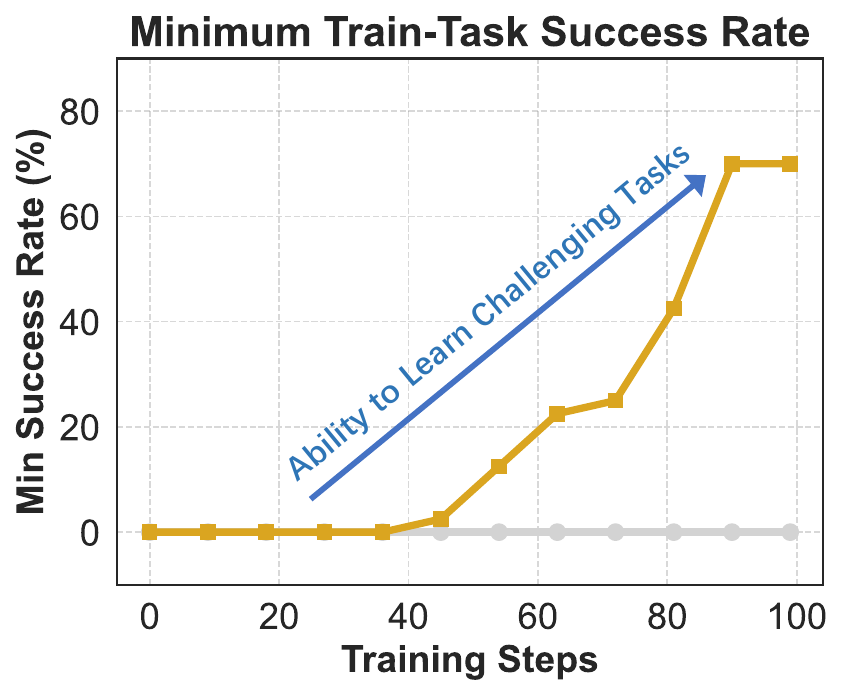}
        \label{fig:rl-ablation-minimum}
    \end{subfigure}
    \begin{subfigure}[tb!]{0.3\textwidth}
        \centering
        \includegraphics[width=1\linewidth]{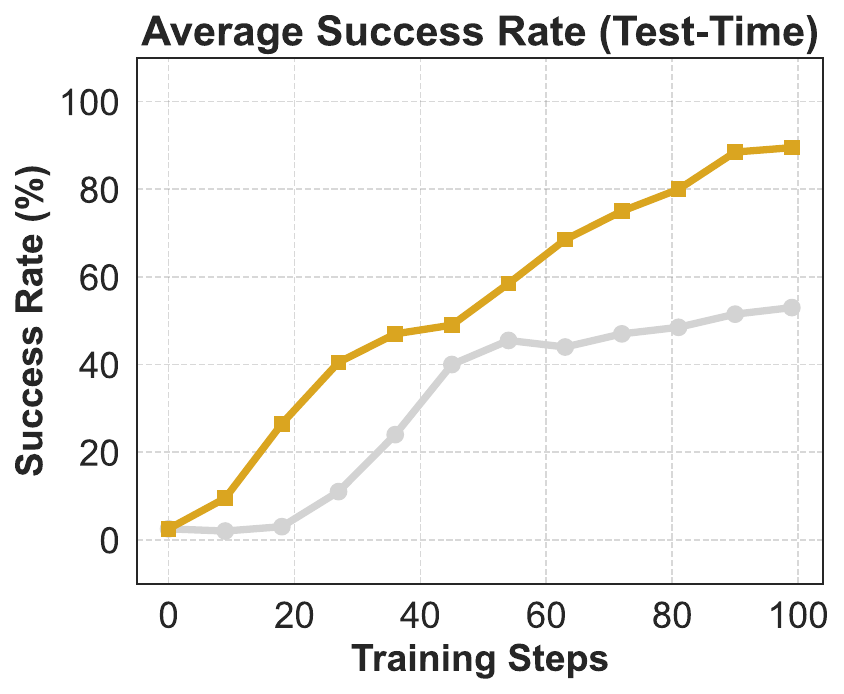}
        \label{fig:rl-ablation-score}
    \end{subfigure}
    \caption{Comparison of \texttt{\textcolor{orange}{IPA}} with \& without Chunk-Level Initialized Resampling (Parallelized Initialization) on a mini-set of the training data. \textbf{Left:} Average success rate on training tasks. The gap between curves in the early stage of training shows that the Chunk-Level Initialized Resampling brings much more diverse reward signals in training batches. \textbf{Middle:} Minimum success rate across train-tasks with test-time setting (sampled from beginning). With Chunk-Level Initialized Resampling, IPA enables the train model to solve extremely hard tasks by learning in a chunk-level curriculum-like manner. \textbf{Right:} Average success rate at test-time. Benefiting from more valuable rollouts, Parallelized Initialization substantially improves the performance of \texttt{\textcolor{orange}{IPA}}.}
    \label{fig:rl-ablation}
\end{figure}

The final training loss of \texttt{\textcolor{orange}{IPA}} is thus:
\begin{equation}
\mathcal{L}_{\text{\texttt{\textcolor{orange}{IPA}}}} = \lambda_{\text{IL}} \cdot 
\underbrace{
    \sum_{{c}^{*}_{k}\in{\tau}^{*}_{\leq c^*_{f}}} \pi^{megatron}_\theta({c}^{*}_{k}) G_{c^*_k} \nabla \log \pi^{megatron}_\theta({c}^{*}_{k} \mid {\tau}^{*}_{\leq {c}^{*}_{k-1}})
}_{\text{Imitation learning style update}} + \lambda_{\text{RL}} \cdot \mathcal{L}^{c \in \tau_{\geq {{c}_{f}}}}_{\text{\textcolor{orange}{Chunk-RL}}}.
\end{equation}
The coefficients $\lambda_{\text{IL}}, \lambda_{\text{RL}}$ balance imitation and exploration. The results in \autoref{fig:rl-ablation} demonstrate that \texttt{\textcolor{orange}{IPA}} effectively enhances the model's generalization ability on challenging agentic tasks, enabling it to overcome performance limits and significantly improve learning efficiency. Based on our \textcolor{orange}{\texttt{IPA}}, we effectively unlock the agentic capabilities of \textcolor{orange!80!black}{\textbf{ROME}}, a 30B MoE model, allowing it to overcome the performance bottleneck associated with its inherent size and achieve capabilities comparable to those of larger models, such as the 480B agentic model.

\subsection{Experiments and Benchmark}
\subsubsection{Evaluation Setup}

To rigorously and holistically evaluate agentic intelligence, we adopt a three-dimensional evaluation framework encompassing \textit{tool-use abilities}, \textit{general agentic capabilities}, and \textit{terminal-based agentic execution}. These dimensions reflect the core competencies required for real-world agent deployment, ranging from tool calling to long-horizon, environment-grounded task completion.

%We first assess tool selection and calling on a set of tool-use benchmarks, including domain-specific subsets of TAU2-Bench (Retail, Airline, Telecom)~\citep{barres2025tau2}, BFCL-V3~\citep{patilberkeley}, and MTU-Bench~\citep{wang2024mtu}.
\begin{itemize}[leftmargin=1.5em]
\item \textbf{Tool-use Abilities.} 
We evaluate tool-use abilities by assessing whether agents can correctly select, invoke, and coordinate external tools in response to user intents. This dimension is evaluated using established tool-use benchmarks, including domain-specific subsets of TAU2-Bench (Retail, Airline, Telecom)~\citep{barres2025tau2}, BFCL-V3~\citep{patilberkeley}, and MTU-Bench~\citep{wang2024mtu}. 

%More importantly, to analyze whether the model can capture the complexities of real-world dynamics and achieve efficient interacting, we evaluate general agentic capabilities on a diverse suite of benchmarks: BrowseComp-ZH~\citep{zhou2025browsecomp}, ShopAgent~\citep{wang2025shopsimulator}, and GAIA~\citep{mialon2023gaia}. These tasks measure an agent's ability to perceive, plan, and act autonomously in open-ended, interactive environments, which go far beyond static question answering.

\item \textbf{General Agentic Capabilities.}  
To evaluate an agent’s ability to solve complex queries through multi-step reasoning and high-level decision-making, we assess general agentic capabilities on a diverse suite of benchmarks, including BrowseComp-ZH~\citep{zhou2025browsecomp}, ShopAgent~\citep{wang2025shopsimulator}, and GAIA~\citep{mialon2023gaia}. 

\item \textbf{Terminal-Based Agentic Execution.}  
%Most significantly, we assess agentic coding proficiency using Terminal-Bench~\citep{tbench_2025}, Terminal-Bench 2.0, SWE-Bench Verified~\citep{swe_verified}, SWE-Bench Multilingual~\citep{yang2025swemultilin} and our extended benchmark, to characterize autonomous software engineering in realistic development workflows, including environment setup, iterative debugging, and system-level integration. These difficult tasks serve as the critical dimension for evaluating a general agent's robustness.
We further assess terminal-based agentic execution using benchmarks that require agents to complete goal-directed workflows within executable environments. Specifically, we evaluate on Terminal-Bench 1.0~\citep{tbench_2025}, Terminal-Bench 2.0, SWE-bench Verified~\citep{swe_verified}, SWE-Bench Multilingual~\citep{yang2025swemultilin}, as well as our extended benchmark.
 \end{itemize}
 
Together, these three dimensions represent demanding and practically significant dimensions of agent evaluation, serving as the primary yardsticks for assessing real-world deployability of agentic models.

Meanwhile, we observe that the aforementioned publicly available datasets exhibit notable limitations in scale, domain balance, difficulty calibration, and contamination control. To further enrich the agentic evaluation ecosystem, we introduce \textbf{Terminal-Bench Pro}, a rigorously curated benchmark designed to offer larger-scale coverage, balanced task domains, calibrated difficulty levels, and stronger safeguards against data leakage. Full details of its construction and corresponding evaluation results are provided in Section~\ref{subsubsec:terminal_bench_pro}.

All models are evaluated using a consistent set of generation hyperparameters to ensure fair comparison. Specifically, we configure the models with \texttt{temperature = 0.7}, \texttt{top-p = 0.8}, and \texttt{top-k = 20}. The maximum output length is restricted to 65,536 tokens and the maximum context length 262,144 tokens.
To ensure consistency across terminal-based agentic tasks, all CLI evaluations are conducted under a unified execution environment using the \textit{iFlow CLI} framework. For the evaluation results, we report Pass@1 as the average over 3 independent runs (Avg@3), and we use * to denote scores obtained from official reports or public leaderboards.
% \newpage

\subsubsection{Terminal Bench Pro: A More Rigorous and Fine-Grained Benchmark for Terminal Agents}
\label{subsubsec:terminal_bench_pro}

\begin{figure}[!h]
      \centering
    \includegraphics[width=0.9\linewidth]{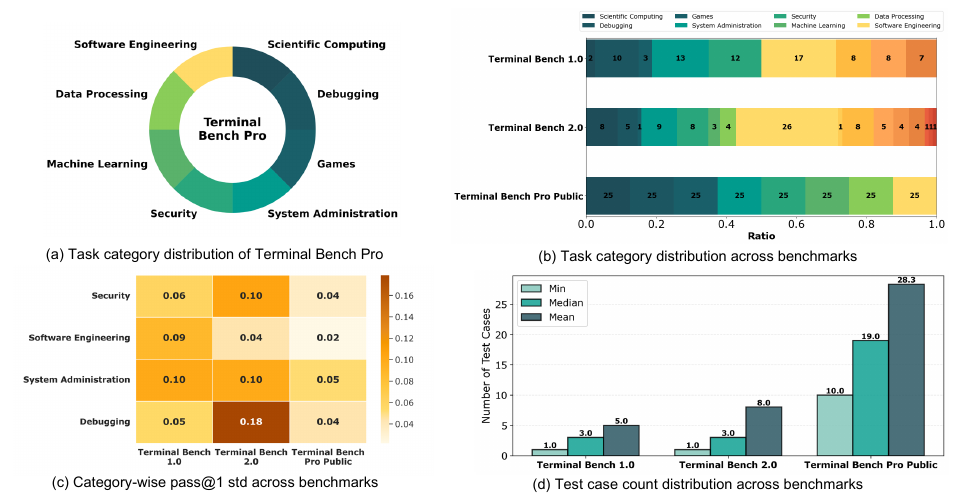}
    \caption{Benchmark characterization and cross-benchmark comparison of Terminal Bench Pro against other benchmarks.}
    \label{fig:dataset_comparison}
\end{figure}
%  (a) Task-category distribution of Terminal Bench Pro. (b) Category proportion comparison across benchmarks. (c) Category-wise standard deviation of pass@1 across benchmarks. (d) Per-instance test-case count statistics (min/median/mean) across benchmarks.

\paragraph{\textcolor{orange!80!black}{Motivation and Limitations of Existing Benchmarks.}} Terminal-based benchmarks are increasingly important for evaluating autonomous coding agents, yet existing benchmarks remain limited in scale, reliability, and diagnostic resolution. In terms of scale, Terminal Bench 1.0 contains only 80 tasks, and Terminal Bench 2.0 expands this marginally to 89 tasks, which renders aggregate metrics susceptible to wide confidence intervals and makes overall rankings sensitive to a small number of idiosyncratic instances. Moreover, the benchmark reliability can be further compromised by task-specific artifacts. For instance, tasks that are highly sensitive to external network conditions (e.g., downloading content from online platforms) introduce environment-induced variance that is orthogonal to agent capability, thereby reducing reproducibility and complicating attribution of observed performance differences. More critically, existing benchmarks lack sufficient granularity for domain-level analysis. As shown in Fig \ref{fig:dataset_comparison}(b), several sub-domains are represented by only one to three tasks (e.g., one task in \textit{games} and three tasks in \textit{machine learning}). This sparsity yields unstable sub-domain level estimates, as reflected by the substantial category-wise variance in pass@1 across benchmarks as shown in Fig \ref{fig:dataset_comparison}(c). And this undermines the statistical significance and confidence of our evaluation results in these domains. In addition, many existing tasks are overly broad yet relying on sparse test suites, resulting in low test coverage, as depicted by the per-instance test-case statistics in Fig \ref{fig:dataset_comparison}(d). Under such conditions, agents may pass evaluations by exploiting underspecified requirements or unintended shortcuts, which undermines the validity of conclusions regarding correctness, robustness, and generalization.

\paragraph{\textcolor{orange!80!black}{Design and Construction.}} To address these limitations, we propose Terminal Bench Pro, a new benchmark designed for rigorous and fine-grained evaluation of terminal-based agents. Its construction follows three core principles: 
\begin{insightblock1}
\begin{enumerate}[leftmargin=1.5em]
    \item \textbf{Comprehensive Domain Coverage:} Terminal Bench Pro is aligned with real-world user demands. It systematically covers eight major CLI-related domains with a balanced number of tasks in each, enabling reliable assessment of fine-grained domain capabilities.
    \item \textbf{Rigorous Data Validation:} All tasks undergo multiple rounds of expert validation to ensure high data quality, unambiguous specifications, and comprehensive test coverage.
    \item \textbf{Deterministic Evaluation Environment:} Every instance is paired with an executable test file and a fully reproducible environment, eliminating non-determinism arising from external network or system dependencies.
\end{enumerate}
\end{insightblock1}

To ground the benchmark in practical usage, we analyze discussions from GitHub issue forums and identify eight key domains where user queries are predominantly concentrated: data processing, games, debugging, system administration, scientific computing, software engineering, machine learning, and security. For each domain, we engage experts to manually construct tasks based on real-world problem scenarios. All problem descriptions and unit tests are authored from scratch by experienced programmers to ensure originality and minimize the risk of data leakage. Each task then undergoes independent review by multiple experts to eliminate ambiguous instructions and false-positive solutions, ensure optimal reference solutions, and achieve high unit-test coverage.

Following this process, we construct \textbf{Terminal Bench Pro} \footnote{\url{https://github.com/alibaba/terminal-bench-pro}} dataset, which consists of 400 evaluation tasks (200 public and 200 private instances) uniformly distributed across the eight domains. As evidenced by Fig.~\ref{fig:dataset_comparison}, the benchmark exhibits high test coverage, rich task diversity, and low evaluation variance, providing a reliable foundation for systematic and trustworthy assessment of terminal-based agentic systems.

\subsubsection{Evaluation Results}
\label{subsubsec:evaluation_results}

\begin{figure}[!h]
      \centering
    \includegraphics[width=\linewidth]{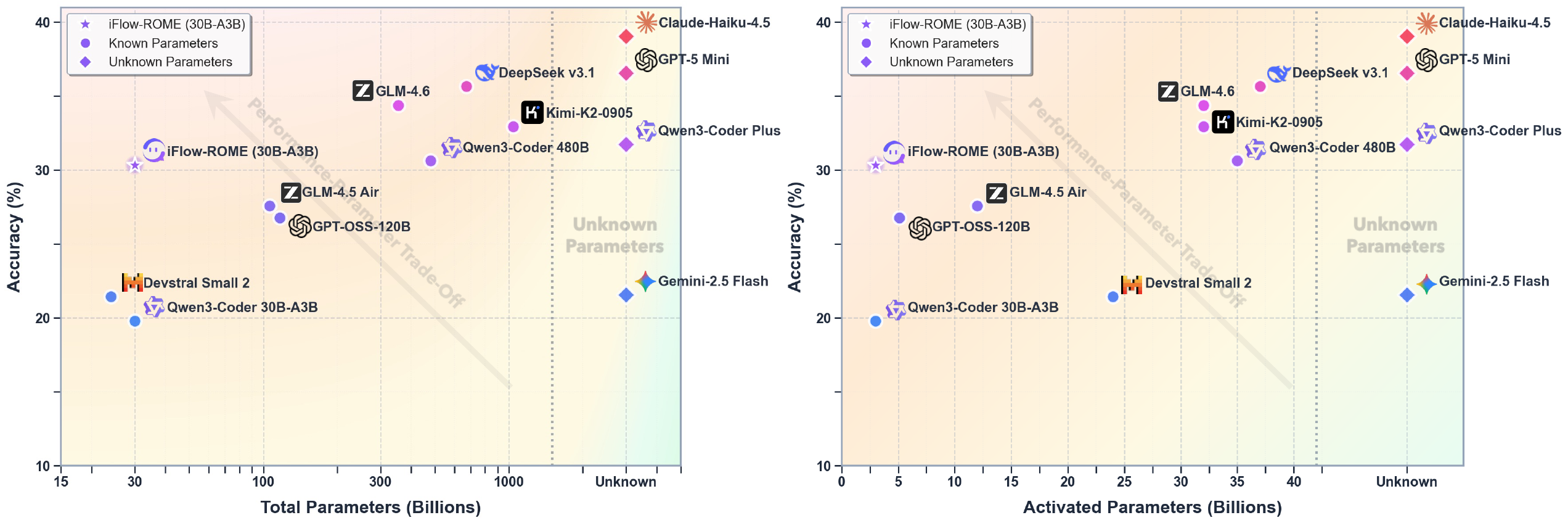}
    \caption{Performance-parameter trade-offs in agentic tasks. Scores represent averages on general agentic and code agent benchmarks. Models with known parameters are shown as circles, while proprietary models with unknown parameters are depicted as diamonds (right side). \textbf{Left}: Total parameters versus overall performance. \textbf{Right}: Activated parameters versus overall performance.}
    \label{fig:performance-size}
\end{figure}

% 整体结果
In this section, we present the detailed and fair evaluation results comprehensively under a structured agentic evaluation setting, to support the outstanding performance of our agentic model, i.e., \textbf{\textcolor{orange!80!black}{ROME}}, trained by \texttt{\textcolor{orange}{ALE}} through three test perspectives. To provide an intuitive overview, \autoref{fig:performance-size} highlights \textbf{\textcolor{orange!80!black}{ROME}}'s agentic performance advantage under comparable or smaller parameter budgets, surpassing the performance ceiling typically observed in standard-sized models. In the rest of the subsection, we present detailed evaluation results and analyses across individual benchmarks.

% CodeAgent相关的结果
\paragraph{\textcolor{orange!80!black}{Evaluation on Terminal-Based Benchmarks.}} 
We evaluate models on a suite of terminal-based agentic benchmarks that emphasize \emph{execution robustness}, \emph{long-horizon multi-turn interaction}, and \emph{environment grounding}. These benchmarks go beyond single-shot code generation and require agents to iteratively reason, invoke tools, recover from execution errors, and maintain state across multiple interaction steps.
As shown in \autoref{tab.5}, \textbf{\textcolor{orange!80!black}{ROME}} achieves \textbf{41.50\%} on \texttt{Terminal-Bench 1.0}, \textbf{24.72\%} on \texttt{Terminal-Bench 2.0}, \textbf{57.40\%} on \texttt{SWE-Bench Verified}, and \textbf{40.00\%} on \texttt{SWE-Bench Multilingual}. These results consistently and substantially outperform other normal-sized models, including Qwen3-Coder-30B-A3B-Instruct, Devstral Small 2, and GPT-OSS-120B, across all evaluated benchmarks. Notably, the improvements are not confined to a single dataset but persist across benchmarks with varying task distributions, programming languages, and interaction lengths, indicating strong robustness and generalization in agentic behavior. From a \emph{scaling-efficiency} perspective, \textbf{\textcolor{orange!80!black}{ROME}} demonstrates a highly favorable performance–parameter trade-off. Despite activating only 3B parameters, it significantly surpasses dense and MoE models with substantially larger total or activated parameter counts. This highlights the effectiveness of \texttt{\textcolor{orange}{ALE}} in enhancing agentic reasoning and action execution, effectively breaking through the performance ceiling typically observed in normal-sized models. 
% This highlights the power of \texttt{\textcolor{orange}{ALE}} in enhancing agentic crafting capabilities, effectively breaking through the learning bottlenecks of normal-sized models (\autoref{fig:performance-size}).

\begin{table}[t]
\centering
\caption{Performance on Terminal-Based Benchmarks (Normal Models).}\label{tab.5}
\resizebox{\linewidth}{!}{
\begin{tabular}{l|c|cccccc}
\toprule
\textbf{Benchmark} & \textbf{ROME} 
& \begin{tabular}[c]{@{}c@{}}\textbf{Qwen3-Coder}\\\textbf{30B-A3B-Instruct}\end{tabular} 
& \begin{tabular}[c]{@{}c@{}}\textbf{Devstral}\\\textbf{Small 2}\end{tabular} 
& \begin{tabular}[c]{@{}c@{}}\textbf{GPT-OSS-}\\\textbf{120B}\end{tabular} 
& \begin{tabular}[c]{@{}c@{}}\textbf{Gemini-2.5}\\\textbf{Flash}\end{tabular} 
& \begin{tabular}[c]{@{}c@{}}\textbf{GLM-4.5}\\\textbf{Air}\end{tabular} 
& \begin{tabular}[c]{@{}c@{}}\textbf{GPT-5}\\\textbf{Mini}\end{tabular} \\ \midrule
 Architecture & MoE & MoE & Dense & MoE & - & MoE & -  \\
 \# Total Params & 30B & 30B & 24B & 117B & - & 106B & - \\
 \# Activated Params & 3B & 3B &- & 5.1B & - & 12B & - \\ \midrule
Terminal-Bench 1.0 & 41.50 & 28.50 & 28.33 & 31.25 & 23.75 & 30.00 & 33.75 \\
Terminal-Bench 2.0 & 24.72 & 13.48 & 18.20 & 21.12 & 16.40* & 17.30 & 20.97 \\
SWE-Bench Verified & 57.40 & 46.33 & 51.87 & 43.93 & 28.73* & 56.20 & 59.30 \\ 
SWE-Bench Multilingual & 40.00 & 30.00 & 27.00 & 34.84 & 11.50 & 38.16 & 49.67 \\ 
Terminal-Bench-Pro-Public & 40.50 & 26.00 & 32.17 & 32.00 & 23.67 & 33.00 & 34.75 \\
Terminal-Bench-Pro-Private & 21.50 & 11.33 & 17.00 & 27.83 & 15.17 & 15.83 & 29.50 \\ \midrule
\textbf{Avg.} & \textbf{37.60} & \textbf{25.94} & \textbf{29.10} & \textbf{31.83} & \textbf{19.87} & \textbf{31.75} & \textbf{37.99} \\ \bottomrule
\end{tabular}
}
\end{table}

\begin{table}[t]
\centering
\caption{Performance on Terminal-Based Benchmarks (Large Models).}\label{tab.6}
\resizebox{\linewidth}{!}{
\begin{tabular}{l|c|cccccc}
\toprule
\textbf{Benchmark}
& \textbf{ROME} 
& \begin{tabular}[c]{@{}c@{}}\textbf{Qwen3-Coder}\\\textbf{Plus}\end{tabular}
& \begin{tabular}[c]{@{}c@{}}\textbf{Qwen3-Coder}\\\textbf{480B-A35B-Instruct}\end{tabular} 
& \begin{tabular}[c]{@{}c@{}}\textbf{DeepSeek}\\\textbf{V3.1}\end{tabular} 
& \begin{tabular}[c]{@{}c@{}}\textbf{GLM-}\\\textbf{4.6}\end{tabular} 
& \begin{tabular}[c]{@{}c@{}}\textbf{Kimi-}\\\textbf{K2}\end{tabular} 
& \begin{tabular}[c]{@{}c@{}}\textbf{Claude-}\\\textbf{Haiku-4.5}\end{tabular}  \\
\midrule
 Architecture & MoE & MoE & MoE & MoE & MoE & MoE & -  \\
 \# Total Params & 30B & - & 480B & 671B & 355B & 1043B & - \\
 \# Activated Params & 3B & - &  35B & 37B & 32B & 32B & - \\ \midrule
Terminal-Bench 1.0 & 41.50 & 39.58 & 37.92 & 38.75 & 41.25 & 39.25 & 47.08 \\
Terminal-Bench 2.0 & 24.72 & 32.36 & 26.97 & 28.47 & 26.29 & 30.90 & 34.83 \\
SWE-Bench Verified & 57.40 & 65.87 & 65.20 & 62.20 & 62.67 & 64.80 & 69.60 \\
SWE-Bench Multilingual & 40.00 & 54.16 & 49.50 & 48.16 & 53.84 & 48.67 & 60.34 \\
Terminal-Bench-Pro-Public & 40.50 & 39.67 & 38.33 & 39.33 & 41.50 & 40.50 & 45.83 \\
Terminal-Bench-Pro-Private & 21.50 & 28.50 & 26.50 & 28.33 & 29.17 & 29.00 & 35.33 \\
\midrule
\textbf{Avg.} & \textbf{37.60} & \textbf{43.36} & \textbf{40.74} & \textbf{40.87} & \textbf{42.45} & \textbf{42.19} & \textbf{48.84} \\ \bottomrule
\end{tabular}
}
\end{table}
% More impressively, as a medium-scale model, \textbf{\textcolor{orange!80!black}{ROME}} has attained scores in most benchmarks that almost reach or exceed those of large-scale models (\autoref{tab.6}). Notably, our model excels in the widely used Terminal-Bench and \texttt{SWE-Bench Verified}. On \texttt{Terminal-Bench}, \textbf{\textcolor{orange!80!black}{ROME}} outperforms super large-scale models such as Qwen3-Coder-480B-A35B-Instruct and DeepSeek v3.1, which have a total parameter count of 671B, as well as advanced proprietary models like Qwen3-Coder-Plus and Kimi-K2. Similarly, on \texttt{SWE-Bench Verified}, our model surpasses leading models, including GLM-4.5 Air, Gemini-2.5-Flash, and GPT-OSS-120B. Furthermore, all agentic models exhibited only rudimentary performance on the more rigorous \texttt{Terminal Bench Pro} that we proposed. This underscores the considerable potential for improvement in the current agentic capabilities of LLMs. Therefore, we encourage the community to leverage Terminal Bench Pro to further explore the potential agentic abilities, fostering iterative progress in the community.
More impressively, As a Small-scale model, \textbf{\textcolor{orange!80!black}{ROME}} attains performance that approaches or even exceeds that of multiple large-scale and ultra-large-scale models across several benchmarks. On \texttt{Terminal-Bench 1.0}, \textbf{\textcolor{orange!80!black}{ROME}} (41.50\%) surpasses super large-scale models such as Qwen3-Coder-480B-A35B-Instruct (37.92\%) and DeepSeek-V3.1 (38.75\%), while achieving performance comparable to advanced proprietary systems including Qwen3-Coder-Plus (39.58\%) and Kimi-K2 (39.25\%), despite operating at a substantially smaller scale. Similarly, on the widely adopted \texttt{SWE-Bench Verified}, \textbf{\textcolor{orange!80!black}{ROME}} (57.40\%) surpasses or matches leading proprietary models such as GLM-4.5 Air (56.20\%), Gemini-2.5 Flash (28.73\%), and GPT-OSS-120B (43.93\%). This indicates that the benefits of \texttt{\textcolor{orange}{ALE}} extend beyond terminal-based interaction and generalize to real-world software engineering tasks involving repository understanding, patch generation, and regression validation.

Despite these encouraging results, all evaluated agentic models, including \textbf{\textcolor{orange!80!black}{ROME}} and large-scale baselines, exhibit only limited performance on the more challenging \textbf{Terminal Bench Pro} benchmark. This benchmark introduces stricter success criteria, deeper interaction horizons, and more complex environment dynamics, exposing systematic weaknesses such as error compounding, suboptimal recovery strategies, and brittle long-term planning. The uniformly low absolute scores highlight that current agentic LLMs, regardless of scale, remain far from solving realistic, high-difficulty terminal-based tasks. 
Taken together, these findings underscore both the effectiveness and the limitations of current agentic learning approaches. While \textbf{\textcolor{orange!80!black}{ROME}} significantly improves agentic efficiency and narrows the gap between medium-scale and large-scale models, the results on \texttt{Terminal-Bench-Pro} reveal substantial headroom for future research. 
% We therefore encourage the community to adopt Terminal-Bench-Pro as a stress test for next-generation agentic systems, facilitating more rigorous evaluation and iterative progress in autonomous code agents.

% 工具调用相关的结果
\paragraph{\textcolor{orange!80!black}{Evaluation on Tool-Use Benchmarks.}} 
% After establishing robust fundamental reasoning performance, we further analyze whether ROME has effectively integrated the appropriate tool usage patterns. 
We further evaluate the tool-use abilities of the models, with the results summarized in ~\autoref{tab.1}.
Overall, the results reveal that \textbf{\textcolor{orange!80!black}{ROME}} excels across the benchmarks. In six benchmark tests, our model achieved an average score of 49.46\%, significantly outperforming similar-sized models such as Qwen3-Coder-30B-A3B (40.87\%), and Devstral Small 2 (39.35\%), demonstrating remarkable efficiency. Meanwhile, we find that even when compared with slightly larger-size models, such as GPT-5-mini (close-sourced) and GLM-4.5 Air (106B-A12B), our model still achieves competitive performance. A granular analysis of the benchmarks indicates that our model excels particularly in the \texttt{MTU-Bench (Single-Turn)}, reaching a score of $62.45\%$, which is substantially higher than Gemini-2.5 Flash ($45.93\%$) and GPT-OSS-120B ($54.16\%$). While some models like GPT-5 Mini show volatility across different domains, our model maintains consistent efficacy, particularly in the \texttt{Tau2-Bench Retail} ($62.28\%$) and \texttt{Airline} ($50.50\%$) tasks. These results suggest that the architectural optimizations in \textbf{\textcolor{orange!80!black}{ROME}} provide a more stable foundation for complex tool-calling logic than many of its direct competitors.

Furthermore, as shown in ~\autoref{tab.2}, when compared to significantly larger models, \textbf{\textcolor{orange!80!black}{ROME}} remains highly competitive, often matching or exceeding the performance of models with vastly greater parameter counts. Despite having only a fraction of the activated parameters (3B) compared to models like DeepSeek-V3.1 (37B activated) and Qwen3-Coder 480B (35B activated), our model maintains a highly competitive average performance of 49.46\%. Specifically, \textbf{\textcolor{orange!80!black}{ROME}} outperforms Qwen3-Coder Plus (47.41\%) and performs on par with DeepSeek-V3.1 (49.94\%) across the aggregate suite. In the MTU-Bench (Single-Turn) category, our model's performance (62.45\%) actually exceeds that of DeepSeek-V3.1 (61.71\%) and several other large-scale alternatives. This "scaling efficiency" highlights \textbf{\textcolor{orange!80!black}{ROME}}’s ability to bridge the performance gap between medium-scale and large-scale models, suggesting that its specialized training for tool-use tasks provides a more effective path to achieving agentic capabilities than sheer parameter scaling alone.

% In particular, our model's performance on Tau2-Bench Airline ranks just below that of GLM-4.5-Air, reaching 50.50\%; and its performance on MTU (Single-Turn) is only surpassed by Qwen3-Coder-480B, reaching 62.45\%. This analysis suggests that \textbf{\textcolor{orange!80!black}{ROME}} achieves greater parameter efficiency and task adaptability, attributable to our advanced training pipeline.

\begin{table}[t]
\centering
\caption{Performance on Tool-Use Benchmarks (Normal Models).}\label{tab.1}
\resizebox{\linewidth}{!}{
\begin{tabular}{l|c|cccccc}
\toprule
\textbf{Benchmark} & \textbf{ROME} 
& \begin{tabular}[c]{@{}c@{}}\textbf{Qwen3-Coder}\\\textbf{30B-A3B-Instruct}\end{tabular} 
& \begin{tabular}[c]{@{}c@{}}\textbf{Devstral}\\\textbf{Small 2}\end{tabular} 
& \begin{tabular}[c]{@{}c@{}}\textbf{GPT-OSS-}\\\textbf{120B}\end{tabular} 
& \begin{tabular}[c]{@{}c@{}}\textbf{Gemini-2.5}\\\textbf{Flash}\end{tabular} 
& \begin{tabular}[c]{@{}c@{}}\textbf{GLM-4.5}\\\textbf{Air}\end{tabular} 
& \begin{tabular}[c]{@{}c@{}}\textbf{GPT-5}\\\textbf{Mini}\end{tabular} \\ \midrule
 Architecture & MoE & MoE & Dense & MoE & - & MoE & -  \\
 \# Total Params & 30B & 30B & 24B & 117B & - & 106B & - \\
 \# Activated Params & 3B & 3B &- & 5.1B & - & 12B & - \\ \midrule
Tau2-Bench Retail & 62.28 & 59.87 & 58.33 & 64.30 & 64.30$^*$ & 74.60 & 74.12 \\
Tau2-Bench Airline & 50.50 & 45.50 & 30.00 & 53.50 & 42.50$^*$ & 69.00 & 60.00 \\
Tau2-Bench Telecom & 30.92 & 30.04 & 20.40 & 54.61 & 16.90$^*$ & 46.90 & 73.46 \\
BFCL-v3 (Multi-Turn) & 43.00 & 29.75 & 30.12 & 53.62 & 36.25$^*$ & 66.88 & 27.25 \\
MTU-Bench (Single-Turn) & 62.45 & 50.69 & 63.08 & 54.16 & 45.93 & 57.74 & 59.82 \\
MTU-Bench (Multi-Turn) & 47.63 & 29.38 & 34.15 & 58.61 & 57.01 & 37.55 & 55.61 \\ \midrule
\textbf{Avg.} & \textbf{49.46} & \textbf{40.87} & \textbf{39.35} & \textbf{56.47} & \textbf{43.82} & \textbf{58.78} & \textbf{58.38} \\ \bottomrule
\end{tabular}
}
\end{table}

\begin{table}[t]
\centering
\caption{Performance on Tool-Use Benchmarks (Large Models).}\label{tab.2}
\resizebox{\linewidth}{!}{
\begin{tabular}{l|c|cccccc}
\toprule
\textbf{Benchmark}
& \textbf{ROME} 
& \begin{tabular}[c]{@{}c@{}}\textbf{Qwen3-Coder}\\\textbf{Plus}\end{tabular}
& \begin{tabular}[c]{@{}c@{}}\textbf{Qwen3-Coder}\\\textbf{480B-A35B-Instruct}\end{tabular} 
& \begin{tabular}[c]{@{}c@{}}\textbf{DeepSeek}\\\textbf{V3.1}\end{tabular} 
& \begin{tabular}[c]{@{}c@{}}\textbf{GLM-}\\\textbf{4.6}\end{tabular} 
& \begin{tabular}[c]{@{}c@{}}\textbf{Kimi-}\\\textbf{K2}\end{tabular} 
& \begin{tabular}[c]{@{}c@{}}\textbf{Claude-}\\\textbf{Haiku-4.5}\end{tabular}  \\
\midrule
 Architecture & MoE & MoE & MoE & MoE & MoE & MoE & -  \\
 \# Total Params & 30B & - & 480B & 671B & 355B & 1043B & - \\
 \# Activated Params & 3B & - &  35B & 37B & 32B & 32B & - \\ \midrule
Tau2-Bench Retail & 62.28 & 62.28 & 59.00 & 71.50 & 76.10 & 67.54 & 67.32 \\
Tau2-Bench Airline & 50.50 & 48.00 & 48.00 & 52.00 & 65.00 & 49.00 & 47.50 \\
Tau2-Bench Telecom & 30.92 & 52.19 & 58.55 & 40.35 & 71.05 & 86.40 & 36.40 \\
BFCL-v3 (Multi-Turn) & 43.00 & 27.75 & 42.38 & 20.62 & 67.50 & 50.63$^*$ & 53.50 \\
MTU-Bench (Single-Turn) & 62.45 & 56.68 & 63.87 & 61.71 & 49.54  & 56.21 & 61.19 \\
MTU-Bench (Multi-Turn) & 47.63 & 37.56 & 34.85 & 53.44 & 37.55& 53.31 & 55.43 \\ \midrule
% \textbf{Avg.} & \textbf{46.86} & \textbf{45.56} & \textbf{48.56} & \textbf{47.58} & \textbf{63.44} & \textbf{61.37} & \textbf{52.03} \\ 
\textbf{Avg.} & \textbf{49.46} & \textbf{47.41} & \textbf{51.11} & \textbf{49.94} & \textbf{61.12} & \textbf{60.52} & \textbf{53.56} \\
\bottomrule
\end{tabular}
}
\end{table}

% 通用Agent任务
% 三类任务的侧重点介绍；取得的效果；
\paragraph{\textcolor{orange!80!black}{Evaluation on General Agentic Benchmarks.}} 
After establishing a robust fundamental tool-use performance of our model, we conducted a unified analysis of the models on general agentic benchmarks that require multi-turn interactions and action decision-making. Specifically, we consider \texttt{GAIA}, which focuses on everyday queries requiring coordinated use of multiple tools (e.g., web search, data analysis, and logical reasoning), and \texttt{BrowseComp-ZH}, which emphasizes Chinese multi-hop web search with evidence aggregation across heterogeneous webpages. In addition, we introduce \texttt{ShopAgent}, a high-quality proprietary benchmark constructed from real-world e-commerce assistant scenarios, designed to systematically evaluate an agent’s ability to infer user preferences, retrieve and compare products, reason over structured attributes, and adapt to evolving user intent through multi-step interactions. Both Single-Turn and Multi-Turn settings require long-horizon, multi-step interactions, where the Multi-Turn setting is particularly challenging as users may revise or refine their intentions during subsequent interactions, demanding robust intent clarification and adaptive planning.

\autoref{tab.3} and \autoref{tab.4} report the performance of \textbf{\textcolor{orange!80!black}{ROME}} compared with a wide range of strong baselines, including both normal-scale and large-scale models. 
% Compared to normal-sized models (\autoref{tab.3}), our model demonstrates a comprehensive advantage, with an average score of 25.64\%. This score significantly exceeds that of Qwen3-Coder-30B-A3B-Instruct and Devstral Small 2.
% Among normal-sized models, \textbf{\textcolor{orange!80!black}{ROME}} achieves the highest overall average score (\textbf{25.64\%}), consistently outperforming other competitive baselines such as Qwen3-Coder-30B-A3B-Instruct(15.69\%), Devstral Small 2(16.30\%), Gemini-2.5 Flash (22.66\%) and GLM-4.5 Air(24.78\%). The advantage of \textbf{\textcolor{orange!80!black}{ROME}} is particularly pronounced on \texttt{ShopAgent}, where it obtains \textbf{34.53\%} on the Single-Turn setting and \textbf{29.61\%} on the Multi-Turn setting, substantially surpassing all other normal-sized models. These results highlight ROME’s strong capability in long-horizon planning, user preference modeling, and adaptive interaction, which are critical for realistic shopping assistant tasks involving intent clarification and personalized recommendation. 
Additionally, \textbf{\textcolor{orange!80!black}{ROME}} achieves performance comparable to that of larger open-source agentic models across most benchmarks, as shown in \autoref{tab.4}. Notably, our model even surpassed the super-large scale GLM-4.6 in the complex \texttt{ShopAgent} task. These results demonstrate strong generalization across diverse agentic workloads.
Overall, \textbf{\textcolor{orange!80!black}{ROME}} significantly outperforms other models of comparable scale, achieving an average score of \textbf{25.64\%}, markedly higher than Qwen3-Coder-30B-A3B-Instruct (15.69\%) and Devstral Small 2 (16.30\%). Beyond scale-matched comparisons, \textbf{\textcolor{orange!80!black}{ROME}} also demonstrates strong competitiveness against substantially larger models, outperforming Gemini-2.5 Flash (22.66\%), GLM-4.5 Air (24.78\%), Qwen3-Coder-Plus (23.99\%), and Qwen3-Coder-480B-A35B-Instruct (23.88\%). Moreover, \textbf{\textcolor{orange!80!black}{ROME}} achieves performance close to Kimi-K2, despite the latter having a total parameter count of 1043B with 32B activated parameters.
The advantage of \textbf{\textcolor{orange!80!black}{ROME}} is particularly pronounced on the \texttt{ShopAgent} benchmark, where it attains \textbf{34.53\%} in the Single-Turn setting and \textbf{29.61\%} in the more challenging Multi-Turn setting, substantially surpassing all other normal-sized models. These results highlight \textbf{\textcolor{orange!80!black}{ROME}}’s strong capability in long-horizon planning, user preference modeling, and adaptive interaction—key competencies for realistic shopping assistant scenarios involving intent clarification and personalized recommendation.

\begin{table}[t]
\centering
\caption{Performance on General-Agent Benchmarks (Normal Models).}\label{tab.3}
\resizebox{\linewidth}{!}{
\begin{tabular}{l|c|cccccc}
\toprule
\textbf{Benchmark} & \textbf{ROME} 
& \begin{tabular}[c]{@{}c@{}}\textbf{Qwen3-Coder}\\\textbf{30B-A3B-Instruct}\end{tabular} 
& \begin{tabular}[c]{@{}c@{}}\textbf{Devstral}\\\textbf{Small 2}\end{tabular} 
& \begin{tabular}[c]{@{}c@{}}\textbf{GPT-OSS-}\\\textbf{120B}\end{tabular} 
& \begin{tabular}[c]{@{}c@{}}\textbf{Gemini-2.5}\\\textbf{Flash}\end{tabular} 
& \begin{tabular}[c]{@{}c@{}}\textbf{GLM-4.5}\\\textbf{Air}\end{tabular} 
& \begin{tabular}[c]{@{}c@{}}\textbf{GPT-5}\\\textbf{Mini}\end{tabular} \\ \midrule
 Architecture & MoE & MoE & Dense & MoE & - & MoE & -  \\
 \# Total Params & 30B & 30B & 24B & 117B & - & 106B & - \\
 \# Activated Params & 3B & 3B &- & 5.1B & - & 12B & - \\ \midrule
GAIA & 24.24 & 20.00 & 21.21 & 33.54 & 34.14 & 31.92 & 51.52 \\
BrowseComp-ZH & 14.19 & 7.27 & 7.27 & 20.42 & 18.11 & 21.11 & 40.83 \\
ShopAgent (Single-Turn) & 34.53 & 22.11 & 19.44 & 21.11 & 20.89 & 25.97 & 23.58 \\
ShopAgent (Multi-Turn) & 29.61 & 13.38 & 17.28 & 18.54 & 17.51 & 20.12 & 26.41 \\ \midrule
% \textbf{Avg.} & \textbf{22.68} & \textbf{13.55} & \textbf{15.25} & \textbf{24.17} & \textbf{23.25} & \textbf{24.38} & \textbf{39.59} \\ 
\textbf{Avg.} & \textbf{25.64} & \textbf{15.69} & \textbf{16.30} & \textbf{23.40} & \textbf{22.66} & \textbf{24.78} & \textbf{35.59} \\
\bottomrule
\end{tabular}

}
\end{table}

\begin{table}[t]
\centering
\caption{Performance on General-Agent Benchmarks (Large Models).}\label{tab.4}
\resizebox{\linewidth}{!}{
\begin{tabular}{l|c|cccccc}

\toprule
\textbf{Benchmark}
& \textbf{ROME} 
& \begin{tabular}[c]{@{}c@{}}\textbf{Qwen3-Coder}\\\textbf{Plus}\end{tabular}
& \begin{tabular}[c]{@{}c@{}}\textbf{Qwen3-Coder}\\\textbf{480B-A35B-Instruct}\end{tabular} 
& \begin{tabular}[c]{@{}c@{}}\textbf{DeepSeek}\\\textbf{V3.1}\end{tabular} 
& \begin{tabular}[c]{@{}c@{}}\textbf{GLM-}\\\textbf{4.6}\end{tabular} 
& \begin{tabular}[c]{@{}c@{}}\textbf{Kimi-}\\\textbf{K2}\end{tabular} 
& \begin{tabular}[c]{@{}c@{}}\textbf{Claude-}\\\textbf{Haiku-4.5}\end{tabular}  \\
\midrule
 Architecture & MoE & MoE & MoE & MoE & MoE & MoE & -  \\
 \# Total Params & 30B & - & 480B & 671B & 355B & 1043B & - \\
 \# Activated Params & 3B & - &  35B & 37B & 32B & 32B & - \\ \midrule
GAIA & 24.24 & 31.52 & 33.74 & 31.92 & 35.76 & 34.55 & 41.01 \\ 
BrowseComp-ZH & 14.19 & 15.80 & 13.15 & 23.88 & 24.33 & 15.22 & 22.15 \\
ShopAgent (Single-Turn) & 34.53 & 26.54 & 27.66 & 38.87 & 33.80 & 30.97 & 36.21 \\ 
ShopAgent (Multi-Turn) & 29.61 & 22.08 & 20.98 & 33.97 & 22.12 & 26.26 & 30.65 \\ \midrule
% \textbf{Avg.} & \textbf{22.68} & \textbf{23.13} & \textbf{22.62} & \textbf{29.92} & \textbf{27.40} & \textbf{25.34} & \textbf{31.27} \\ 
\textbf{Avg.} & \textbf{25.64} & \textbf{23.99} & \textbf{23.88} & \textbf{32.16} & \textbf{29.00} & \textbf{26.75} & \textbf{32.51} \\
\bottomrule
\end{tabular}
}
\end{table}

\section{Conclusion}

The pursuit of \emph{agentic crafting} represents a significant advancement in the capabilities of LLMs, moving beyond simple one-shot responses to operate effectively within dynamic, real-world environments. This shift necessitates a robust agentic ecosystem to facilitate the planning, execution, and reliability required for complex tasks. Through our introduction of the \textbf{Agentic Learning Ecosystem (\texttt{\textcolor{orange}{ALE}})}, we lay the groundwork for streamlining the development and deployment of agentic LLMs. Specifically, by integrating systematic components, i.e., \textbf{ROLL}, \textbf{ROCK}, and \textbf{iFlow CLI}, we provide a comprehensive infrastructure that optimizes the complete production pipeline for agent LLMs.  Anchored by our proposed policy optimization algorithm \texttt{\textcolor{orange}{IPA}}, the training pipeline ultimately fosters a smoother transition into the agent era. The deployment of \textcolor{orange!80!black}{\textbf{\textbf{ROME}}}, an open-source agent built upon this ecosystem and trained on extensive trajectories, demonstrates the potential of this approach. 

Our empirical evaluations, supported by various benchmarks and our newly proposed \textbf{Terminal Bench Pro} benchmark, underscore \textcolor{orange!80!black}{\textbf{\textbf{ROME}}}'s strong performance across diverse contexts, reaffirming the practicality and effectiveness of the \texttt{\textcolor{orange}{ALE}} framework. This foundational infrastructure not only enhances agent model development but also bridges the gap in the open-source community, addressing the challenges that have impeded the practical implementation and adoption of agents. As we continue to refine and expand upon this ecosystem, we anticipate that our efforts will significantly contribute to the evolution of agentic modeling and the broader landscape of AGI applications.

\clearpage
\section{Authors}

Within each role, authors are listed alphabetically.

\begin{multicols}{2}

\textbf{\textcolor{orange!80!black}{Project Lead}}\\
\vspace{-3mm}
\begin{itemize}[leftmargin=1.5em, itemsep=1pt]
\item Weixun Wang
\item XiaoXiao Xu
\end{itemize}

% \vspace{1em}
\textbf{\textcolor{orange!80!black}{Core Contributors}}\\
\vspace{-3mm}
\begin{itemize}[leftmargin=1.5em, itemsep=1pt]
\item Wanhe An
\item Fangwen Dai
\item Wei Gao
\item Yancheng He
\item Ju Huang
\item Qiang Ji
\item Hanqi Jin
\item Xiaoyang Li
\item Yang Li
\item Zhongwen Li
\item Shirong Lin
\item Jiashun Liu
\item Zenan Liu
\item Tao Luo
\item Dilxat Muhtar
\item Yuanbin Qu
\item Jiaqiang Shi
\item Qinghui Sun
\item Yingshui Tan
\item Hao Tang
\item Runze Wang
\item Yi Wang
\item Zhaoguo Wang
\item Yanan Wu
\item Shaopan Xiong
\item Binchen Xu
\item Xander Xu
\item Yuchi Xu
\item Qipeng Zhang
\item Xixia Zhang
\item Haizhou Zhao
\item Jie Zhao
\item Shuaibing Zhao
\item Baihui Zheng
\item Jianhui Zheng
\item Suhang Zheng
\item Yanni Zhu
\end{itemize}

% \vspace{1em}
\textbf{\textcolor{orange!80!black}{Contributors}}\\
\vspace{-3mm}
\begin{itemize}[leftmargin=1.5em, itemsep=1pt]
\item Mengze Cai
\item Kerui Cao
\item Xitong Chen
\item Yue Dai
\item Lifan Du
\item Tao Feng
\item Tao He
\item Jin Hu
\item Yijie Hu
\item Ziyu Jiang
\item Cheng Li
\item Xiang Li
\item Jing Liang
\item Xin Lin
\item Chonghuan Liu
\item ZhenDong Liu
\item Zhiqiang Lv
\item Haodong Mi
\item Yanhu Mo
\item Junjia Ni
\item Shixin Pei
\item Jingyu Shen
\item XiaoShuai Song
\item Cecilia Wang
\item Chaofan Wang
\item Kangyu Wang
\item Pei Wang
\item Tao Wang
\item Wei Wang
\item Ke Xiao
\item Mingyu Xu
\item Tiange Xu
\item Nan Ya
\item Siran Yang
\item Jianan Ye
\item Yaxing Zang
\item Duo Zhang
\item Junbo Zhang
\item Boren Zheng
\end{itemize}

% \vspace{1em}
\textbf{\textcolor{orange!80!black}{Supervision}}\\
\vspace{-5mm}
\begin{itemize}[leftmargin=1.5em, itemsep=1pt]
\item Wanxi Deng
\item Ling Pan
\item Lin Qu
\item Wenbo Su
\item Jiamang Wang
\item Wei Wang
\item Hu Wei
\item Minggang Wu
\item Cheng Yu
\item Bing Zhao
\item Zhicheng Zheng
\item Bo Zheng
\end{itemize}

\end{multicols}

\clearpage
% \appendix

\section{Appendix}
\label{sec:appendix}

\subsection{Real-world Case Study and Subjective Evaluation}
Here, we present several concrete real-world task cases to further demonstrate the superiority of our model in agentic crafting capabilities.

As summarized in ~\autoref{tab:app.1}, we conduct a comprehensive evaluation of the model’s capability to execute real-world tasks. We curate a benchmark of 100 distinct tasks (from de-identified real user logs collected via the iFlow CLI) and assess outputs along five dimensions: (1) Functionality \& Interaction Implementation, which emphasizes correct core logic, smooth user interaction, and absence of critical defects; (2) Layout \& Style Replication, which measures visual fidelity, responsiveness, and adherence to design specifications; (3) Code Quality \& Robustness, focusing on structural clarity, standardized naming, maintainability, and error-free execution; (4) Structural \& Semantic Correctness, evaluating compliance with HTML5 semantic conventions; and (5) Innovation \& Prompt Understanding, capturing accurate requirement comprehension and reasonable value-added enhancements. For comparison, we select two similarly sized models (Qwen3-Coder-30B-A3B-Instruct and Devstral-Small-2) as well as two large-scale models (GLM-4.6 and Qwen3-Coder-Plus) as our reference baselines. To improve reliability and reduce evaluator bias, we employ a blinded annotation protocol involving 20 independent domain experts, who judge results without access to model identity. Final labels are determined via majority voting and used to compute the overall win rate. The aggregated quantitative results, together with representative qualitative examples and selected screenshots, are reported in the following sections.

% \begin{table}[htbp]
% \centering
% \begin{tabular}{p{0.32\linewidth} p{0.12\linewidth} p{0.50\linewidth}}
% \toprule
% Evaluation Dimension & Weight & Core Requirements \\
% \midrule
% Functionality \& Interaction Implementation & 40\% & Complete core logic, smooth interaction, no critical defects \\
% Layout \& Style Replication & 20\% & Visually appealing, responsive, compliant with design specifications \\
% Code Quality \& Robustness & 20\% & Clear structure, standardized naming, error-free, maintainable \\
% Structural \& Semantic Correctness & 10\% & HTML5 semantics \\
% Innovation \& Prompt Understanding & 10\% & Accurate understanding of requirements + reasonable feature enhancements \\
% \bottomrule
% \end{tabular}
% \caption{Evaluation rubric for real-world case study, detailing the five assessment dimensions, their point weights, and the corresponding core requirements.}
% \label{tab:app.1}
% \end{table}
\begin{table}[htbp]
\centering
\begin{tabular}{
p{0.24\linewidth} | p{0.08\linewidth} | p{0.60\linewidth}
}
\toprule
Evaluation Dimension & Weight & Core Requirements \\
\midrule
Functionality \& Interaction Implementation & 40\% & Complete core logic, smooth interaction, no critical defects \\
Layout \& Style Replication & 20\% & Visually appealing, responsive, compliant with design specifications \\
Code Quality \& Robustness & 20\% & Clear structure, standardized naming, error-free, maintainable \\
Structural \& Semantic Correctness & 10\% & HTML5 semantics \\
Innovation \& Prompt Understanding & 10\% & Accurate understanding of requirements + reasonable feature enhancements \\
\bottomrule
\end{tabular}
\caption{Evaluation rubric for real-world case study, detailing the five assessment dimensions, their point weights, and the corresponding core requirements.}
\label{tab:app.1}
\end{table}

\begin{wrapfigure}{r}{0.42\textwidth} % r = right, l = left
  \centering
  \vspace{-0.5\baselineskip} % optional: tweak vertical spacing
  \includegraphics[width=0.40\textwidth]{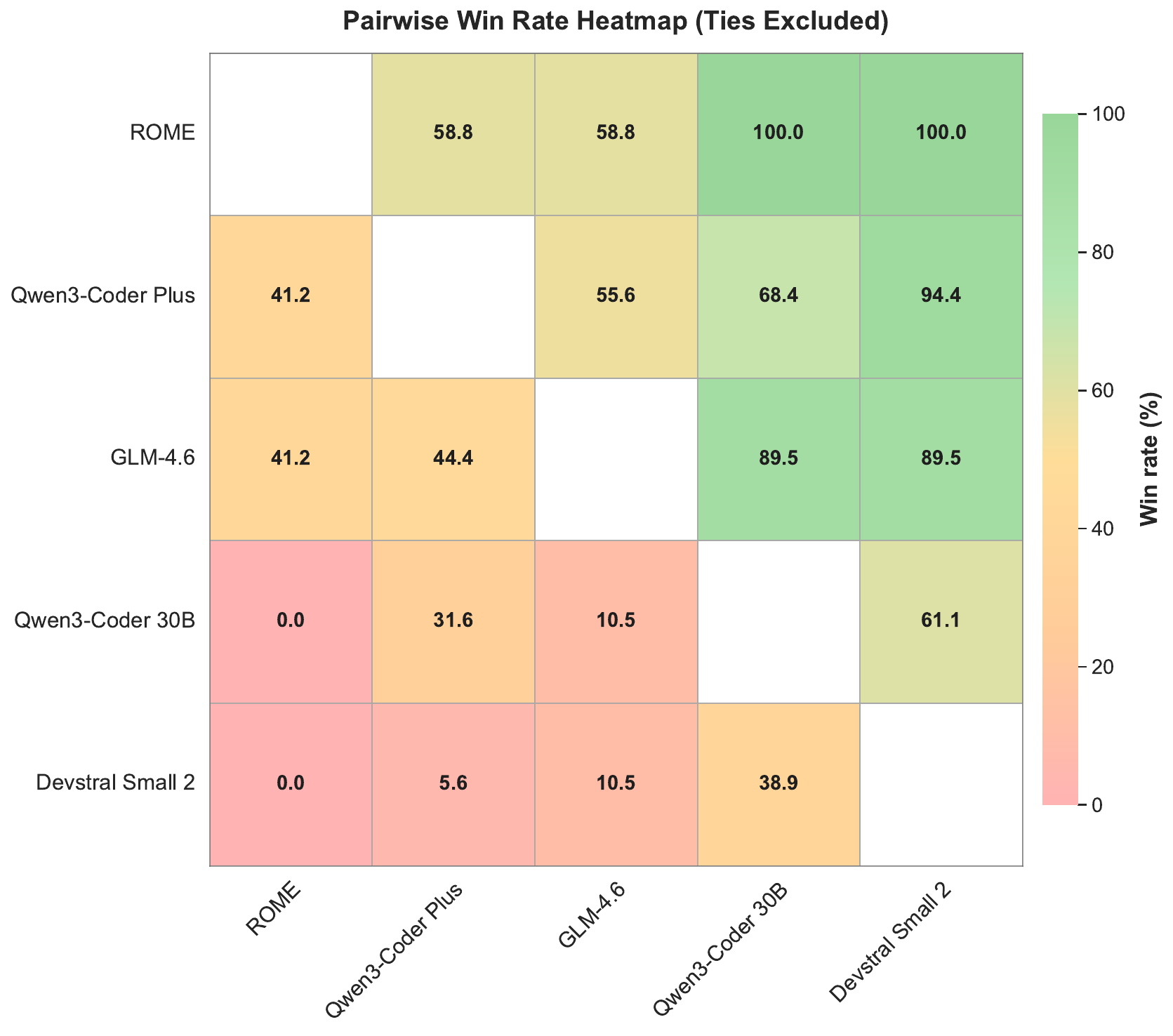}
  \caption{Pairwise win-rate matrix (\%) on the 100-task real-world benchmark under 30-expert blinded majority voting. Each cell reports the percentage of tasks where the row model is judged better than the column model; higher values (green) indicate stronger performance.}
  \label{fig:win-rate-heatmap-appendix}
  \vspace{-0.8\baselineskip} % optional: tweak vertical spacing
\end{wrapfigure}

As shown in \autoref{fig:win-rate-heatmap-appendix}, across the 100-task benchmark, \textcolor{orange!80!black}{\textbf{ROME}} demonstrates consistent advantages over all evaluated baselines in overall task execution quality. Notably, these gains persist even when compared against larger, same-series model (e.g., Qwen3-Coder-Plus) and a strong state-of-the-art agentic model (GLM-4.6), indicating that \textcolor{orange!80!black}{\textbf{ROME}}’s improvements are not merely attributable to parameter scale. This result suggests that \textcolor{orange!80!black}{\textbf{ROME}} more effectively translates high-level requirements into executable plans and reliably completes multi-step workflows, yielding outputs that are not only functionally correct but also better aligned with interaction design, robustness expectations, and semantic structure. In practice, ROME exhibits fewer critical failures in core logic and integration, maintains higher stability under varied task specifications, and provides more consistent end-to-end deliverables across task types. Overall, the findings imply that ROME achieves a form of ``scale-breaking'' agentic capability—i.e., stronger real-task completion performance than would be expected from model size alone—highlighting the effectiveness of our approach for enhancing agentic execution beyond scaling laws. 

\begin{figure}[!htbp]
  \centering
  % 关键参数：三列均匀分布；每张图同高（height），并保持比例（keepaspectratio）
  \setlength{\tabcolsep}{0pt} % 去掉列间默认空白
  \renewcommand{\arraystretch}{0} % 去掉行间默认空白（如需更紧凑）
  \newcommand{\imgH}{0.1\textheight} % 5 行 => 每行约 0.2，高度略留空隙
  \newcommand{\imgW}{0.3\textwidth}  % 3 列 + \hfill 间距

  % --- Row 1 ---
  \begin{subfigure}[t]{\imgW}
    \centering
    \includegraphics[height=\imgH,keepaspectratio]{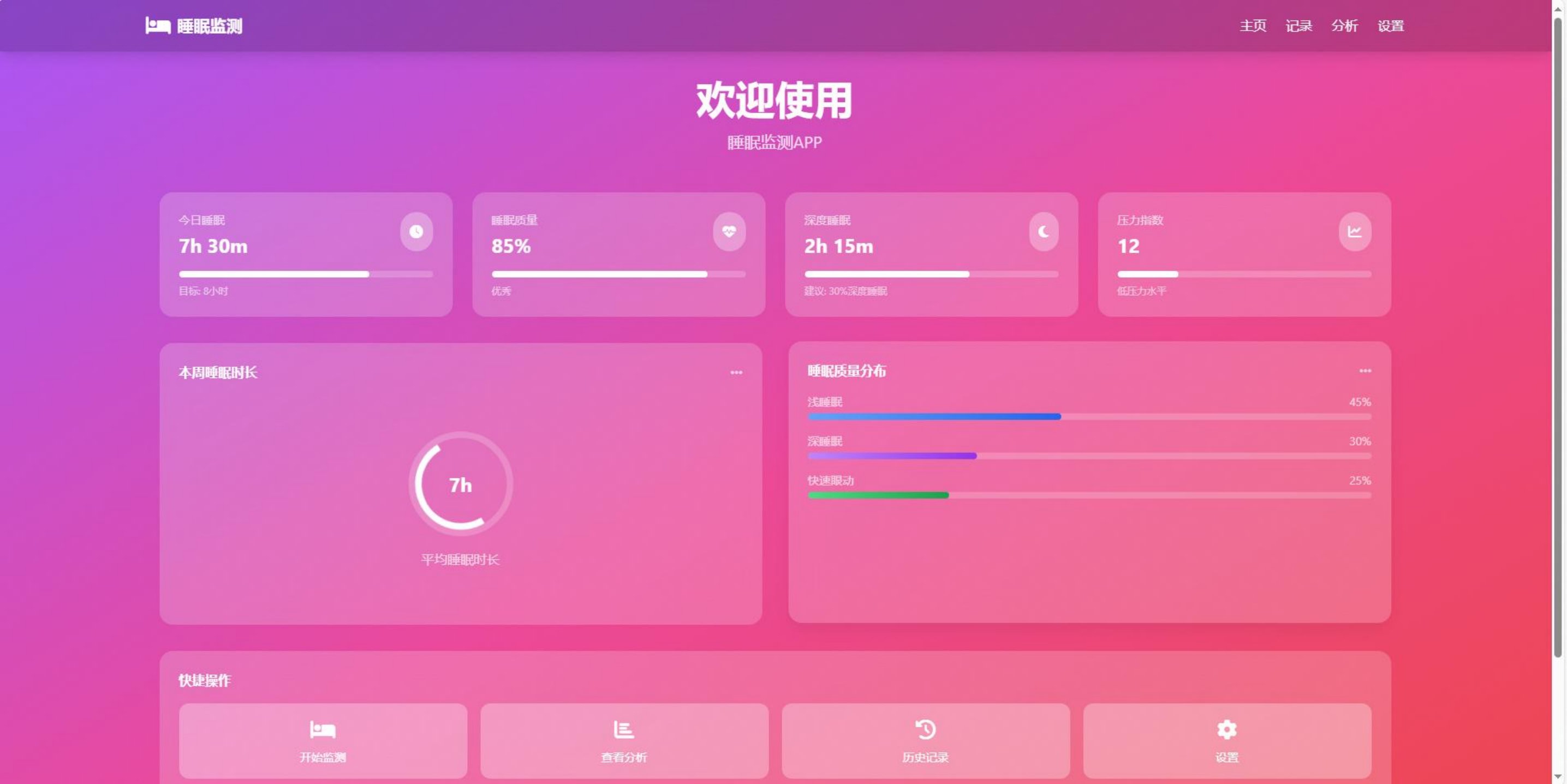}
    \caption{ROME screenshot 1}
    \label{fig:ph01}
  \end{subfigure}\hfill
  \begin{subfigure}[t]{\imgW}
    \centering
    \includegraphics[height=\imgH,keepaspectratio]{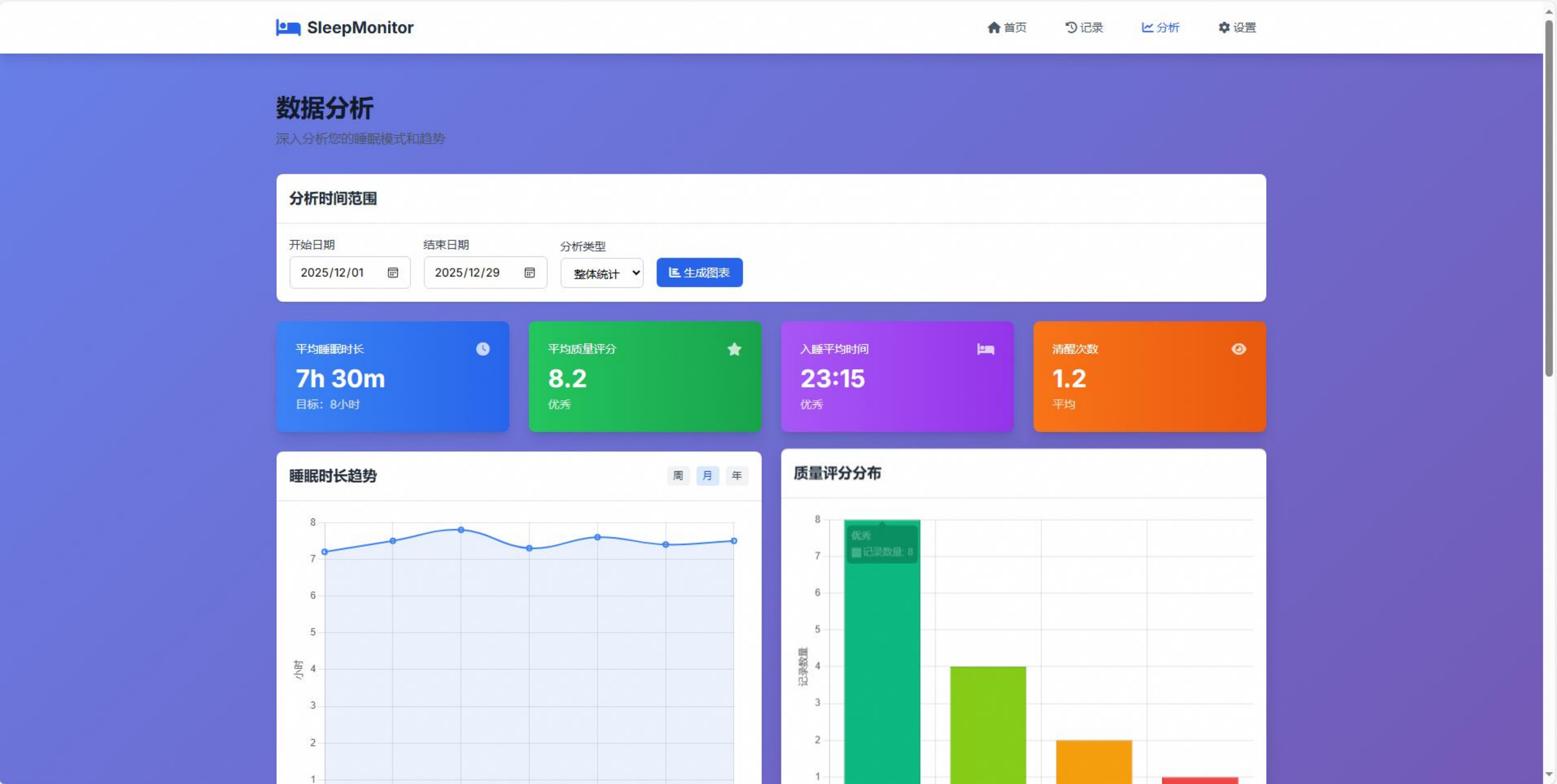}
    \caption{ROME screenshot 2}
    \label{fig:ph02}
  \end{subfigure}\hfill
  \begin{subfigure}[t]{\imgW}
    \centering
    \includegraphics[height=\imgH,keepaspectratio]{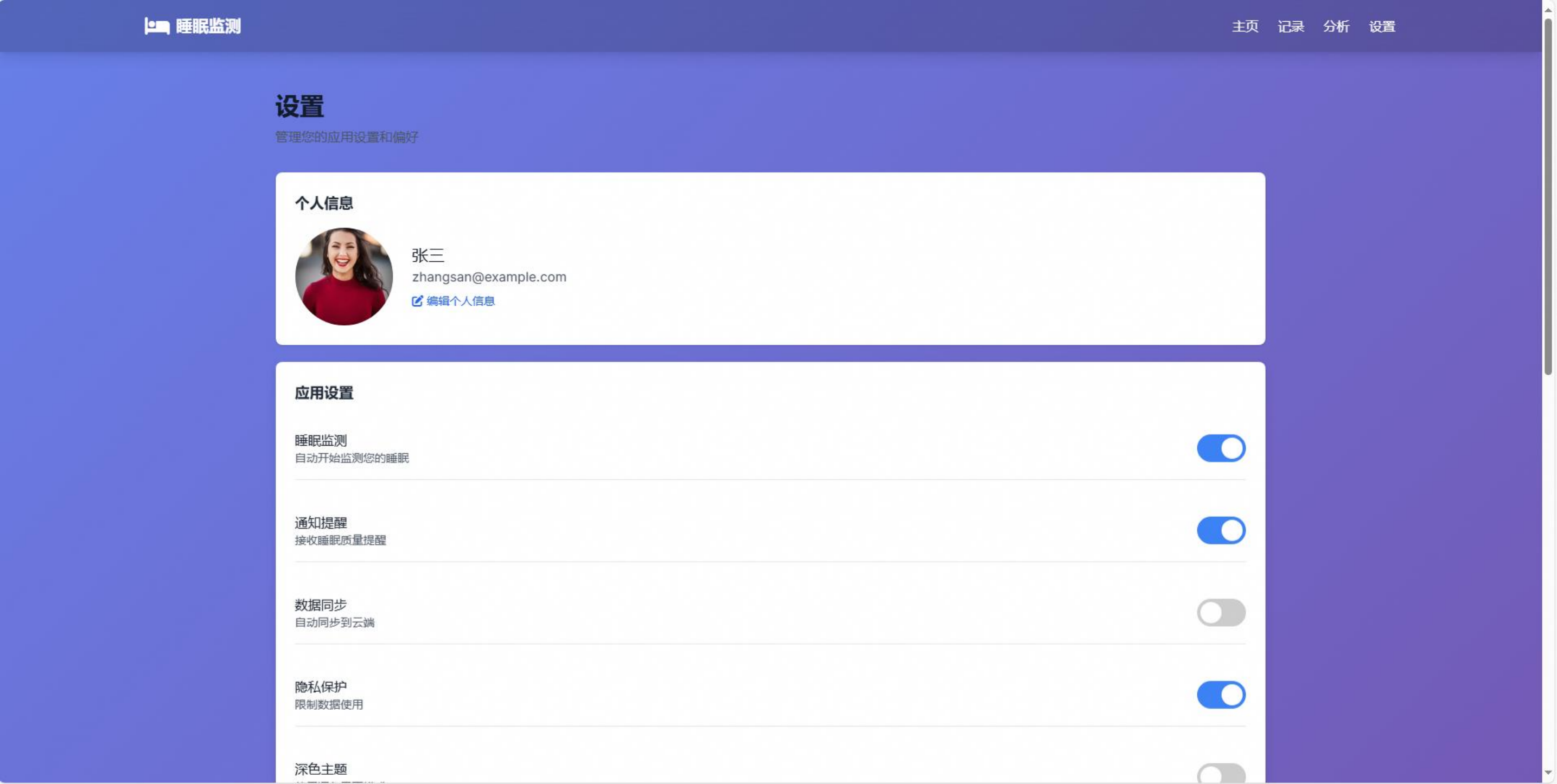}
    \caption{ROME screenshot 3}
    \label{fig:ph03}
  \end{subfigure}

  \vspace{0.012\textheight}

  % --- Row 2 ---
  \begin{subfigure}[t]{\imgW}
    \centering
    \includegraphics[height=\imgH,keepaspectratio]{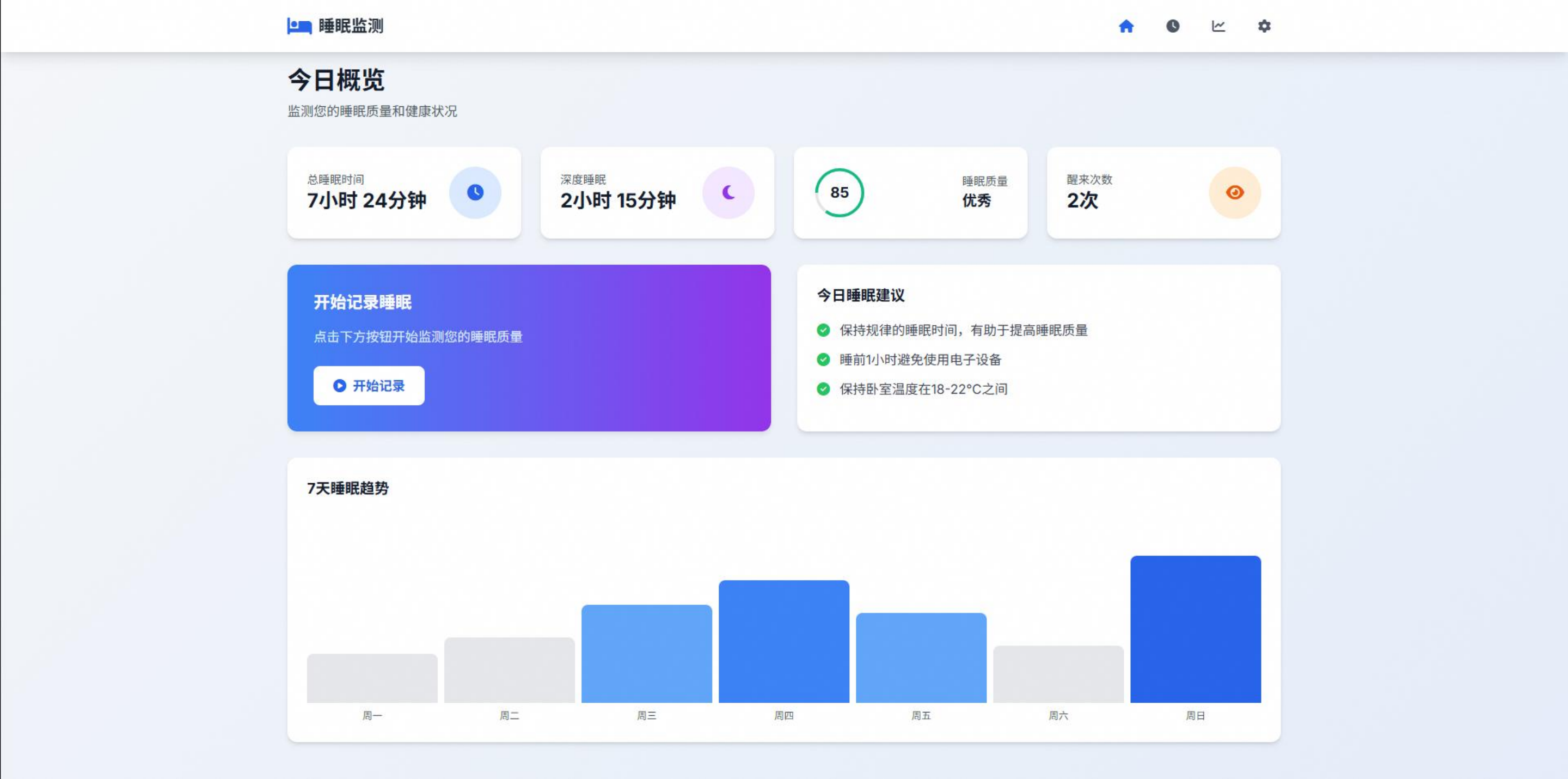}
    \caption{Qwen3-Coder-Plus screenshot1}
    \label{fig:ph04}
  \end{subfigure}\hfill
  \begin{subfigure}[t]{\imgW}
    \centering
    \includegraphics[height=\imgH,keepaspectratio]{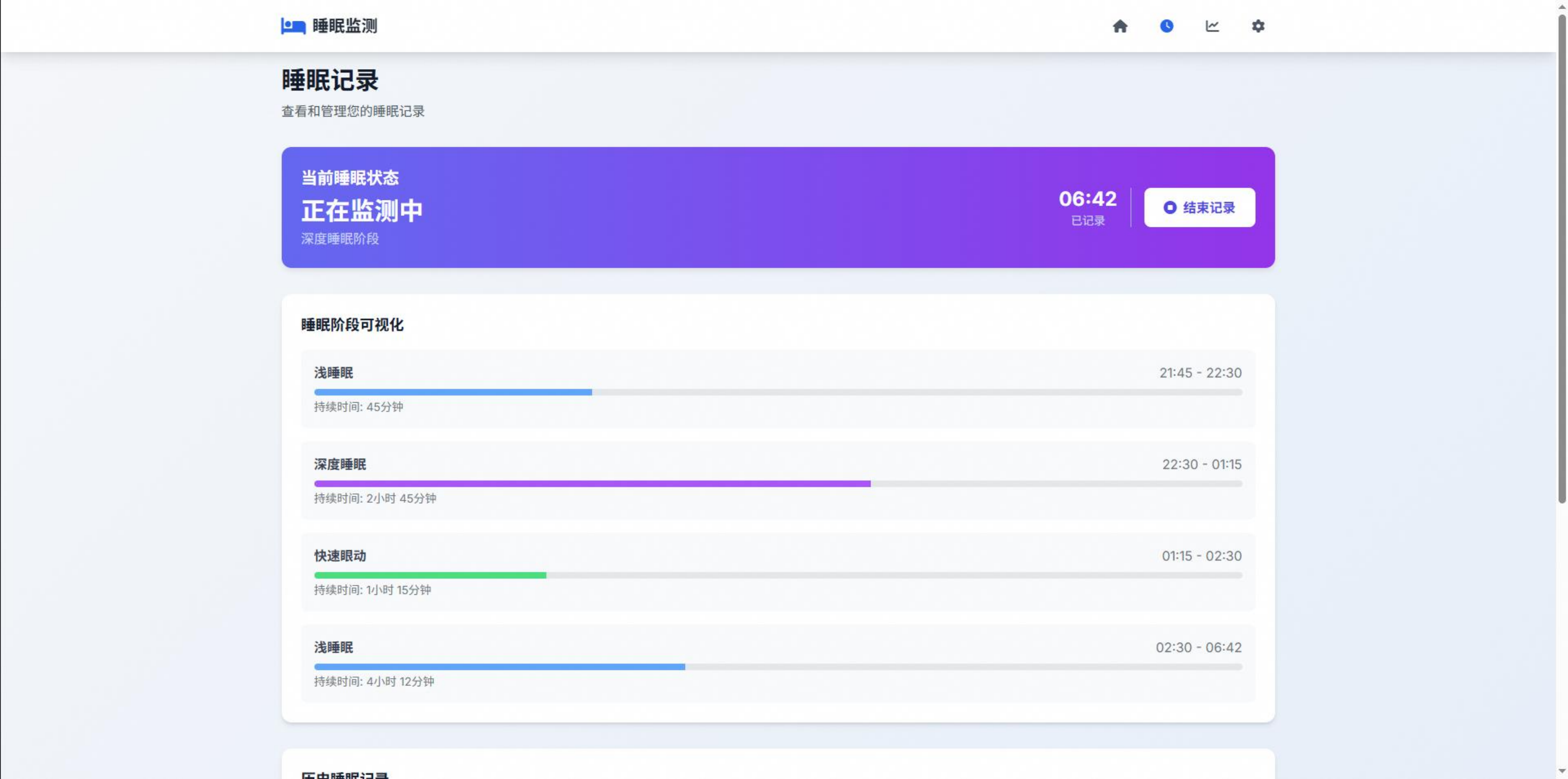}
    \caption{Qwen3-Coder-Plus screenshot 2}
    \label{fig:ph05}
  \end{subfigure}\hfill
  \begin{subfigure}[t]{\imgW}
    \centering
    \includegraphics[height=\imgH,keepaspectratio]{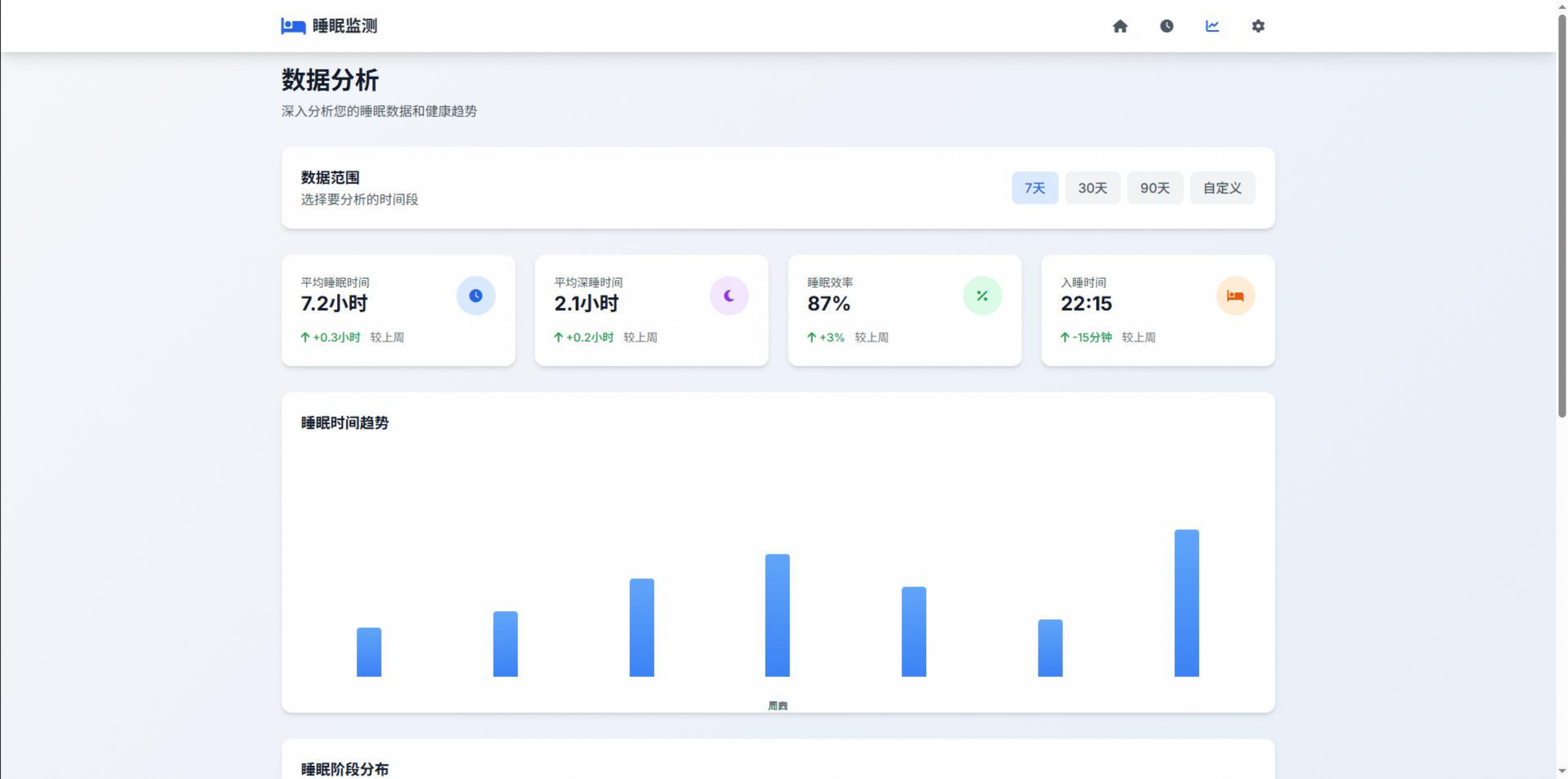}
    \caption{Qwen3-Coder-Plus screenshot 3}
    \label{fig:ph06}
  \end{subfigure}

  \vspace{0.012\textheight}

  % --- Row 3 ---
  \begin{subfigure}[t]{\imgW}
    \centering
    \includegraphics[height=\imgH,keepaspectratio]{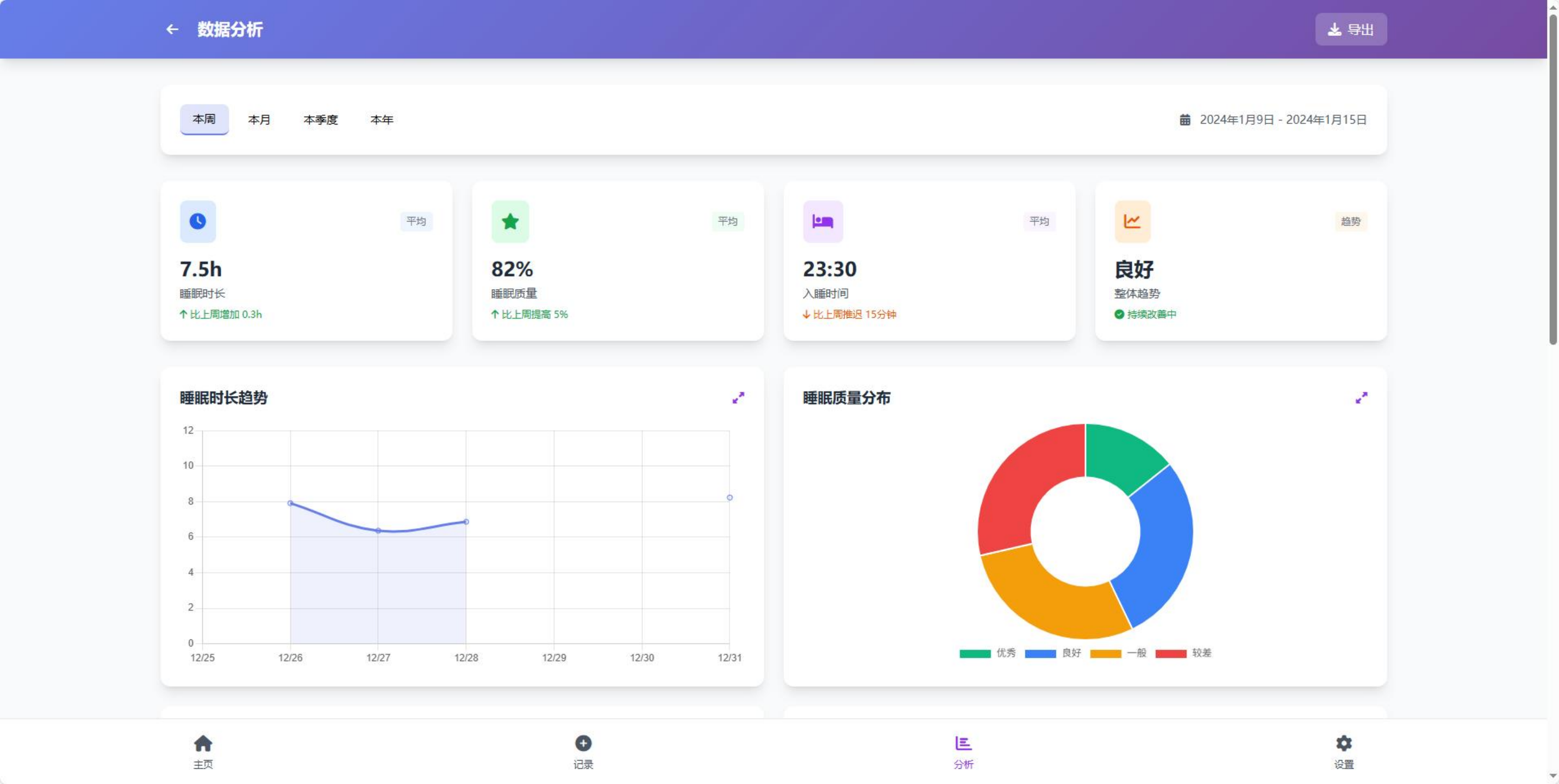}
    \caption{GLM-4.6 screenshot 1}
    \label{fig:ph07}
  \end{subfigure}\hfill
  \begin{subfigure}[t]{\imgW}
    \centering
    \includegraphics[height=\imgH,keepaspectratio]{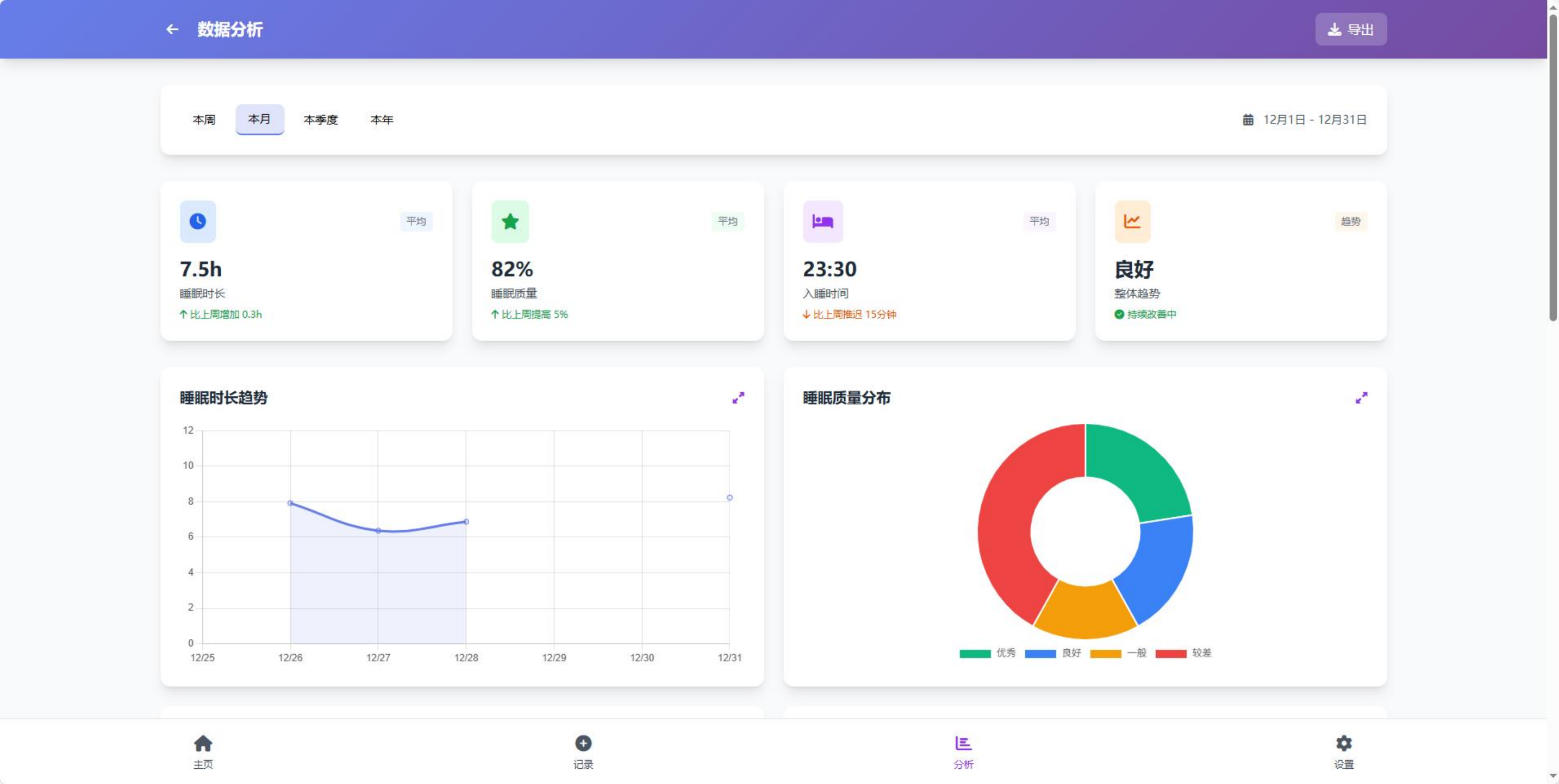}
    \caption{GLM-4.6 screenshot 2}
    \label{fig:ph08}
  \end{subfigure}\hfill
  \begin{subfigure}[t]{\imgW}
    \centering
    \includegraphics[height=\imgH,keepaspectratio]{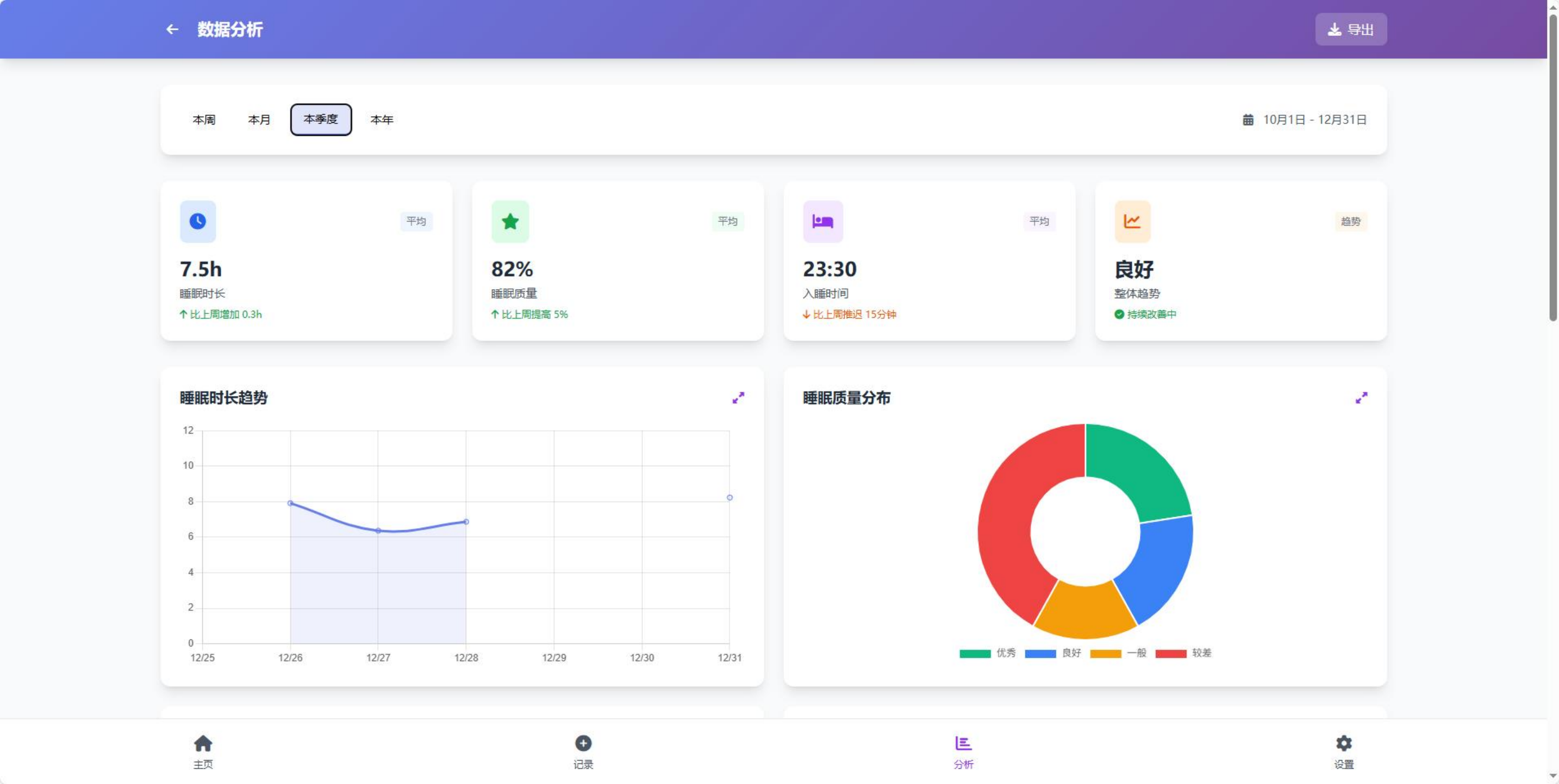}
    \caption{GLM-4.6 screenshot 3}
    \label{fig:ph09}
  \end{subfigure}

  \vspace{0.012\textheight}

  % --- Row 4 ---
  \begin{subfigure}[t]{\imgW}
    \centering
    \includegraphics[height=\imgH,keepaspectratio]{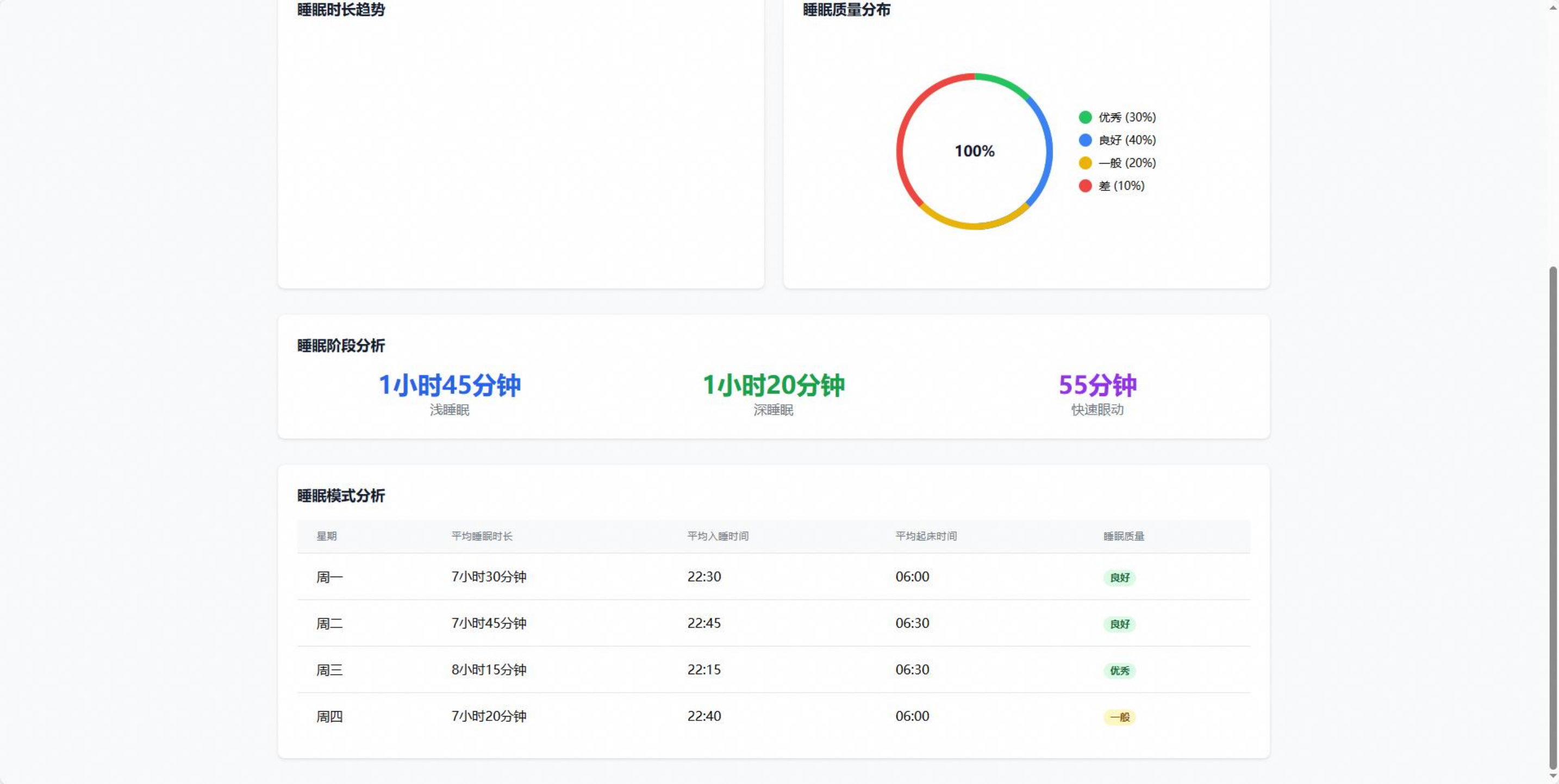}
    \caption{Qwen3-coder-30B screenshot 1}
    \label{fig:ph10}
  \end{subfigure}\hfill
  \begin{subfigure}[t]{\imgW}
    \centering
    \includegraphics[height=\imgH,keepaspectratio]{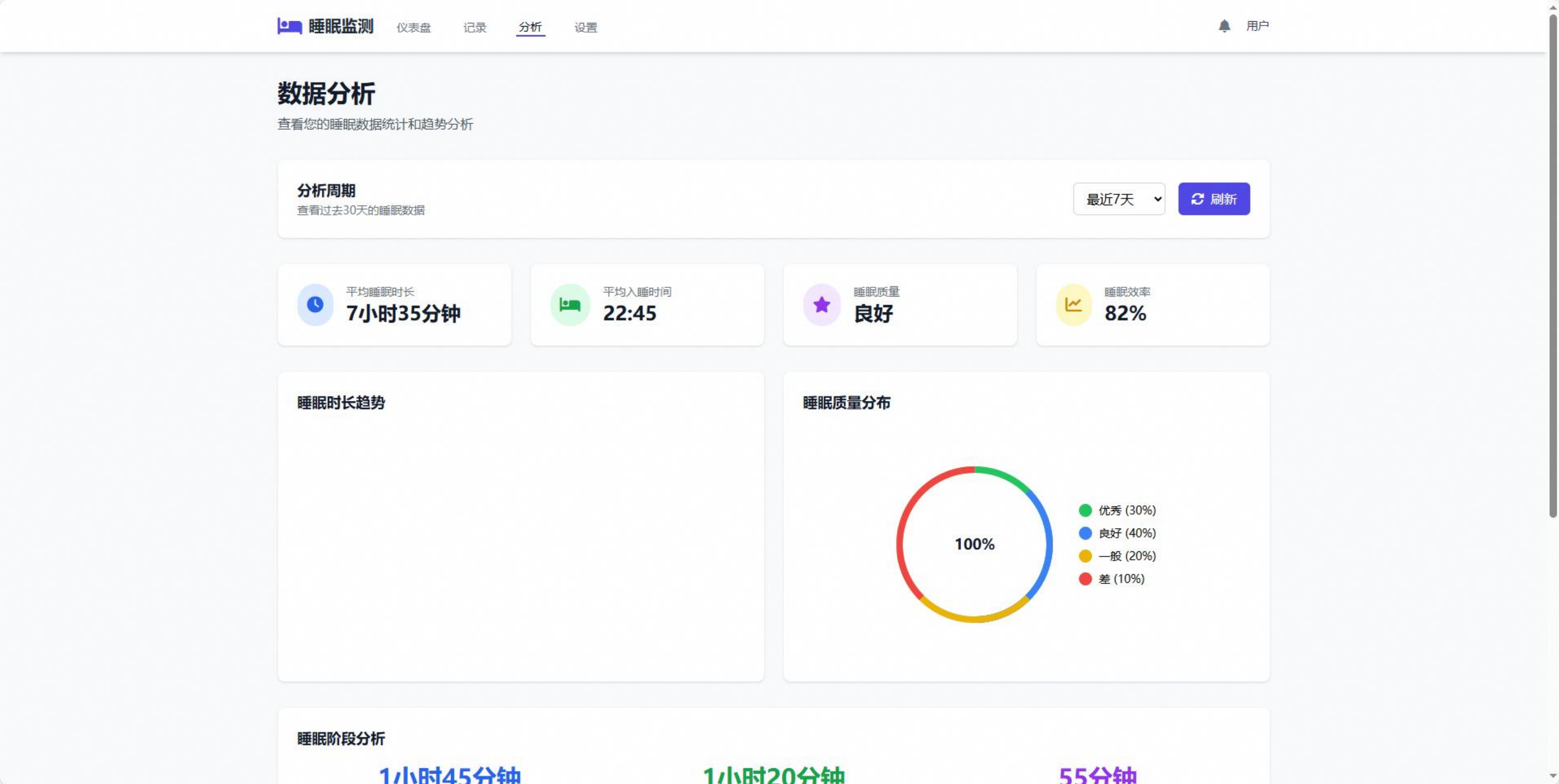}
    \caption{Qwen3-coder-30B screenshot 2}
    \label{fig:ph11}
  \end{subfigure}\hfill
  \begin{subfigure}[t]{\imgW}
    \centering
    \includegraphics[height=\imgH,keepaspectratio]{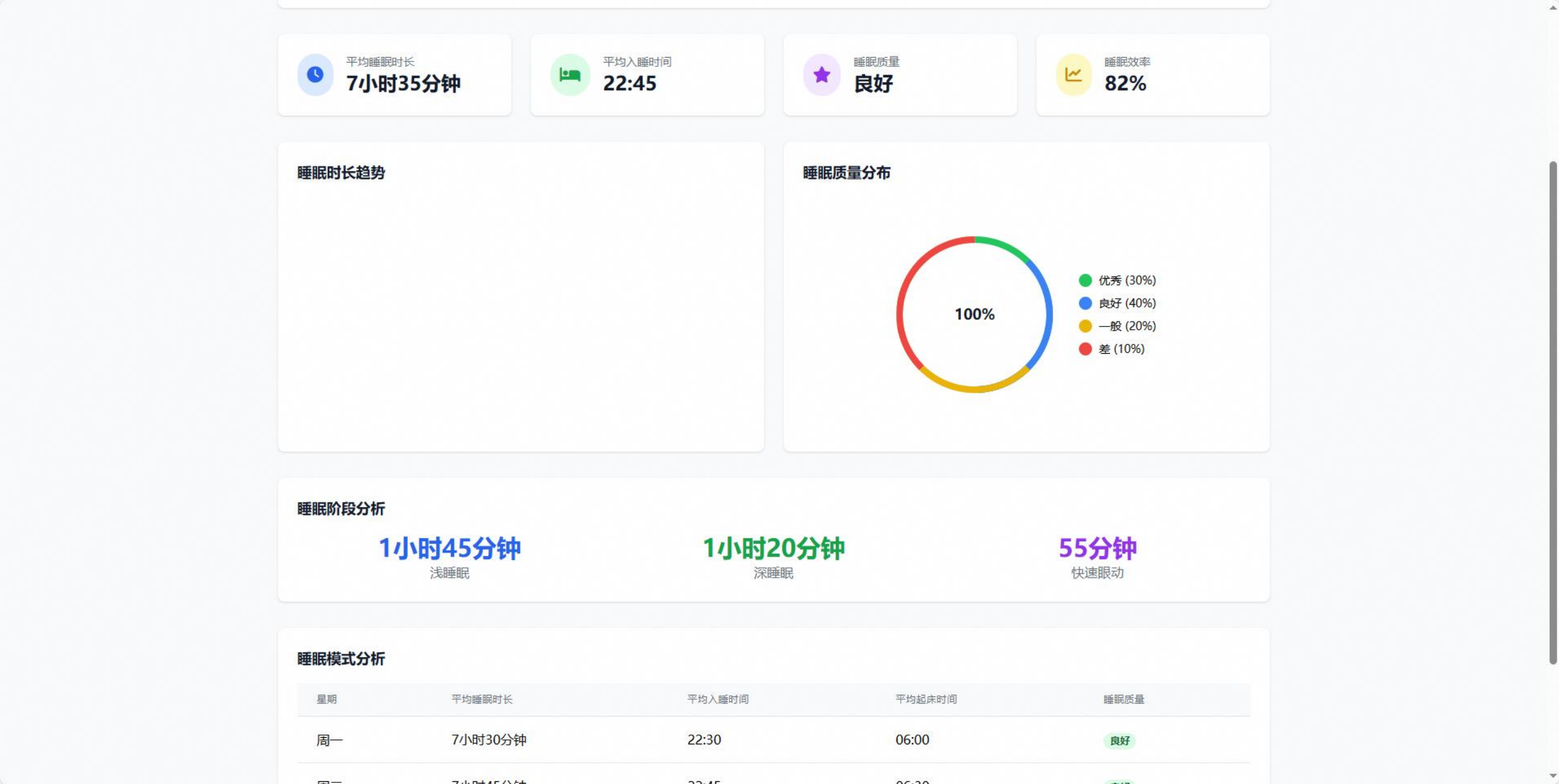}
    \caption{Qwen3-coder-30B screenshot 3}
    \label{fig:ph12}
  \end{subfigure}

  \vspace{0.012\textheight}

  % --- Row 5 ---
  \begin{subfigure}[t]{\imgW}
    \centering
    \includegraphics[height=\imgH,keepaspectratio]{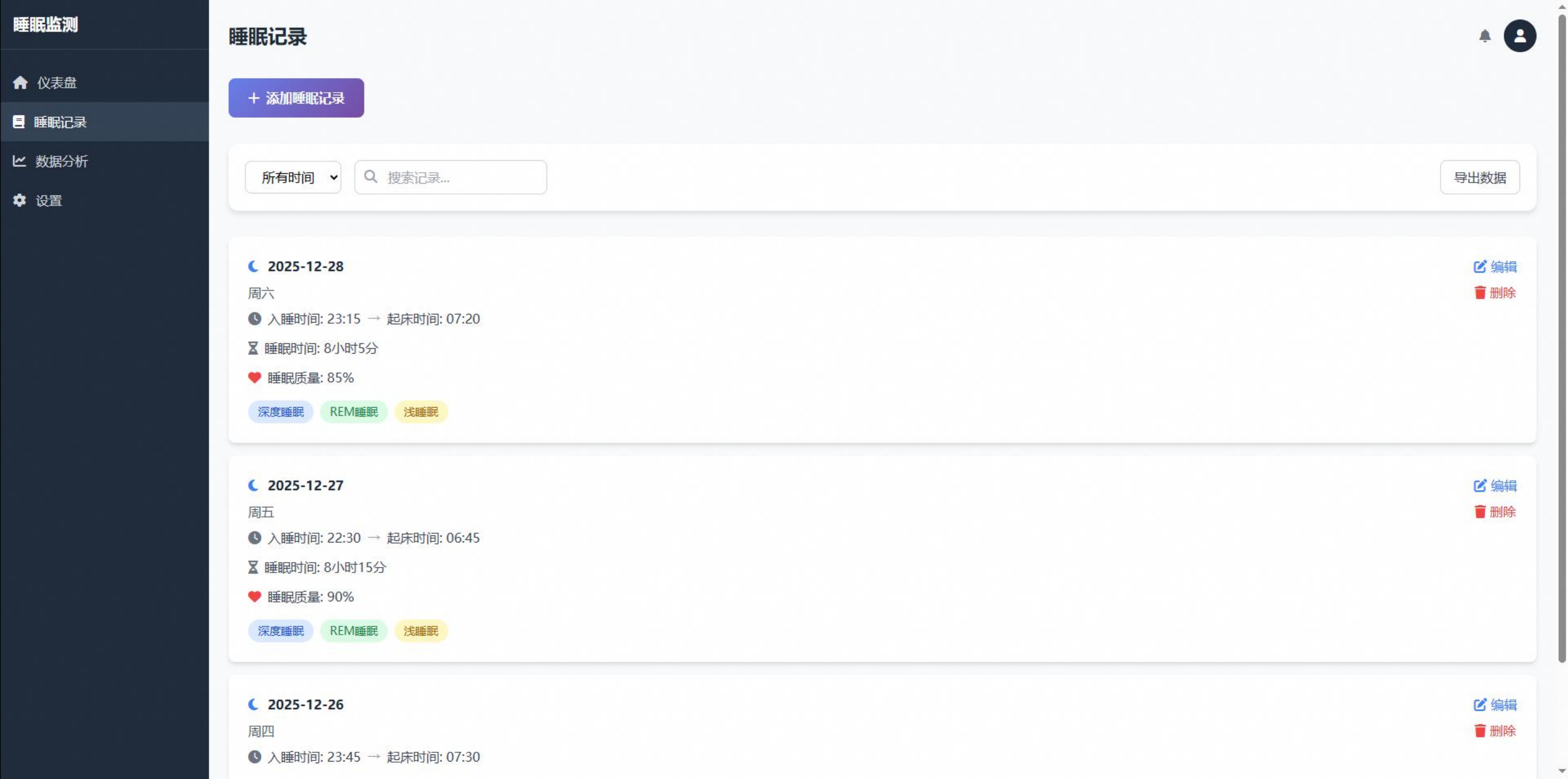}
    \caption{Devstral-Small-2 screenshot 1}
    \label{fig:ph13}
  \end{subfigure}\hfill
  \begin{subfigure}[t]{\imgW}
    \centering
    \includegraphics[height=\imgH,keepaspectratio]{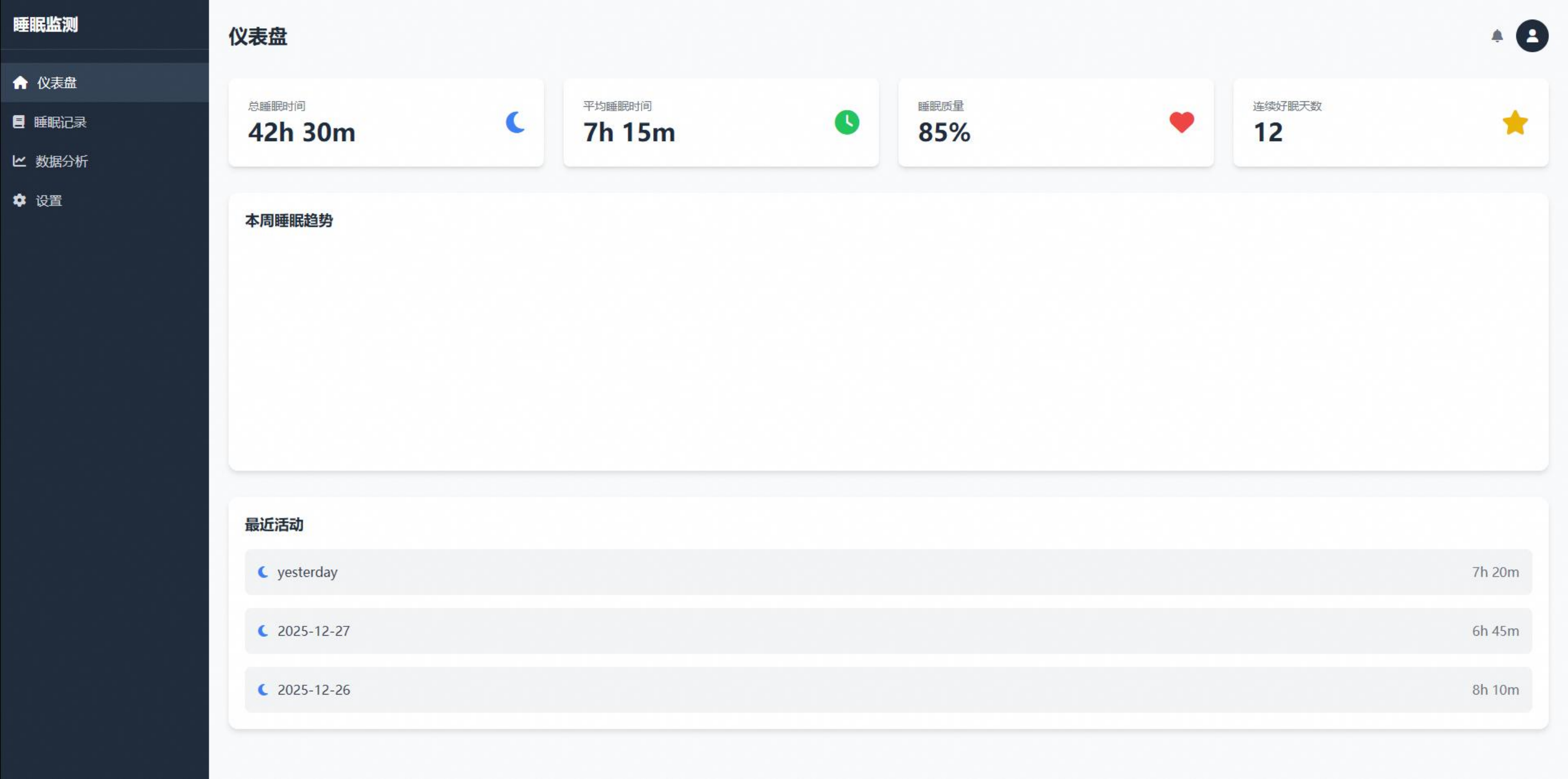}
    \caption{Devstral-Small-2 screenshot 2}
    \label{fig:ph14}
  \end{subfigure}\hfill
  \begin{subfigure}[t]{\imgW}
    \centering
    \includegraphics[height=\imgH,keepaspectratio]{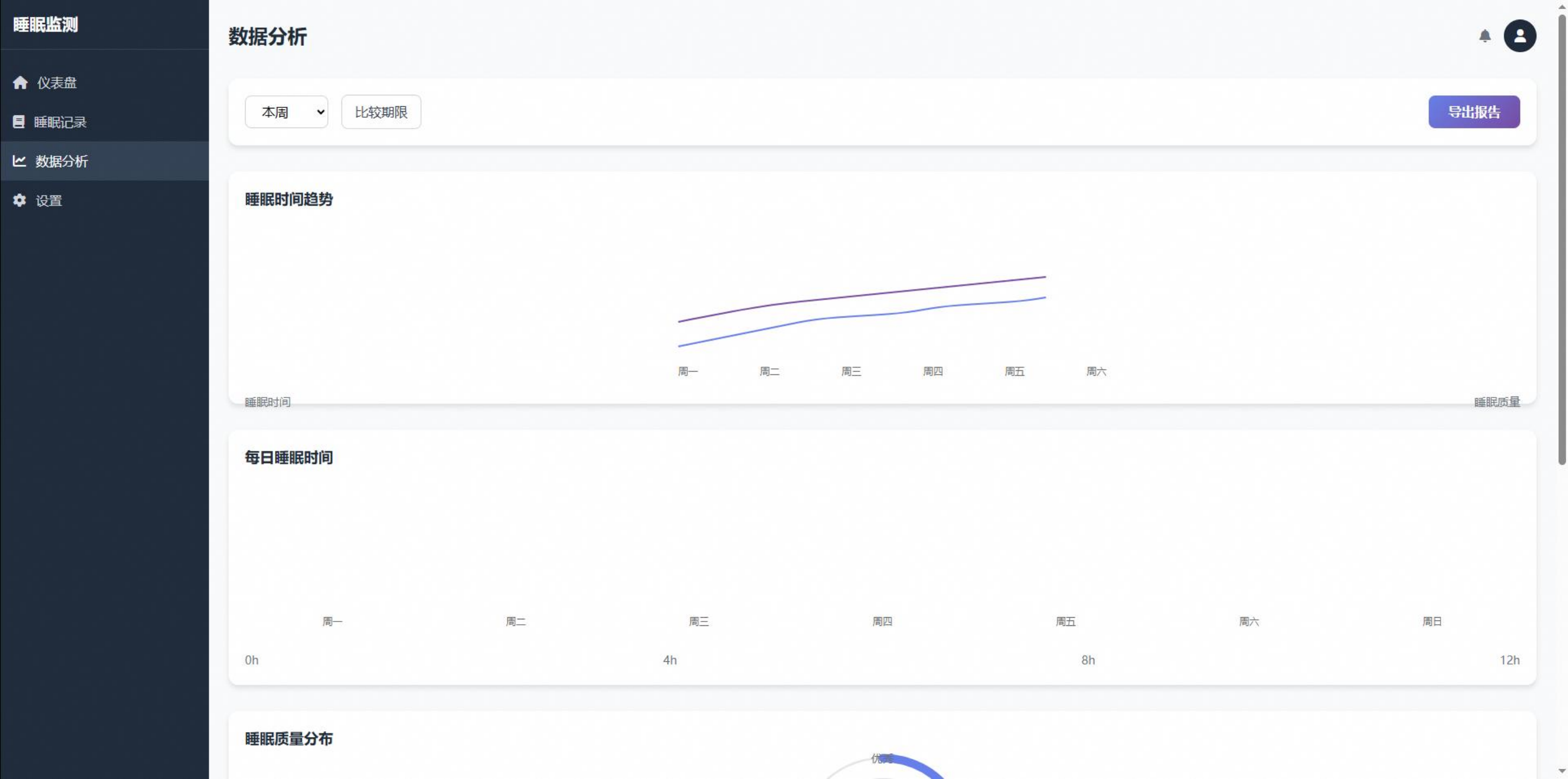}
    \caption{Devstral-Small-2 screenshot 3}
    \label{fig:ph15}
  \end{subfigure}

  \caption{Case study 1 screenshot examples: Sleep Management System Generation.}
  \label{fig:app-case-sleep}
\end{figure}

\begin{figure}[!htbp]
  \centering
  % 关键参数：三列均匀分布；每张图同高（height），并保持比例（keepaspectratio）
  \setlength{\tabcolsep}{0pt} % 去掉列间默认空白
  \renewcommand{\arraystretch}{0} % 去掉行间默认空白（如需更紧凑）
  \newcommand{\imgH}{0.1\textheight} % 5 行 => 每行约 0.2，高度略留空隙
  \newcommand{\imgW}{0.32\textwidth}  % 3 列 + \hfill 间距

  % --- Row 1 ---
  \begin{subfigure}[t]{\imgW}
    \centering
    \includegraphics[height=\imgH,keepaspectratio]{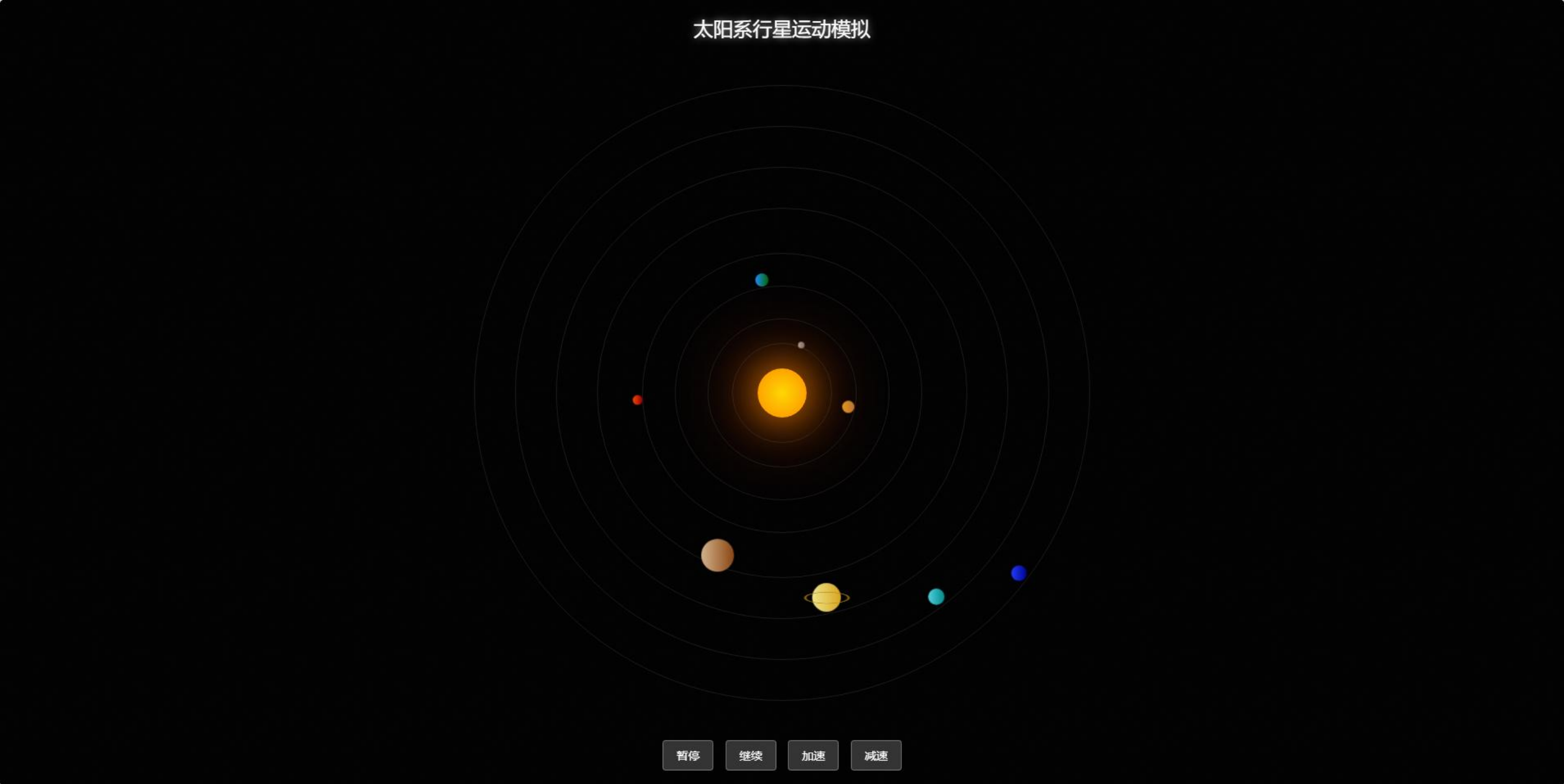}
    \caption{ROME screenshot 1}
    \label{fig:ph01}
  \end{subfigure}\hfill
  \begin{subfigure}[t]{\imgW}
    \centering
    \includegraphics[height=\imgH,keepaspectratio]{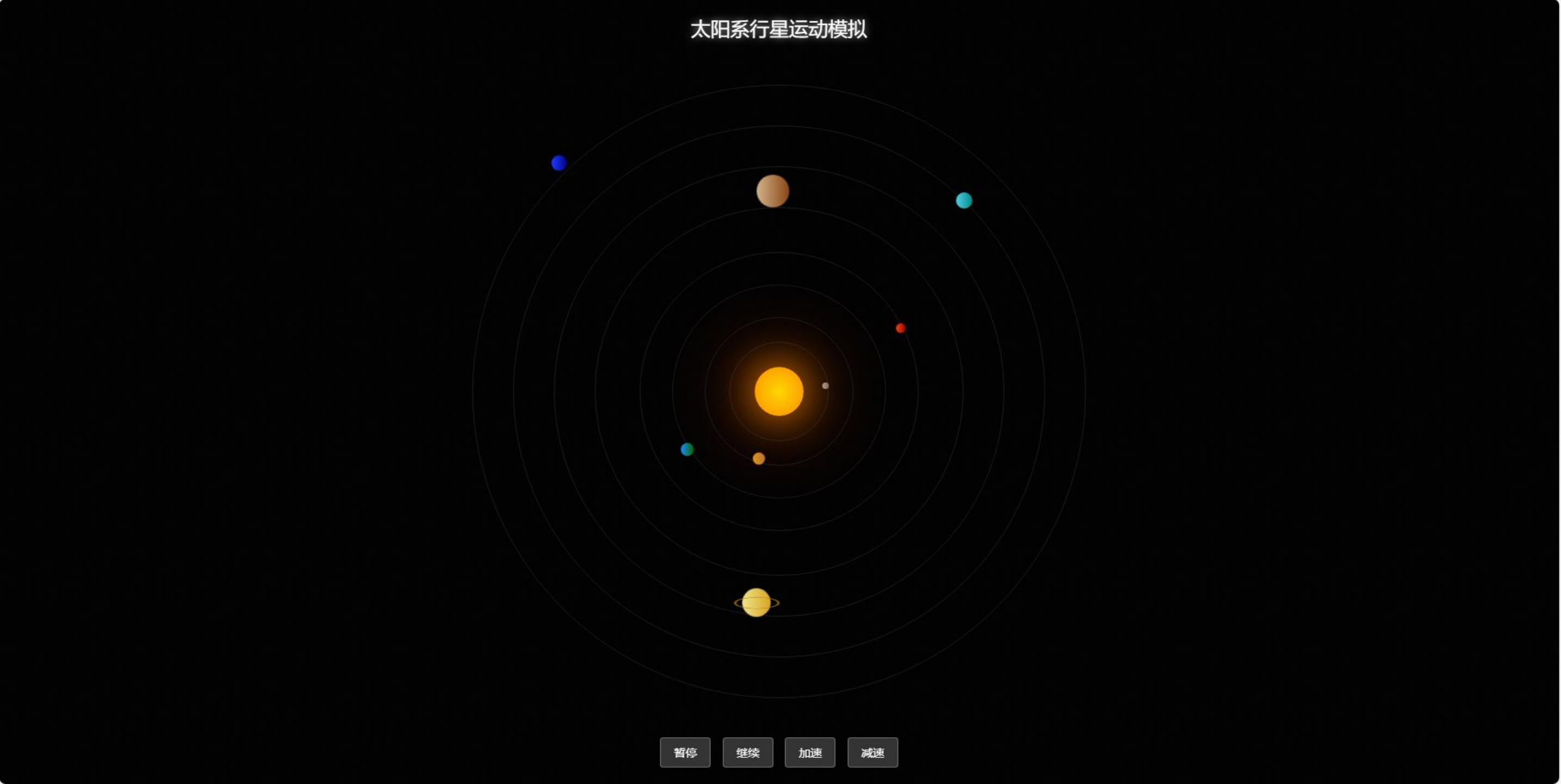}
    \caption{ROME screenshot 2}
    \label{fig:ph02}
  \end{subfigure}\hfill
  \begin{subfigure}[t]{\imgW}
    \centering
    \includegraphics[height=\imgH,keepaspectratio]{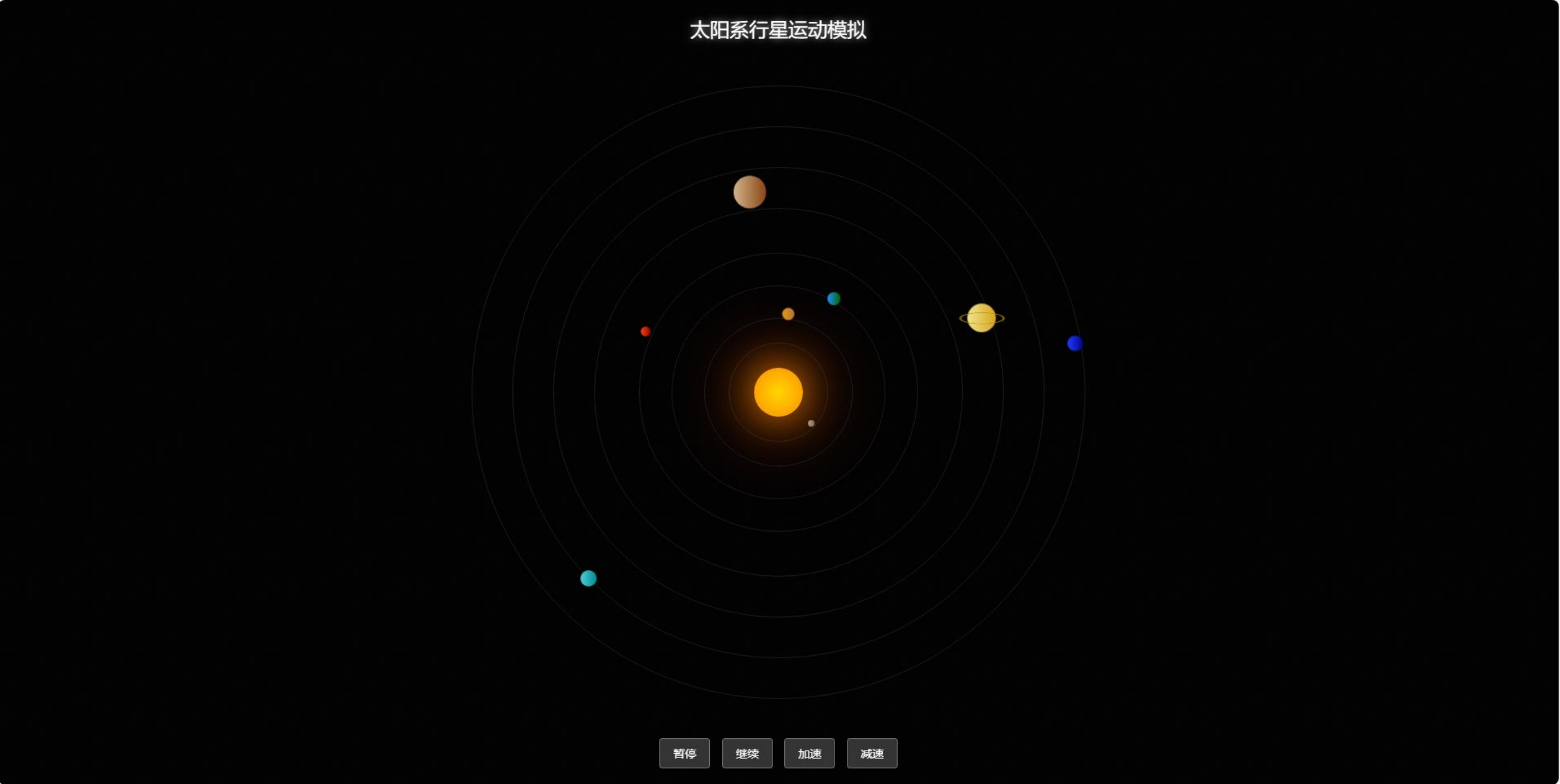}
    \caption{ROME screenshot 3}
    \label{fig:ph03}
  \end{subfigure}

  \vspace{0.012\textheight}

  % --- Row 2 ---
  \begin{subfigure}[t]{\imgW}
    \centering
    \includegraphics[height=\imgH,keepaspectratio]{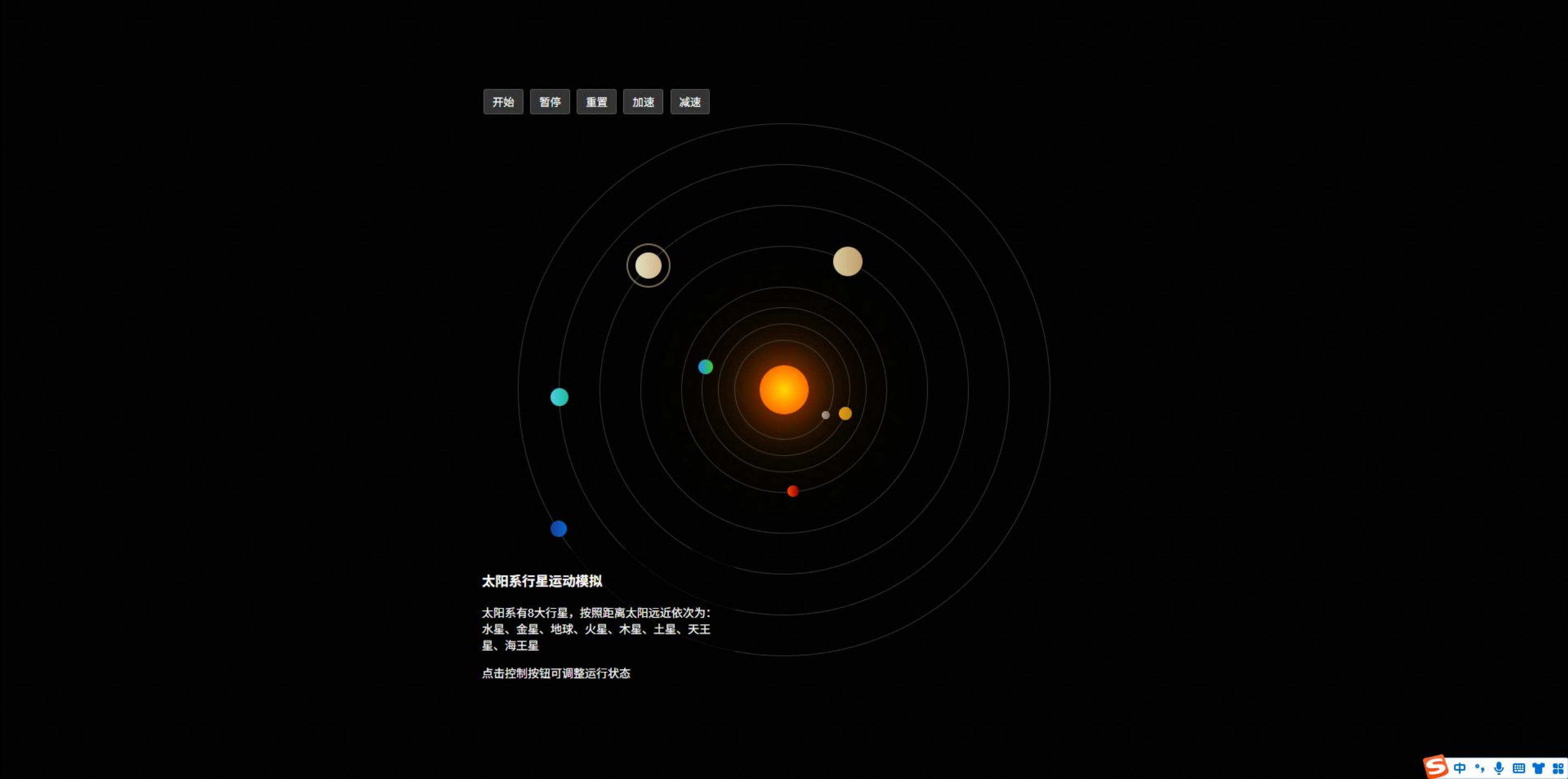}
    \caption{Qwen3-Coder-Plus screenshot 1}
    \label{fig:ph04}
  \end{subfigure}\hfill
  \begin{subfigure}[t]{\imgW}
    \centering
    \includegraphics[height=\imgH,keepaspectratio]{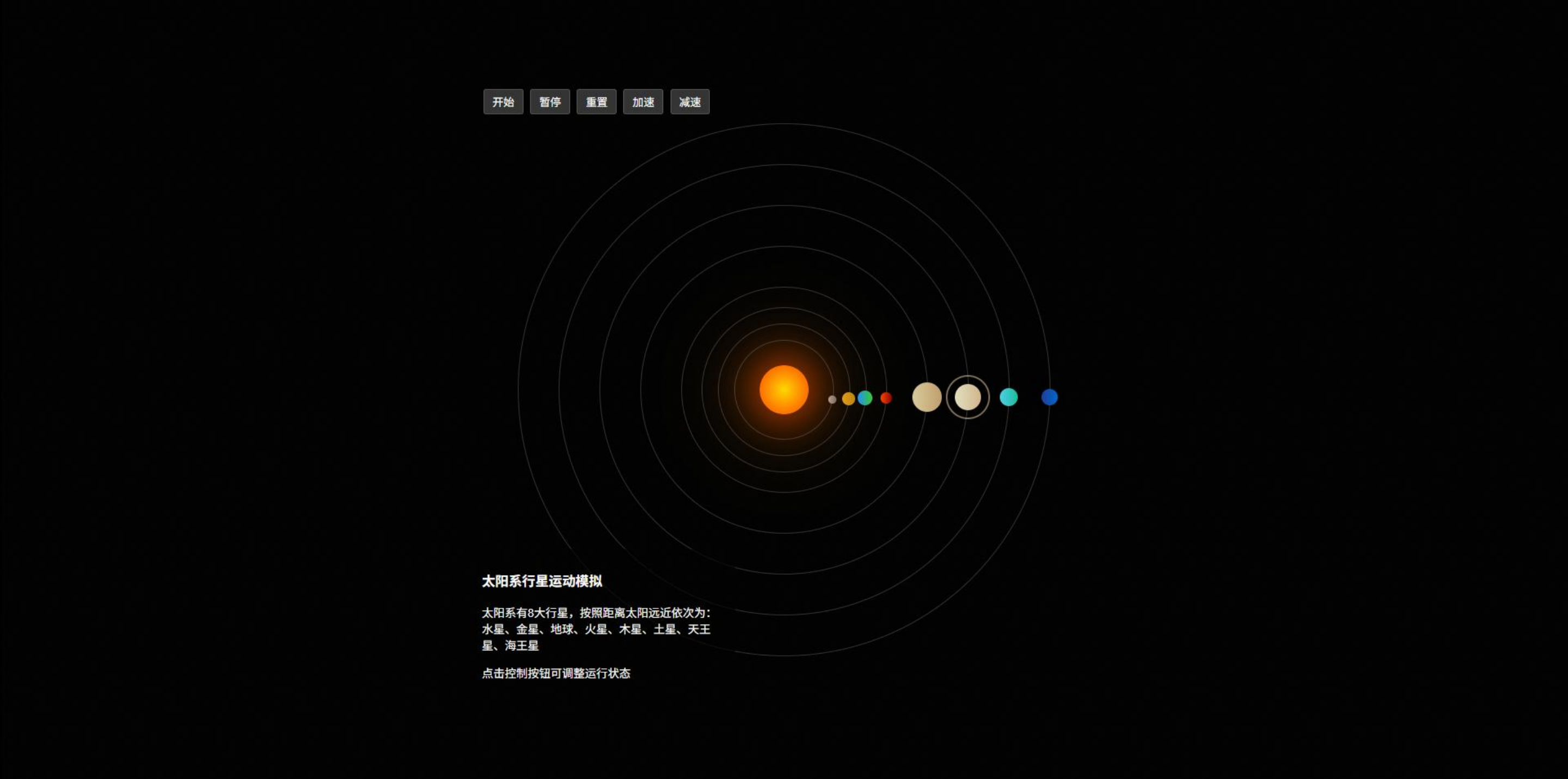}
    \caption{Qwen3-Coder-Plus screenshot 2}
    \label{fig:ph05}
  \end{subfigure}\hfill
  \begin{subfigure}[t]{\imgW}
    \centering
    \includegraphics[height=\imgH,keepaspectratio]{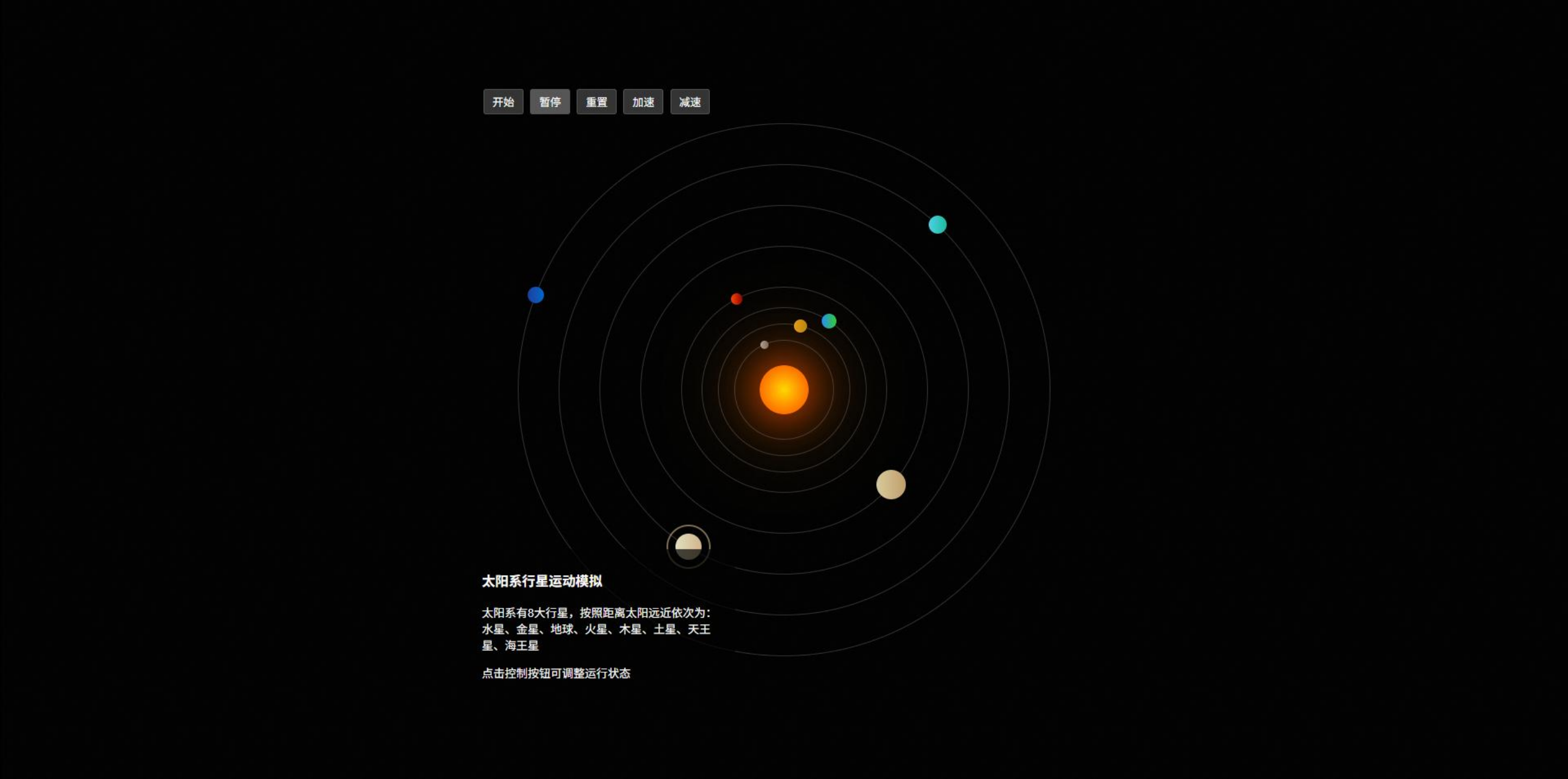}
    \caption{Qwen3-Coder-Plus screenshot 3}
    \label{fig:ph06}
  \end{subfigure}

  \vspace{0.012\textheight}

  % --- Row 3 ---
  \begin{subfigure}[t]{\imgW}
    \centering
    \includegraphics[height=\imgH,keepaspectratio]{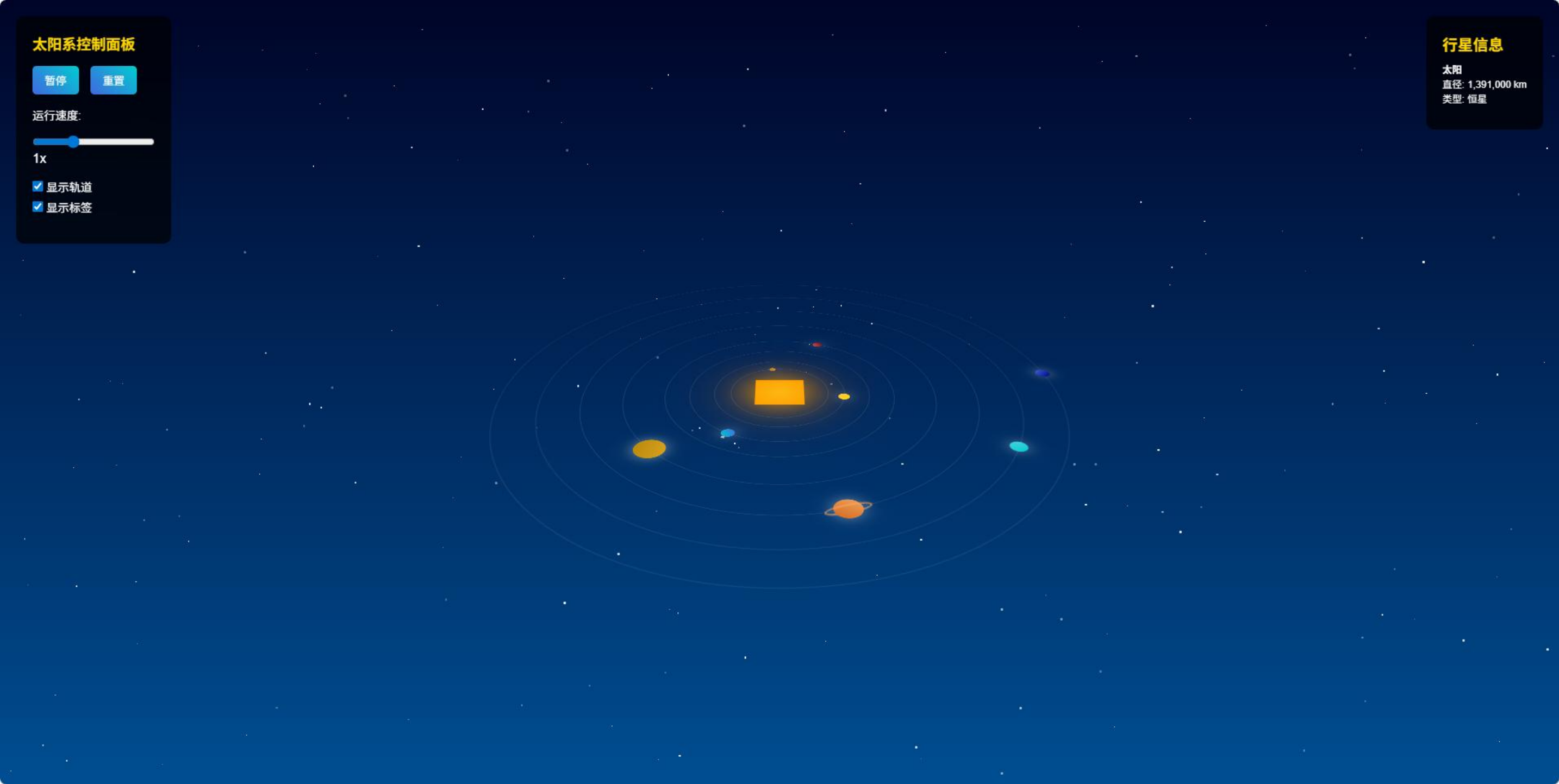}
    \caption{GLM-4.6 screenshot 1}
    \label{fig:ph07}
  \end{subfigure}\hfill
  \begin{subfigure}[t]{\imgW}
    \centering
    \includegraphics[height=\imgH,keepaspectratio]{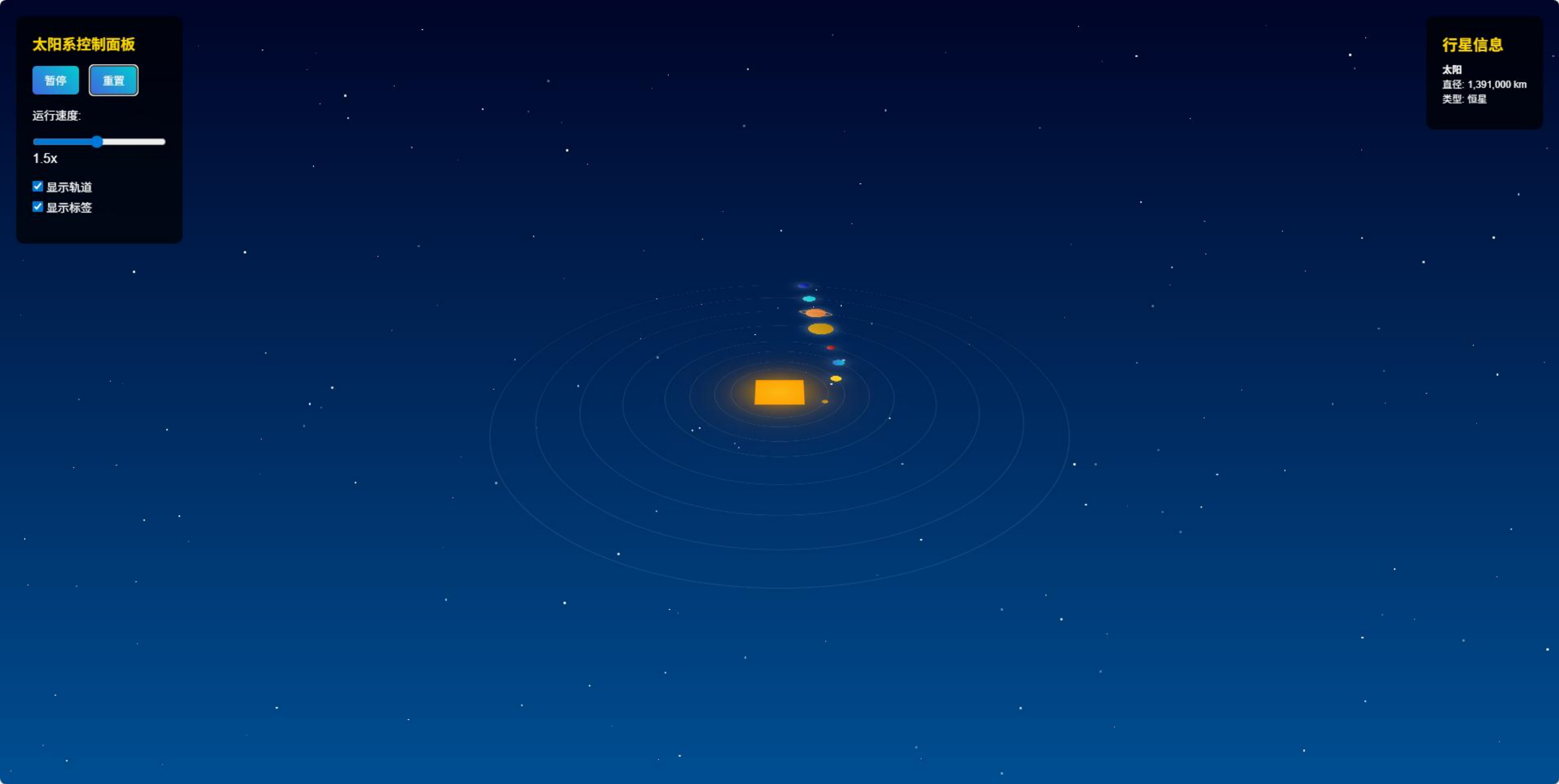}
    \caption{GLM-4.6 screenshot 2}
    \label{fig:ph08}
  \end{subfigure}\hfill
  \begin{subfigure}[t]{\imgW}
    \centering
    \includegraphics[height=\imgH,keepaspectratio]{figures/appendix/solar-system/glm_46/3-glm_46-5.pdf}
    \caption{GLM-4.6 screenshot 3}
    \label{fig:ph09}
  \end{subfigure}

  \vspace{0.012\textheight}

  % --- Row 4 ---
  \begin{subfigure}[t]{\imgW}
    \centering
    \includegraphics[height=\imgH,keepaspectratio]{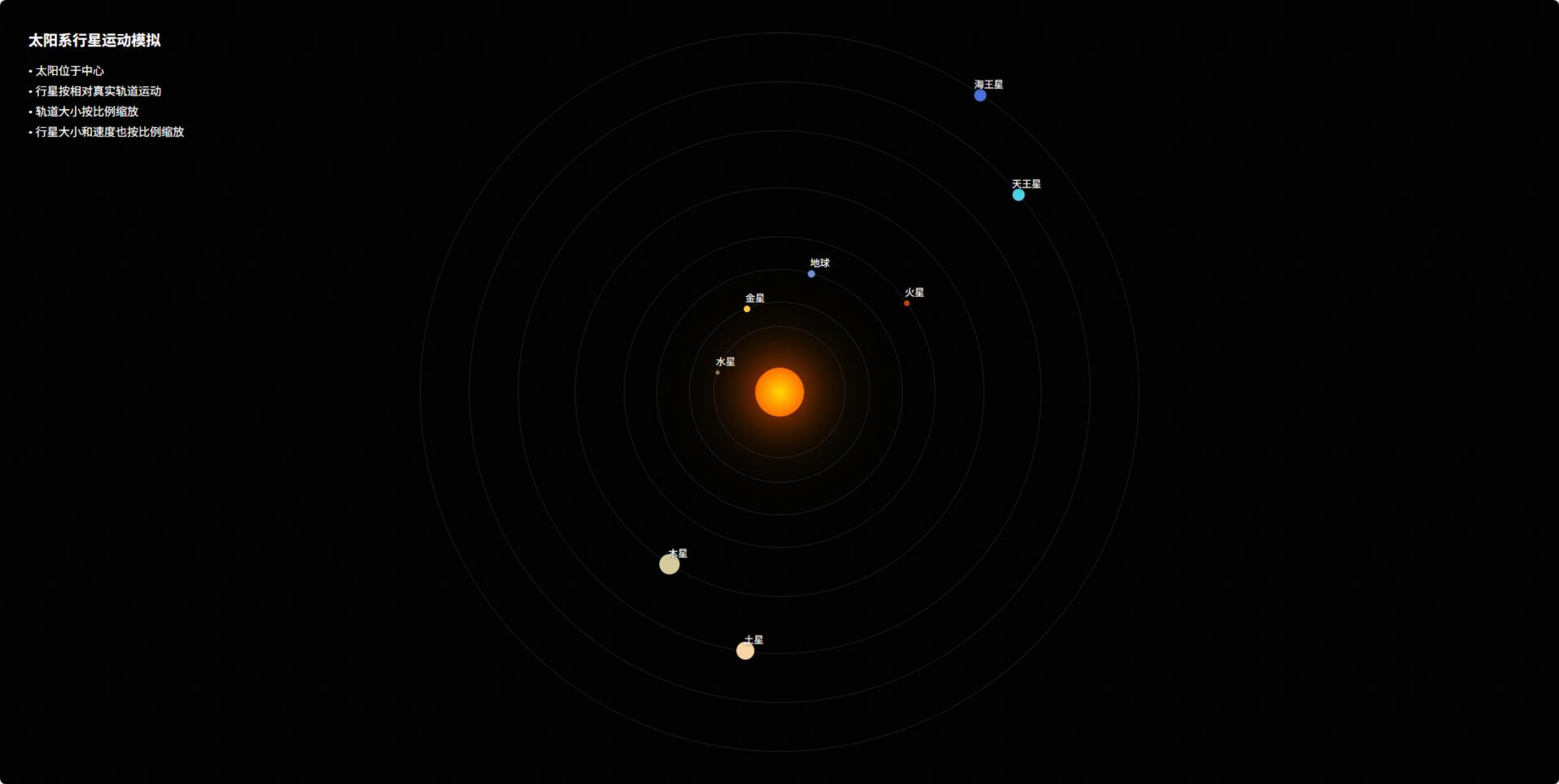}
    \caption{Qwen3-coder-30B screenshot 1}
    \label{fig:ph10}
  \end{subfigure}\hfill
  \begin{subfigure}[t]{\imgW}
    \centering
    \includegraphics[height=\imgH,keepaspectratio]{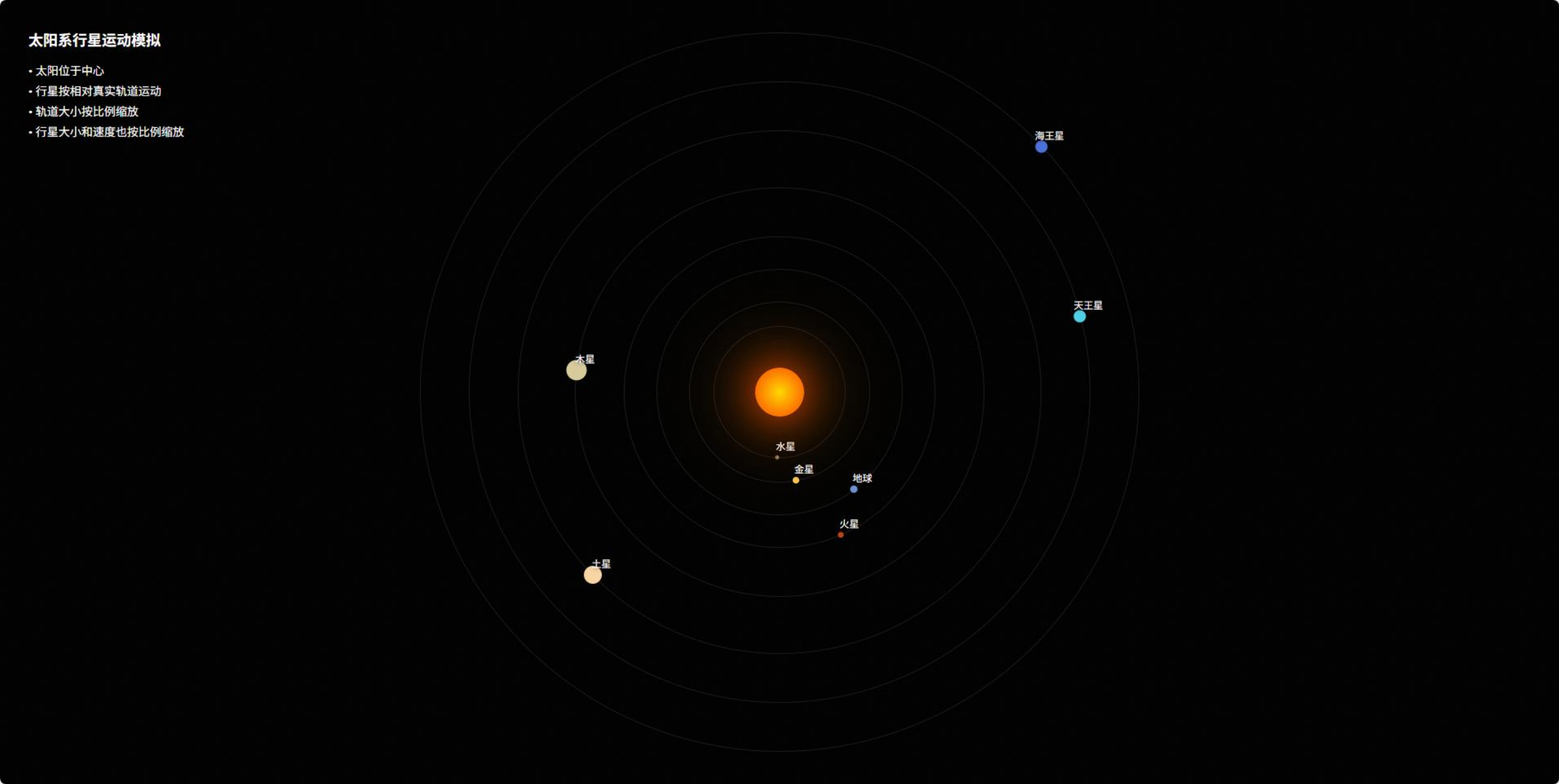}
    \caption{Qwen3-coder-30B screenshot 2}
    \label{fig:ph11}
  \end{subfigure}\hfill
  \begin{subfigure}[t]{\imgW}
    \centering
    \includegraphics[height=\imgH,keepaspectratio]{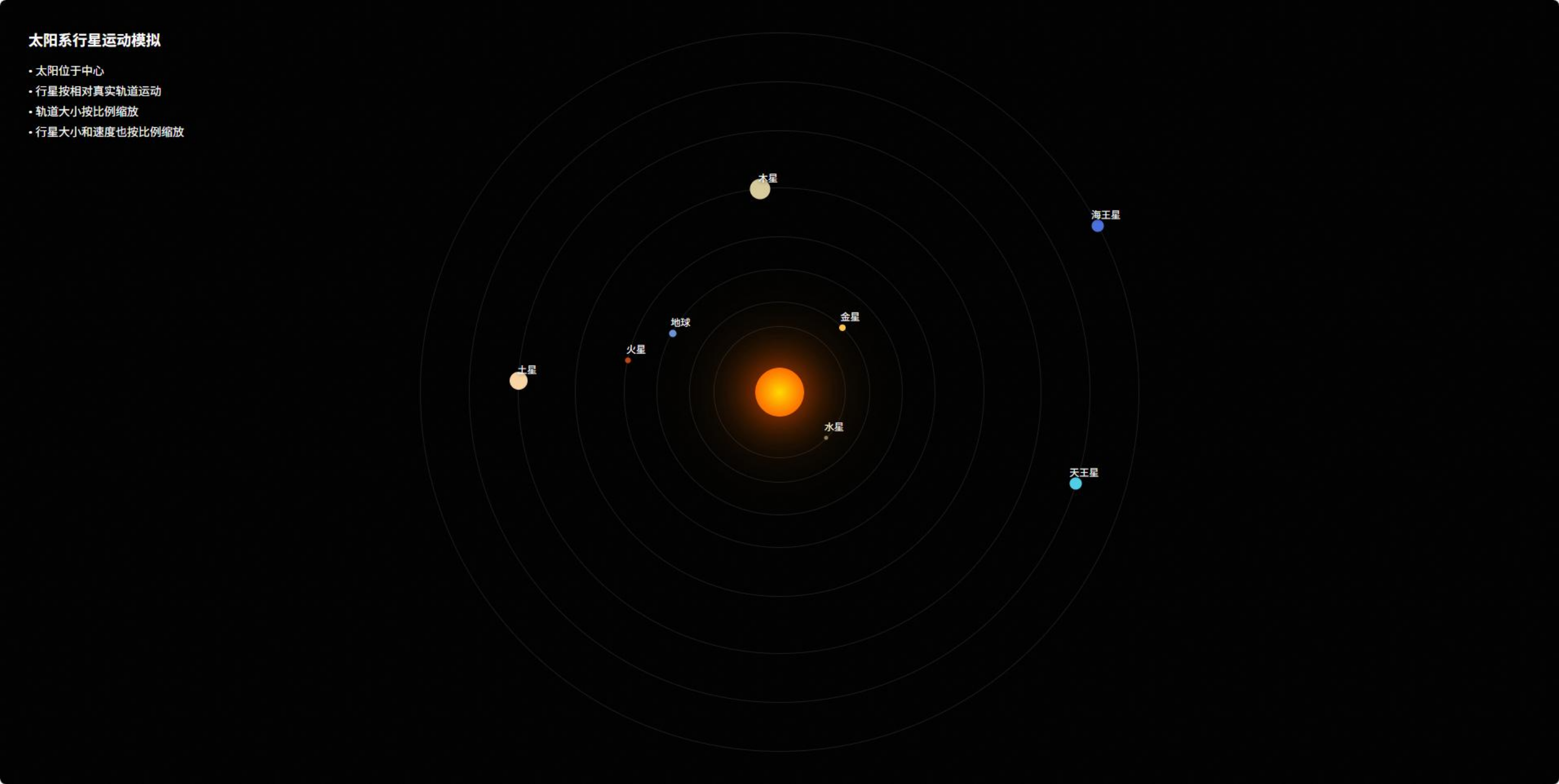}
    \caption{Qwen3-coder-30B screenshot 3}
    \label{fig:ph12}
  \end{subfigure}

  \vspace{0.012\textheight}

  % --- Row 5 ---
  \begin{subfigure}[t]{\imgW}
    \centering
    \includegraphics[height=\imgH,keepaspectratio]{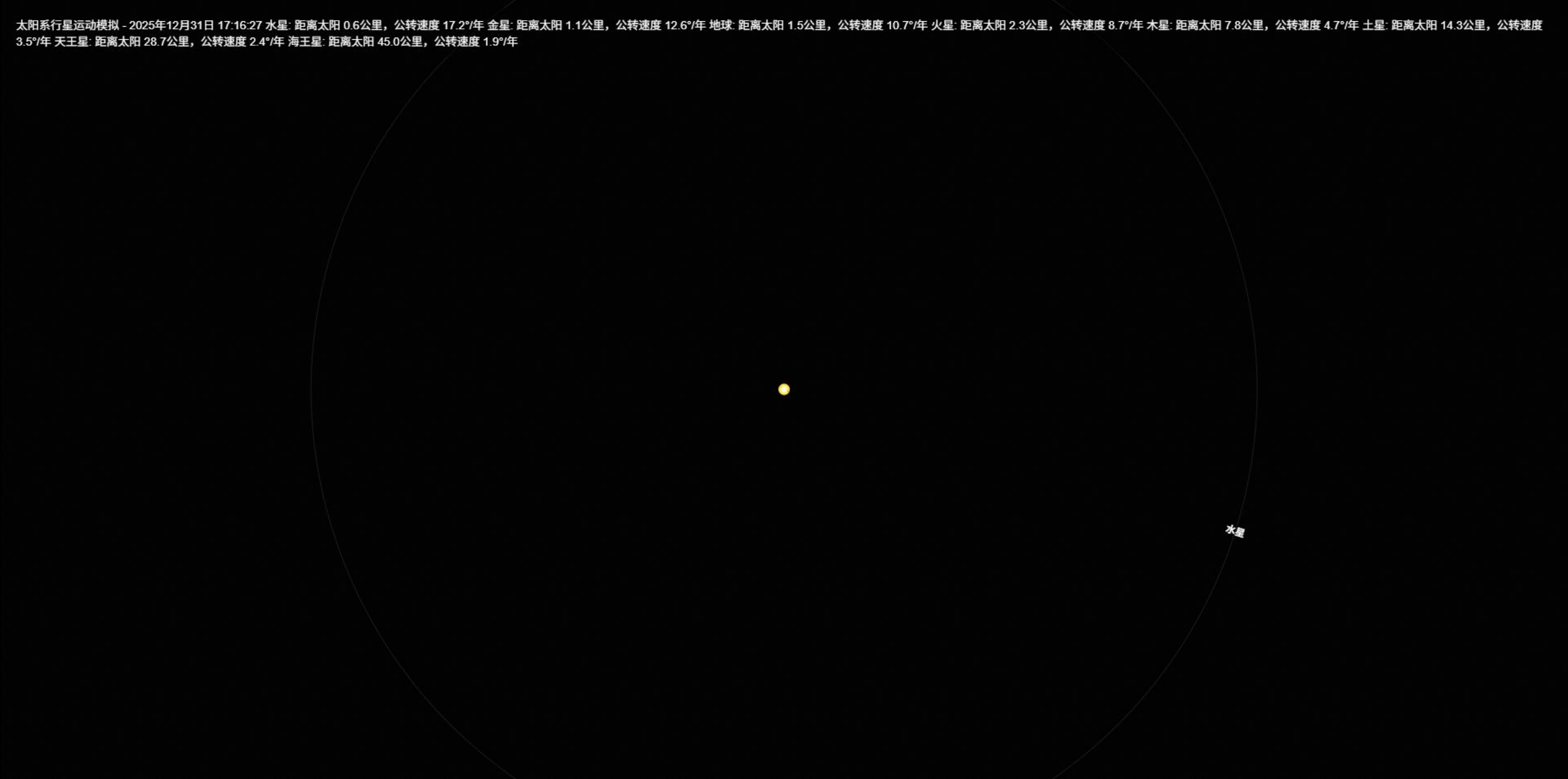}
    \caption{Devstral-Small-2 screenshot 1}
    \label{fig:ph13}
  \end{subfigure}\hfill
  \begin{subfigure}[t]{\imgW}
    \centering
    \includegraphics[height=\imgH,keepaspectratio]{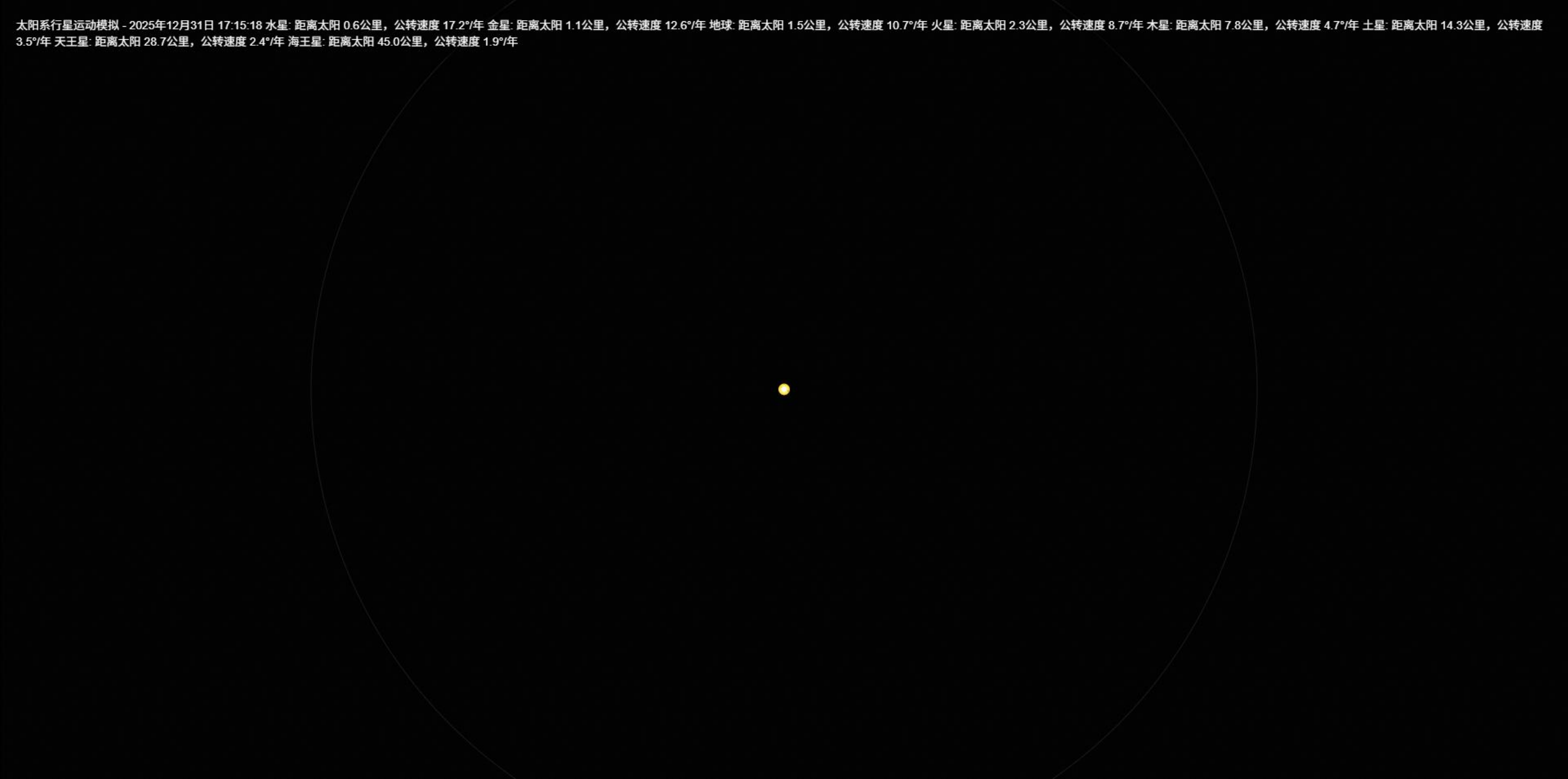}
    \caption{Devstral-Small-2 screenshot 2}
    \label{fig:ph14}
  \end{subfigure}\hfill
  \begin{subfigure}[t]{\imgW}
    \centering
    \includegraphics[height=\imgH,keepaspectratio]{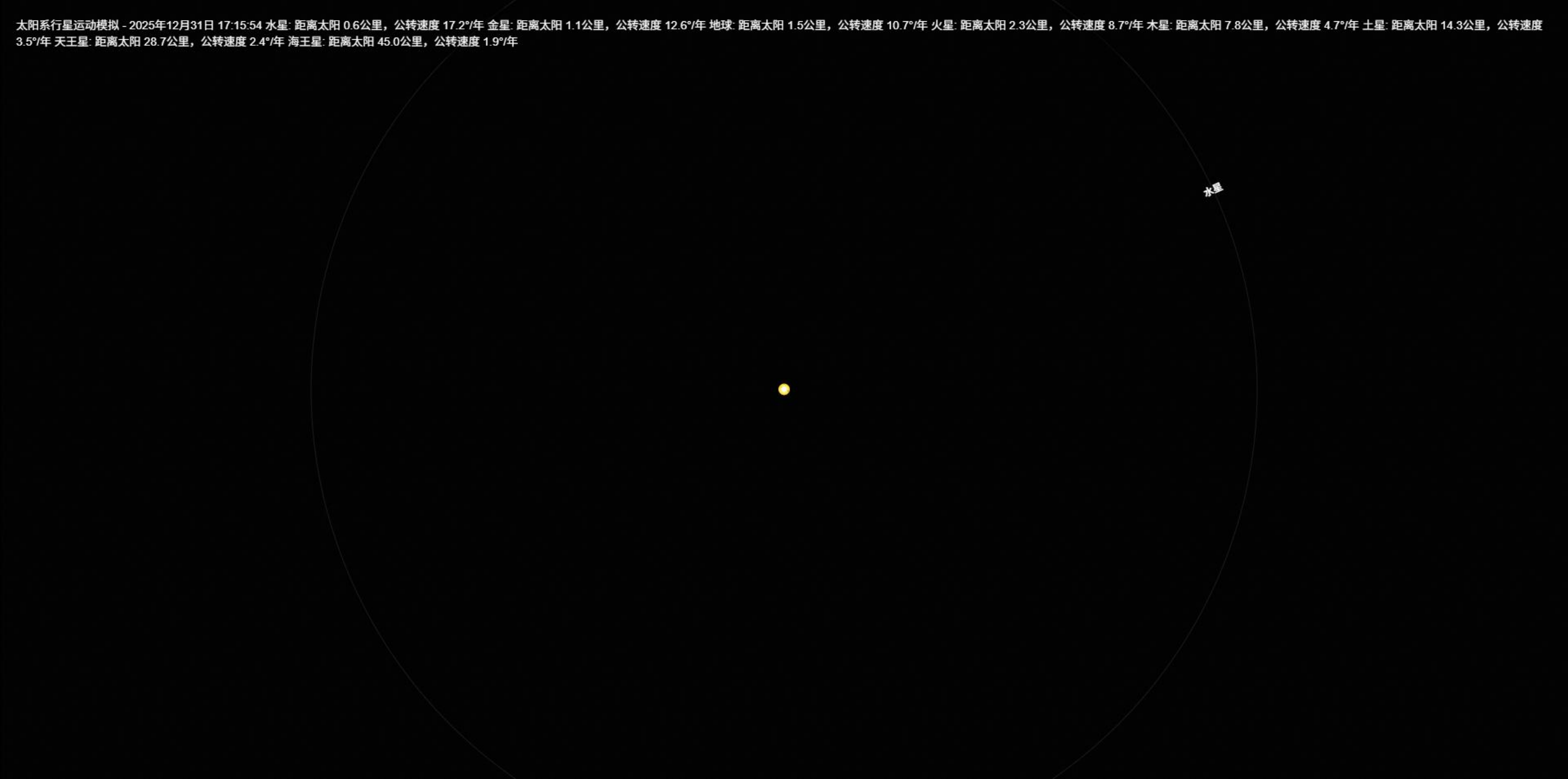}
    \caption{Devstral-Small-2 screenshot 3}
    \label{fig:ph15}
  \end{subfigure}

  \caption{Case study 2 screenshot examples: Solar System Modeling.}
  \label{fig:app-case-solar}
\end{figure}

We also select two representative case studies(Sleep Management System Generation and Solar System Modeling) and present task screenshots in \autoref{fig:app-case-sleep} and \autoref{fig:app-case-solar}, respectively. The detailed scoring rubric is provided in \autoref{tab:app.1}. From the case examples, we can also observe that ROME achieves stronger task-execution performance and better visual/page quality than other models of comparable size, and its results are competitive with those large-scale models.

\begin{table}[t]
\centering
\caption{Case-study evaluation scores, reported as the average ratings across 30 experts.}
\label{tab.app.1}
\resizebox{\linewidth}{!}{
\begin{tabular}{l l | c | c c c c}
\toprule
\textbf{Metric} & \textbf{Sub-metric} 
& \textbf{ROME}
& \textbf{Qwen-Coder-30B} 
& \textbf{Qwen3-Coder-Plus} 
& \textbf{Devstral-Small} 
& \textbf{GLM-4.6} \\
\midrule

\multicolumn{7}{c}{\textbf{Sleep Management System Generation}} \\
\midrule
Functionality & Interaction          & 39 & 39 & 39 & 39 & 40 \\
Layout        & Style Restoration    & 18 & 16 & 18 & 15 & 18 \\
Code Quality  & Robustness           & 20 & 20 & 20 & 19 & 20 \\
Structure     & Semantic Correctness & 7  & 7  & 7  & 6  & 7  \\
Innovation    & Prompt Understanding & 8  & 7  & 8  & 7  & 8  \\
\midrule
\textbf{Total Score} & -- & \textbf{92} & 89 & 92 & 86 & 93 \\
\midrule

\multicolumn{7}{c}{\textbf{Solar System Modeling}} \\
\midrule
Functionality & Interaction          & 34 & 35 & 36 & 10 & 34 \\
Layout        & Style Restoration    & 20 & 20 & 20 & 5  & 20 \\
Code Quality  & Robustness           & 20 & 20 & 20 & 7  & 16 \\
Structure     & Semantic Correctness & 10 & 10 & 10 & 3  & 10 \\
Innovation    & Prompt Understanding & 10 & 6  & 10 & 5  & 10 \\
\midrule
\textbf{Total Score} & -- & \textbf{94} & 91 & 96 & 30 & 90 \\
\bottomrule
\end{tabular}
}
\end{table}

\clearpage
\bibliography{biblio}
\bibliographystyle{colm2024_conference}

\end{document}